%% file: thesis.tex
\documentclass[12pt,a4paper,twoside,titlepage]{book}

\usepackage[inner=3.3cm,outer=2.7cm]{geometry}
\usepackage{ragged2e}

\usepackage{algorithm}
\usepackage{algpseudocode}
\usepackage{setspace}  
\usepackage{changepage}

\usepackage{lmodern}

\usepackage{tgtermes}
\usepackage[T1]{fontenc}

\usepackage[utf8]{inputenc}

\usepackage[ngerman,english]{babel}

\usepackage{newunicodechar}
\newunicodechar{ }{~}

\usepackage{pdfpages}

\usepackage[nottoc,notlof,notlot]{tocbibind}

\usepackage{graphicx}

\usepackage{pgfplots}
\pgfplotsset{compat=newest}

\usepackage[square, numbers]{natbib}
\usepackage{bm}
\usepackage{amsmath}
\usepackage{amssymb}
\usepackage{mathtools}
\usepackage{amsthm}
\usepackage{siunitx}
\usepackage{setspace}
\usepackage{array}
\usepackage{booktabs}
\usepackage{multirow}
\usepackage{multicol}
\usepackage{comment}
\usepackage{lscape}
\usepackage{fancyhdr}
\usepackage{colortbl}
\usepackage{listings}
\usepackage{collcell}
\usepackage{hhline}
\usepackage{dcolumn}
\usepackage{makecell}
\usepackage{sectsty}
\usepackage{caption}
\usepackage{subcaption}

\usepackage{todonotes}
\usepackage{subfiles}
\usepackage{siunitx}
\definecolor{asphalt}{RGB}{228, 26, 28}
\definecolor{grass}{RGB}{152, 78, 163}
\definecolor{gravel}{RGB}{55, 126, 184}
\definecolor{parking}{RGB}{77, 175, 74}
\definecolor{cobble}{RGB}{255, 127, 0}

\def\colorModel{hsb} 

\newcolumntype{P}[1]{>{\centering\arraybackslash}p{#1}}
\newcolumntype{d}[1]{D{.}{.}{#1}}

\newcommand\ColCell[1]{
  \pgfmathparse{#1<100?1:0}  
    \ifnum\pgfmathresult=0\relax\color{white}\fi
  \pgfmathsetmacro\compA{0}      
  \pgfmathsetmacro\compB{#1/200} 
  \pgfmathsetmacro\compC{1}      
  \edef\x{\noexpand\centering\noexpand\cellcolor[\colorModel]{\compA,\compB,\compC}}\x #1
  } 
\newcolumntype{E}{>{\collectcell\ColCell}m{0.6cm}<{\endcollectcell}}  

\usepackage[
unicode=true, 
bookmarks=true,
bookmarksnumbered=true,
bookmarksopen=false,
breaklinks=false,
backref=false,
colorlinks=false,
hidelinks]{hyperref}

\usepackage{cleveref}

\usepackage{sectsty}  

\DeclareMathAlphabet{\pazocal}{OMS}{zplm}{m}{n}

\DeclareMathOperator*{\argmin}{arg\,min}

\newcommand{\etal}{\emph{et al.}}

\theoremstyle{definition}

\definecolor{asphalt}{RGB}{228, 26, 28}
\definecolor{grass}{RGB}{152, 78, 163}
\definecolor{gravel}{RGB}{55, 126, 184}
\definecolor{parking}{RGB}{77, 175, 74}
\definecolor{cobble}{RGB}{255, 127, 0}
\definecolor{marking}{RGB}{230,120,10}

\definecolor{sky}{RGB}{70, 130, 180}
\definecolor{road}{RGB}{70, 70, 70}
\definecolor{building}{RGB}{128, 64, 128}
\definecolor{sidewalk}{RGB}{244, 35, 232}
\definecolor{bicycle}{RGB}{190, 153, 153}
\definecolor{vegetation}{RGB}{107, 142, 35}
\definecolor{pole}{RGB}{255, 165, 0}
\definecolor{person}{RGB}{220, 20, 60}
\definecolor{car}{RGB}{0, 255, 0}
\definecolor{terrain}{RGB}{255, 255, 0}
\definecolor{fence}{RGB}{0, 255, 255}
\definecolor{curb}{RGB}{168, 168, 168}

\definecolor{road}{RGB}{128, 148, 255}
\definecolor{crossing}{RGB}{42, 255, 61}
\definecolor{pedestrian}{RGB}{255, 226, 108}
\definecolor{obstacle}{RGB}{178, 73, 73}

\definecolor{codegreen}{rgb}{0,0.6,0}
\definecolor{codegray}{rgb}{0.5,0.5,0.5}
\definecolor{codepurple}{rgb}{0.58,0,0.82}
\definecolor{backcolour}{rgb}{0.95,0.95,0.92}

\definecolor{magenta}{RGB}{255,32,255}
\definecolor{green}{RGB}{0,255,0}
\definecolor{blue}{RGB}{0,0,255}
\definecolor{tablehighlight}{rgb}{0.8,0.8,0.8}

\renewcommand{\baselinestretch}{0.985}

\newcolumntype{R}[2]{%
    >{\adjustbox{angle=#1,lap=\width-(#2)}\bgroup}%
    l%
    <{\egroup}%
}

\lstdefinestyle{mystyle}{
    backgroundcolor=\color{backcolour},   
    commentstyle=\color{codegreen},
    keywordstyle=\color{magenta},
    numberstyle=\tiny\color{codegray},
    stringstyle=\color{codepurple},
    basicstyle=\ttfamily\footnotesize,
    breakatwhitespace=false,         
    breaklines=true,                 
    captionpos=b,                    
    keepspaces=true,                 
    numbers=left,                    
    numbersep=5pt,                  
    showspaces=false,                
    showstringspaces=false,
    showtabs=false,                  
    tabsize=2
}

\usepackage{pifont}

\renewcommand{\baselinestretch}{0.98}

\renewcommand{\eqref}[1]{Eq.~(\ref{#1})}

\newcolumntype{P}[1]{>{\centering\arraybackslash}p{#1}}
\newcolumntype{d}[1]{D{.}{.}{#1}}

\usepackage{helvet}

\hypersetup{
	pdftitle={Dissertation <Title>},
	pdfauthor={<Your name>},
	pdfsubject={%
		Dissertation zur Erlangung des Doktorgrades an der Technischen Universität Nürnberg},
	pdfkeywords={Dissertation},
	pdfpagelayout={OneColumn}
}

\begin{document}

\selectlanguage{english}
	
	\frontmatter
	\pagenumbering{Roman}
	
	\begin{titlepage}
		\input{0-title_page}

	\end{titlepage}
	\thispagestyle{empty}
	\cleardoublepage
	
 
	\selectlanguage{ngerman}
	\input{abstract/abstract_de.tex}
	
	\selectlanguage{english}
	\input{abstract/abstract_en.tex}

        \clearpage
	\input{acknowledgements}

	\cleardoublepage
	\tableofcontents
	
	\mainmatter

         
	\input{1-introduction}

\subfile{publications/fmloc}

\subfile{publications/langrasp}

\subfile{publications/vlmvac}

\subfile{publications/ARRO}

    \input{3-conclusion}

\input{4-remark}

    \cleardoublepage
    \listoffigures

    \cleardoublepage
    \listoftables

	\bibliographystyle{IEEEtran}
	\bibliography{thesis}

	\cleardoublepage
	\thispagestyle{empty}
	\hbox{}
	\newpage

\end{document}

%% file: 0-title_page.tex

\begin{titlepage}
\thispagestyle{empty}
    \centering

    \vspace{1cm}
    
    {\LARGE \textbf{Leveraging Foundation Models \\[0.5cm] for Enhancing Robot Perception and Action}}\\[1.5cm]
    \vspace{20mm}
    \includegraphics[width=4cm]{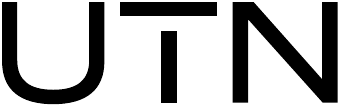} 
    \\
    \vspace{20mm}
    {\large Dissertation submitted for the Degree of}\\
    \vspace{3mm}
    {\large Dr.~rer.~nat.}\\[1.5cm]
    
    {\large by}\\[0.5cm]
    {\Large \textbf{Reihaneh Mirjalili}}\\[1.5cm]
    
    {\large Department Computer Science \& Artificial Intelligence}\\[.3cm]
    
    {\large University of Technology Nuremberg}\\[1cm]

    {\large Supervisor: Prof. Dr. Wolfram Burgard}

    \vspace{1cm}
    {\large June, 2025}\\[1cm]
    
\end{titlepage}

\newpage
\thispagestyle{empty}

\begin{center}
    \textbf{\Large University of Technology Nuremberg} \\
    \vspace{0.2cm}
    \textbf{Department Computer Science \& Artificial Intelligence}
\end{center}

\vspace{0.5cm}

\begin{tabbing}
\hspace{4cm} \= \kill
\textbf{Doctoral} \> Reihaneh Mirjalili \\
\textbf{Researcher} \> \\
\\
\textbf{Submission} \> 15.06.2025 \\
\textbf{Date} \> \\
\\
\textbf{Examiner/Reviewer:} \> Prof. Dr. Tamim Asfour \\
    \> Faculty of Computer Science\\
    \> Karlsruhe Institute of Technology \\
\\
\textbf{Examiner/Reviewer:} \> Prof. Dr. Yuki Asano  \\
    \> Department Computer Science \& Artificial Intelligence\\
    \> University of Technology Nuremberg \\
\\

\textbf{Examiner:} \> Prof. Dr. Michael Roth  \\
    \> Department Computer Science \& Artificial Intelligence\\
    \> University of Technology Nuremberg \\
\\

\textbf{Advisor} \> Prof. Dr. Wolfram Burgard \\
\> Department Computer Science \& Artificial Intelligence \\
\\
\textbf{Declaration} \= 
\end{tabbing}

\vspace{-1.5em}
\begin{adjustwidth}{4.1cm}{0cm}
\justifying
I hereby declare, under oath, that this dissertation complies with the scientific standards set forth in the statutes on good scientific practice. In particular, I affirm that I have written the dissertation independently, used only the aids listed in the bibliography and the notes, and have properly indicated the sources of all passages taken literally or in substance from other works or legal rulings.

Furthermore, I declare that this dissertation has not been submitted, either in the same or a substantially similar form, at any university in the Federal Republic of Germany for the purpose of obtaining a doctoral degree.
\end{adjustwidth}

\vfill
\vspace{1.5cm}

\begin{tabbing}
\hspace{9cm} \= \hspace{9cm} \= \kill
\makebox[6cm]{\hrulefill} \> \makebox[6cm]{\hrulefill} \\
Place, Date \> Signature
\end{tabbing}

%% file: abstract/abstract_de.tex

\begin{otherlanguage}{ngerman}

\section*{Zusammenfassung}

Roboter, die in menschenzentrierten Umgebungen operieren, müssen nicht nur ihre Umgebung präzise wahrnehmen, sondern auch ihre Handlungen an vielfältige und unvorhersehbare Situationen anpassen. Traditionelle Robotikpipelines, die häufig auf aufgabenspezifische Modelle angewiesen sind, zeigen Schwierigkeiten, über ihre Trainingsverteilungen hinaus zu generalisieren. Basismodelle – großskalige, vortrainierte Modelle mit weitreichenden Fähigkeiten in den Bereichen Bildverarbeitung, Sprachverarbeitung und multimodale Verarbeitung – bieten eine skalierbare Alternative. Durch die Nutzung ihrer reichhaltigen, übertragbaren Repräsentationen können Roboter komplexe Szenen besser interpretieren, neue Aufgaben generalisieren und flexibel auf reale Situationen reagieren – ohne zusätzliche manuelle Datenerhebung oder -annotation.

Diese Dissertation untersucht, wie Basismodelle systematisch genutzt werden können, um robotische Fähigkeiten zu verbessern und eine effektivere Lokalisierung, Interaktion und Manipulation in unstrukturierten Umgebungen zu ermöglichen. Die Arbeit ist entlang von vier zentralen Fragestellungen strukturiert, die jeweils eine grundlegende Herausforderung in der Robotik betrachten und zusammen ein kohärentes Rahmenwerk für semantikbewusste robotische Intelligenz bilden.

Zunächst widmen wir uns dem Problem der robusten visuellen Ortsbestimmung – einer Schlüsselaufgabe der visuellen Lokalisierung, insbesondere unter erheblichen Veränderungen in Erscheinung, Anordnung und Perspektive. Unsere Methode erzeugt semantisch hochgradige Bilddeskriptoren, indem sie Objekte erkennt und Szenenkategorien ableitet, um Orte bedeutungsvoll zu repräsentieren. Diese Deskriptoren erfassen sowohl die Präsenz relevanter Elemente als auch den Gesamtkontext einer Szene und ermöglichen eine effektive Zuordnung zwischen Abfrage- und Referenzbildern. Durch Zero-Shot-Inferenz anstelle einer Modellanpassung bleibt der Ansatz robust gegenüber einer Vielzahl visueller Bedingungen, ohne dass zusätzliche Daten erhoben werden müssen. Experimente in realen Innenräumen zeigen starke Leistungen trotz deutlicher Szenenveränderungen und wechselnder Kameraperspektiven.

Im nächsten Schritt untersuchen wir den Einsatz von Basismodellen im Kontext des semantischen Greifens, bei dem es nicht nur darum geht, ein Objekt zu greifen, sondern dies auch in einer Weise zu tun, die mit menschlichen Präferenzen übereinstimmt und durch die funktionale Semantik des Objekts geleitet wird. Unsere Methode nutzt große Sprachmodelle (Large Language Models oder LLMs), um zu bestimmen, welche Teile eines Objekts sich sicher und sinnvoll greifen lassen, und Bild-Sprach-Modelle (Vision-Language-Models oder VLMs), um diese Stellen im visuellen Input zu lokalisieren. Dadurch kann das System Griffe erzeugen, die semantisch bedeutungsvoll, intuitiv und kontextbewusst in Bezug auf die Umgebung des Objekts sind – insbesondere in Situationen, in denen bestimmte Bereiche vermieden werden müssen. Zusätzlich führen wir einen Feedback-Mechanismus zur Machbarkeit ein, der es dem Roboter ermöglicht, seine Greifstrategie dynamisch zu bewerten und anzupassen, was die Robustheit in komplexen oder unsicheren Manipulationsszenarien erhöht.

Im dritten Teil dieser Arbeit widmen wir uns der Herausforderung der aktionsbasierten Objektklassifikation in Haushaltsumgebungen, mit besonderem Fokus auf intelligente Staubsaugerroboter. Obwohl Build-Sprach-Modelle es Robotern ermöglichen, Objekte in handlungsrelevante Klassen (z. B. „zu vermeiden“ oder „aufzusaugen“) einzuordnen – und das im Zero-Shot-Modus – sind ihre hohen Rechenanforderungen ein Hindernis für die kontinuierliche Nutzung in Echtzeit. Um diese Einschränkung zu überwinden, entwickeln wir ein Wissensdistillationsframework, das die semantischen Fähigkeiten des Build-Sprach-Modells auf ein kompaktes, effizientes Modell überträgt, das für den Einsatz auf ressourcenbeschränkten Plattformen geeignet ist. Darüber hinaus schlagen wir einen sprachgesteuerten Experience-Replay-Mechanismus vor, der kontinuierliches Lernen ermöglicht, sodass sich der Roboter im Laufe der Zeit an neue Objekte und Umgebungen anpassen kann – ohne dem Problem des katastrophalen Vergessens zu erliegen.

Abschließend greifen wir die Herausforderung auf, die Robustheit visuomotorischer Policies bei der robotischen Manipulation zu erhöhen, insbesondere bei Änderungen der visuellen Domäne, die durch Hintergrundänderungen, Unterschiede in der Roboterinkarnation oder visuelle Änderungen der Szene entstehen. Hierzu schlagen wir eine kalibrierungsfreie Methode der visuellen Abstraktion vor, die offene Vokabularsegmentierung und Objekterkennung nutzt, um aufgabenrelevante Komponenten – insbesondere den Roboter-Greifer und das Zielobjekt – zu isolieren und irrelevante visuelle Inhalte auszublenden. Die segmentierten Elemente werden anschließend auf einen konsistenten virtuellen Hintergrund projiziert. Durch die konsistente Anwendung dieser Abstraktion während des Trainings und der Inferenz kann unser Ansatz die Auswirkungen von Domänenänderungen erheblich reduzieren, was zu robusterem Verhalten des Roboters in vielfältigen Umgebungen führt – ohne dass zusätzliche Daten erforderlich sind.

\sloppy
Zusammenfassend zeigt diese Dissertation, wie Basismodelle als leistungsstarke Bausteine für semantikbewusste robotische Intelligenz dienen können. Durch ihre Integration in zentrale Aufgaben wie Lokalisierung, Greifen, Klassifikation und Manipulation demonstrieren wir, dass Roboter eine höhere Generalisierungsfähigkeit, Anpassungsfähigkeit und Kontextsensitivität in unstrukturierten Umgebungen erreichen können. Die vorgeschlagenen Methoden – basierend auf Zero-Shot-Reasoning, kompakter Modell-Distillation und visueller Abstraktion – bieten ein skalierbares Rahmenwerk für den Einsatz intelligenter robotischer Systeme in realen Anwendungsszenarien. Zusammengenommen markieren diese Beiträge einen wichtigen Schritt in Richtung robusterer, leistungsfähigerer und autonomer Roboter, die effektiv im komplexen und dynamischen Alltag des Menschen agieren können.

\end{otherlanguage}

\newpage

%% file: abstract/abstract_en.tex
\section*{Abstract}

Robots operating in human-centered environments must not only perceive their surroundings accurately but also adapt their actions to diverse and unpredictable situations. Traditional robotics pipelines, which often depend on task-specific models, struggle to generalize beyond their training distributions. Fundation models—large-scale pretrained models with broad capabilities in vision, language, and multimodal processing—offer a scalable alternative. By leveraging their rich, transferable representations, robots can better interpret complex scenes, generalize to new tasks, and respond flexibly in real-world settings, all without the need for additional data collection and labeling.

This thesis investigates how foundation models can be systematically leveraged to enhance robotic capabilities, enabling more effective localization, interaction, and manipulation in unstructured environments. The work is structured around four core lines of inquiry, each addressing a fundamental challenge in robotics while collectively contributing to a cohesive framework for semantics-aware robotic intelligence.

We first address the problem of robust visual place recognition, a key task in vision-based localization, especially under significant changes in appearance, layout, and viewpoint. Our method generates high-level semantic image descriptors by identifying objects and inferring room labels to represent locations meaningfully. These semantic descriptors are used to compute similarity scores that support reliable matching between query and reference images.
Using zero-shot inference instead of model fine-tuning, the approach remains robust across a wide range of visual conditions without the need for additional data collection. Experiments in real-world indoor environments demonstrate strong performance despite substantial scene modifications and changes in camera viewpoint.

Next, we explore the use of foundation models in semantic grasping, where the goal is not merely to grasp an object, but to do so in a manner aligned with human preferences and informed by the object’s functional semantics. Our method leverages Large Language Models (LLMs) to reason about which parts of an object afford safe and appropriate grasping, and Vision Language Models (VLMs) to localize these parts in visual input. This enables the system to generate grasps that are semantically meaningful, intuitive, and context-aware with respect to the object’s surroundings—particularly in cases where specific parts must be avoided. In addition, we introduce a feasibility feedback mechanism that enables the robot to dynamically assess and revise grasp strategies, enhancing robustness in complex or uncertain manipulation scenarios.

In the third part of this thesis, we address the challenge of action-based object classification in domestic environments, with a particular emphasis on smart robotic vacuum cleaners. Although Vision-Language Models allow robots to categorize objects into actionable classes (e.g., “to avoid” or “to suck”) in a zero-shot manner, their substantial computational demands make frequent inference impractical for real-time use. 
To address this limitation, we introduce a knowledge distillation framework. It gradually transfers the VLM’s task-relevant semantic knowledge—specifically, its ability to categorize objects into action classes—into a lightweight model suitable for deployment on resource-constrained platforms.
Additionally, we propose a language-guided experience replay mechanism to enable continual learning, allowing the robot to adapt to novel objects and environments over time while mitigating catastrophic forgetting.

Finally, we tackle the challenge of enhancing visuomotor policy robustness for robotic manipulation, particularly in the presence of visual domain shifts caused by background variations, changes in robot embodiment, or visual distractors. To address this, we propose a calibration-free visual abstraction method that leverages open-vocabulary segmentation and object detection to isolate task-relevant components—specifically, the robot gripper and target objects—while filtering out irrelevant visual content. These segmented elements are then overlaid onto a consistent virtual background.
By applying this abstraction uniformly during both training and inference, our approach significantly mitigates the effects of visual domain shift, leading to improved policy robustness across diverse environments without requiring additional data collection or retraining.

In summary, this thesis illustrates how foundation models can serve as powerful enablers of semantics-aware robotic intelligence. By integrating these models into core robotic scenarios—such as localization, grasping, classification, and manipulation—we demonstrate enhanced generalization, adaptability, and contextual reasoning in unstructured environments. By leveraging zero-shot reasoning, compact model distillation, and visual abstraction, the proposed methods provide scalable frameworks for real-world deployment. Collectively, these contributions advance the development of autonomous systems that can function reliably and intuitively alongside humans in dynamic environments.

%% file: acknowledgements.tex
\section*{Acknowledgments}

Completing this PhD has been one of the most challenging and rewarding experiences of my life. It would not have been possible without the support, encouragement, and kindness of many people.

First and foremost, I would like to express my deepest gratitude to my parents. Their unwavering support, love, and belief in me have been the foundation upon which this journey was built. Their sacrifices and encouragement have given me the strength to pursue my goals and persevere through difficult times.

I am also sincerely thankful to my friends, who have been a constant source of motivation and comfort. Your companionship, laughter, and wise words helped me stay grounded and resilient. I am deeply grateful for your presence throughout this journey.

To my labmates—thank you for creating such a collaborative and inspiring research environment. Your insights, feedback, and camaraderie have enriched my academic experience, and I have learned so much from working alongside you.

I would also like to thank Dr. Florian Walter for his guidance and support. His contributions, including conceptual input and practical assistance, were valuable to various aspects of this work. I acknowledge and appreciate the time and perspective he provided during the course of this research.

Finally, I would like to express my heartfelt gratitude to my supervisor, Prof. Dr. Wolfram Burgard. Thank you for believing in me, for giving me the opportunity to be part of your group, and for supporting me through every step of this journey. Your scientific insights and consistently great ideas led to exciting and fruitful research directions, and working under your guidance has been a truly inspiring experience. I will always remain grateful for the trust and freedom you gave me to explore and learn. Your mentorship, patience, and vision have profoundly shaped my growth as a researcher.

%% file: 1-introduction.tex
\chapter{Introduction}
\label{chap:introduction}
\section{Motivation}

\begin{quote}
\textit{“It is comparatively easy to make computers exhibit adult-level performance on intelligence tests or playing checkers, and difficult or impossible to give them the skills of a one-year-old when it comes to perception and mobility.”} \\
— Hans Moravec~\cite{moravec1988mind}
\end{quote}

For decades, the idea of bringing robots into everyday human life has captured the imagination of scientists, engineers, and the public alike—fueled by a mix of science fiction, technological ambition, and practical necessity~\citep{moravec1988mind,asimov1950robot, brooks1990elephants}.
From early attempts at building machines that could replicate human motion to industrial robots revolutionizing manufacturing in the 20th century~\citep{siciliano2016springer}, the trajectory of robotics has long hinted at a future where intelligent machines assist us in homes, hospitals, offices, and public spaces~\citep{stone2016artificial}.
The motivation behind this vision is both aspirational and pragmatic: to enhance productivity, support aging populations, improve quality of life, and take over tasks that are dangerous, repetitive, or physically demanding~\citep{bischoff2005service,murphy2009trial, pollack2002pearl}.
As computing power, sensors, and AI algorithms have advanced~\citep{lecun2015deep, krizhevsky2012imagenet, silver2016mastering}, the dream of robots seamlessly integrating into daily life has felt increasingly within reach.
Yet, despite decades of progress, general-purpose domestic and service robots remain rare~\citep{alterovitz2016robot}. The complexity of real-world environments—filled with uncertainty, variability, and subtle social cues—has consistently outpaced the capabilities of traditional robotics approaches~\citep{beetz2012cognition, thrun2005probabilistic}.

This long-standing aspiration to bring robots into everyday life has illuminated not just the potential of robotics, but also its limitations. As we move from conceptual visions and controlled demonstrations to real-world deployment, a fundamental challenge becomes clear: building systems that can operate reliably amid the complexity and unpredictability of human environments~\citep{murphy2014disaster, beetz2015knowrob}.
This transition from theory to practice brings into sharp focus a central and enduring challenge in robotics and artificial intelligence: enabling machines to perceive, reason, and act with a nuanced understanding of context, coupled with the flexibility to adapt to dynamic, unpredictable conditions~\citep{kaelbling2011hierarchical, thrun2005probabilistic}. While robots have made impressive strides in structured industrial environments—such as factory floors or controlled laboratory settings—these systems are often highly specialized~\citep{siciliano2016springer, levine2016end}.
They are typically designed around narrow tasks, trained on carefully curated datasets, and optimized under assumptions that do not hold in the real world~\citep{amodei2016concrete, hendrycks2021robustness}.
As a result, even seemingly minor variations in their operating environment—such as changes in lighting, object positions, human interactions, or scene layout—can significantly degrade their performance~\citep{recht2019imagenet, zhang2019domain, tobin2017domain}.
This fragility highlights the gap between the theoretical capabilities of robotic systems and the practical demands of real-world deployment~\citep{tobin2017domain, rusu2016sim, amodei2016concrete}.

In this context, the emergence of foundation models represents a paradigm shift in artificial intelligence~\citep{bommasani2021opportunities}. Unlike traditional models that are purpose-built for single tasks, foundation models—such as Large Language Models (LLMs) and Vision-Language Models (VLMs)—are trained on massive and diverse corpora encompassing a wide range of modalities, domains, and objectives~\cite{blip, llama, flamingo}. Their training at scale leads to striking generalization capabilities, enabling them to perform complex tasks such as abstraction, cross-modal reasoning, and zero-shot inference without requiring task-specific tuning. For instance, models like GPT-4~\cite{gpt-4o} demonstrate coherent, context-sensitive language understanding and generation, while CLIP~\cite{CLIP} aligns textual and visual representations to enable joint understanding across modalities.
Separately, generative models like diffusion models~\cite{ho2022cascaded} are capable of synthesizing realistic images from abstract prompts, but are typically trained for narrower generative objectives.
These capabilities not only push the boundaries of what is possible in AI, but they also raise a compelling question: can the generality and compositional reasoning of foundation models help close the gap between traditional robotics research and the rich variability of everyday environments?

Despite their promise, the integration of foundation models into robotics remains challenging~\citep{openx2023, roboagent}. Many current robotic systems yet continue to follow end-to-end learning paradigms, where each robot is trained independently for a particular task using domain-specific data~\citep{levine2016end, akkaya2019solving, zeng2021transporter, kalashnikov2018scalable}.
These approaches, while effective in controlled setups, often lack robustness and generalization, necessitating frequent retraining when introduced to new domains, object classes, or environmental variations~\citep{openx2023, hamalainen2019affordance}. 
Furthermore, the cost and complexity of collecting large-scale, task-specific robotic data often becomes a bottleneck~\citep{dasari2019robonet, pinto2017learning, ebert2021bridge}. 
In contrast, foundation models offer a path toward more flexible and scalable systems by providing pretrained knowledge, semantic understanding, and compositional skills that can be adapted to novel scenarios with minimal fine-tuning~\citep{gpt-4o, CLIP, llama, flamingo}.

This leads to a natural and pressing research question:

\textit{How can we systematically incorporate the strengths of foundation models into the design of robotic systems to enable reliable, context-aware behavior in open-ended, unstructured environments?}

Answering this question is critical, as real-world deployment presents a range of subtle and complex challenges for robotic systems. 
For instance, a household vacuum robot must distinguish between toys, jewelry, and crumbs scattered across the floor—deciding which objects to avoid and which to remove. Similarly, a robotic manipulator may need to grasp a delicate item like an ice cream, requiring an intuitive understanding of which part can be safely and stably held without causing damage.
These tasks extend beyond mere perception or control; they demand a deeper level of semantic reasoning, contextual awareness, and adaptability—capabilities that are difficult to achieve through narrowly focused, task-specific training alone~\citep{cliport, jang2022vima}.

To address these demands, foundation models offer a promising path forward. With their ability to encode rich semantic priors, handle multimodal inputs, and perform with minimal supervision, these models are naturally aligned with the needs of robotics in complex, dynamic settings~\citep{saycan2022, rt1x2023}. Their general-purpose design enables out-of-the-box performance across a range of tasks, reducing the need for extensive retraining. This makes them particularly attractive for scalable robotic applications, where data efficiency, robustness, and flexibility are critical.

Motivated by both the challenges of real-world deployment and the capabilities of foundation models, this thesis explores how such models can be systematically integrated into robotic systems. The goal is to enhance semantic understanding, improve adaptability, and enable greater generalization in unstructured environments, ultimately bringing robotic behavior closer to the flexible reasoning seen in natural agents.

\section{Scope and Vision}

This work explores how foundation models can enhance the robustness, adaptability, and semantic understanding of robotic systems. Rather than engineering narrowly optimized solutions for isolated tasks, it aims to leverage the broad generalization capabilities of these models to create robots that naturally adapt to the variability of real-world environments. When carefully integrated, foundation models offer the potential to enable semantically aware systems that generalize across tasks and contexts.

Our focus lies on high-level semantic integration in robotic systems, with an emphasis on using foundation models to support robust and adaptable behavior in open-ended environments. Rather than addressing low-level control or mechanical design, this work targets scenarios where contextual understanding and abstraction are essential—such as visual place recognition, semantic grasping, autonomous vacuuming, and visually-robust manipulation.
To address these challenges, this thesis develops robotic system architectures that embed foundation models—particularly large language models and vision-language models—as core reasoning components. These models are integrated into the robotic decision-making loop, enabling richer semantic understanding. The goal is to design systems that leverage foundation models to tackle the challenges of real-world operation, including real-time execution, limited supervision, and visual variations.

By building on the rich multimodal representations learned by foundation models, this work aims to support robots that can generalize across tasks and environments. For example, identifying cup handles as graspable features, or understanding that certain objects—like toys on the floor—should be avoided during cleaning tasks. These forms of inference are difficult to encode explicitly, but can emerge naturally from the structured priors encoded in foundation models.

The systems developed in this thesis are evaluated under realistic deployment conditions, with a focus on semantic abstraction, multimodal inference, and zero-shot generalization. Beyond the immediate technical contributions, the broader vision is to help shift robotics away from rigid, task-specific pipelines toward systems grounded in high-level semantic understanding. By leveraging foundation models, this work aims to develop agents that are not only more flexible and scalable, but also trustworthy and adaptable—key attributes for effective operation in everyday human environments.

\section{Research Questions}

As mentioned earlier, the central hypothesis of this thesis is that foundation models can be systematically integrated into robotic systems to enable robust, semantically informed behaviors across diverse and unstructured environments. To investigate this hypothesis, the following research questions are addressed, each corresponding to a core dimension of the thesis:

\begin{enumerate}
    \item \textbf{How can foundation models be integrated into robotics to solve key challenges such as localization, grasping, and manipulation?}

    This question explores strategies for integrating foundation models—such as Segment Anything~\cite{sam2}, \mbox{CLIP~\cite{CLIP}} and GPT~\cite{gpt-4o}—into robotic pipelines. The emphasis is on identifying the roles these models can play and how their capabilities can complement or replace traditional, narrowly trained components. Each system in this thesis reflects a different integration philosophy, ranging from generating text description for images to assisting with scene abstraction.

    \item \textbf{Can we develop robotic systems that generalize across domains without retraining or manual data collection?}

    One of the key promises of foundation models is their ability to perform zero-shot or few-shot generalization across a range of contexts. This question investigates whether such generalization holds in embodied, interactive settings—where visual diversity, physical interaction, and environmental noise present unique challenges. By evaluating systems in real-world environments with minimal task-specific fine-tuning, this thesis assesses the practicality of deploying robotic systems that can leverage foundation models to generalize across domains.

    \item \textbf{What is the role of multimodal reasoning in robotic decision-making, and how can we use it effectively?}

    Real-world robotic tasks often require integrating information from multiple modalities—such as language and vision—to make informed, context-sensitive decisions. For instance, a command like “pick up the blue cube that is farther from the yellow cube” cannot be executed using language or visual input in isolation. It requires interpreting spatial relationships expressed in language and accurately grounding them in the visual scene through spatial reasoning and object detection.
    This research question investigates how foundation models can support such multimodal reasoning by aligning language-driven goals with perceptual data. 

    \item \textbf{How can we use foundation models efficiently in robotics, given the constraints of real-time inference and limited hardware?}

    Foundation models are typically computationally intensive, making them unsuitable for direct deployment on embedded or mobile platforms. To address this challenge, knowledge distillation can be employed to transfer the semantic capabilities of large models into smaller, more efficient student models. The objective is to develop lightweight models that retain the essential knowledge of the foundation model while being capable of real-time inference in dynamic environments. Furthermore, these models should support continual learning, enabling them to learn new information over time without experiencing catastrophic forgetting of previously acquired knowledge.
\end{enumerate}

These questions collectively shape the methodological and experimental directions of this thesis. Each chapter addresses a specific challenge while contributing to the main goal of building robotic systems that are more general, reliable, and semantically capable in the face of real-world complexity.

\section{Contributions and Thesis Overview}

This thesis presents four original contributions that demonstrate how foundation models can enhance the semantic reasoning, generalization, and robustness of robotic systems. Each contribution corresponds to a core robotics task and is presented in a dedicated chapter, supported by experimental evaluations.

\begin{itemize}
    \item \textbf{Chapter~\ref{chapter:fm-loc}: FM-Loc — Semantic Place Recognition.} \\ This chapter introduces FM-Loc, a localization pipeline that constructs semantic descriptors by combining object detection with language-based room inference using both large language models and vision language models. Unlike traditional methods that rely on geometric priors or dense visual features, FM-Loc achieves robust place recognition across large viewpoint and appearance changes without the need for data collection or fine-tuning. The chapter includes evaluations in indoor environments with significant visual variations. This work has been published as \textit{``FM-Loc: Using Foundation Models for Improved Vision-Based Localization''} by Reihaneh~Mirjalili, Michael~Krawez, and Wolfram~Burgard, in \textit{IROS~2023}~\cite{fm-loc}. The \href{https://github.com/rmirjalili/FM-Loc}{\textcolor{blue}{\underline{code}}} and 
\href{https://huggingface.co/datasets/rmirjalili/fm-loc}{\textcolor{blue}{\underline{data}}} 
are available online.

    \item \textbf{Chapter~\ref{chapter:langrasp}: Lan-grasp — Language-Guided Semantic Grasping.} \\ Lan-grasp introduces a novel semantic grasping system that leverages large language models to interpret the appropriate part of an object to grasp, and vision-language models to localize that target region. A visual feedback loop enables the robot to iteratively refine its grasp strategy when the initially proposed grasp is not feasible. This chapter presents results from real-world grasping experiments, demonstrating how foundation models can guide more meaningful and context-aware grasping behaviors. This work has been accepted for publication in \textit{ISRR~2024} and is currently available as an arXiv preprint titled \textit{``Lan-grasp: Using Large Language Models for Semantic Object Grasping''} by Reihaneh~Mirjalili, Michael~Krawez, Simone~Silenzi, Yannik~Blei, and Wolfram~Burgard~\cite{Lan-grasp}.

    \item \textbf{Chapter~\ref{chapter:vacuum}: VLM-Vac — Autonomous Vacuuming via Knowledge Distillation.} \\ This chapter introduces VLM-Vac, an intelligent vacuuming framework that uses vision-language models to semantically interpret images and avoid items that are not considered dirt, such as toys, jewelry, or socks. A light-weight student model learns from the VLM and overtime, enables fast inference, supporting autonomous vacuuming while reducing VLM queries. Additionally, we employ a language-guided experience replay strategy to mitigate catastrophic forgetting during continual learning. Experimental results demonstrate improvements in energy efficiency, adaptability to new objects and flooring patterns, and robust action-based object classification. This work has been accepted for publication as \textit{``VLM-Vac: Enhancing Smart Vacuums Through VLM Knowledge Distillation and Language-Guided Experience Replay''} by Reihaneh~Mirjalili, Michael~Krawez, Florian~Walter, and Wolfram~Burgard, in \textit{ICRA~2025}~\cite{vlmvac}. The \href{https://github.com/rmirjalili/VLM-Vac}{\textcolor{blue}{\underline{code}}} and 
\href{https://huggingface.co/datasets/rmirjalili/VLM-Vac-dataset}{\textcolor{blue}{\underline{data}}} 
are available online.

    \item \textbf{Chapter~\ref{chapter:arro}: ARRO — Visual Abstraction for Robust Manipulation.} \\
    ARRO (Augmented Reality for RObots) introduces a preprocessing framework that leverages open-vocabulary segmentation and object detection to suppress irrelevant visual content while preserving only regions essential for task execution.
    This abstraction facilitates the training and deployment of visuomotor policies across diverse visual conditions, including variations in background and robot appearance. This chapter evaluates ARRO’s effectiveness in manipulation tasks under visual domain shifts, demonstrating its robustness to changes in the environment. This work is currently under review and is available as an arXiv preprint titled \textit{``Augmented Reality for RObots (ARRO): Pointing Visuomotor Policies Towards Visual Robustness''} by Reihaneh~Mirjalili and Tobias~Jülg (equal contribution), Florian~Walter, and Wolfram~Burgard~\cite{mirjalili2025augmented}.

    \vspace{1cm}

    \item \textbf{Chapter~\ref{chapter:conclusion}: Conclusion.} \\
    The final chapter summarizes the thesis findings, synthesizes the gained insights, and outlines limitations and future directions. 
\end{itemize}

Together, these chapters advance a unified perspective: that foundation models, when thoughtfully integrated into robotic systems, can serve as powerful tools for semantic reasoning, enabling flexible and generalizable behavior. By leveraging the vision-language representations encoded in foundation models, this thesis shows how robots can move beyond narrow, task-specific execution toward more adaptable and context-aware autonomy in real-world environments.

\clearpage
\subsection*{Author Contribution Statement}

As previously stated, this thesis is composed of four core chapters, each based on a research publication. While all projects involved joint efforts, the following paragraphs outlines my specific contributions to each work to ensure transparency and academic integrity.

In \textbf{Chapter~\ref{chapter:fm-loc} (FM-Loc)}, the high-level research direction was proposed by my supervisor, Prof.~Dr.~Wolfram~Burgard. I developed the core methodology, including the design of the semantic localization pipeline and the integration of foundation models for object and scene-level reasoning. I developed the similarity scoring mechanism, gathered the datasets, and performed the full evaluation and analysis. Data collection and data pre-processing were carried out using infrastructure developed by Michael~Krawez. He also provided the baseline methods and created the trajectory plotting tools; I used these tools to do all the evaluations and compare results. The paper was co-written by me and Michael~Krawez, with substantial contributions from both sides. Feedback from all co-authors was incorporated during the revision process. Prof.~Dr.~Wolfram~Burgard provided technical advice and assisted with writing the manuscript.

In \textbf{Chapter~\ref{chapter:langrasp} (Lan-grasp)}, the high-level research idea was proposed by my supervisor. I designed the core pipeline of the method, including the integration of a Large Language Model and a Vision-Language Model for identifying graspable object parts and incorporating this information into the grasp planning process. I also proposed and realized the Visual Chain-of-Thought feedback mechanism for evaluating grasp feasibility. I conducted all experiments and evaluations both for our method and the baselines, using the experimental infrastructure developed by my co-authors. The ablation studies were carried out jointly by me and Michael~Krawez, and the paper was co-written by us both. Feedback from all co-authors was incorporated during the revision process. Prof.~Dr.~Wolfram~Burgard provided general consultation and helped in writing the manuscript.

In \textbf{Chapter~\ref{chapter:vacuum} (VLM-Vac)}, the core research idea was proposed by my supervisor. I designed and implemented the entire method, including the integration of the Vision-Language Model for action-based object detection, language-guided experience replay, and the knowledge distillation framework. I conducted all experiments, performed the evaluations, and created all visualizations. The manuscript was written primarily by me, with input and revisions from co-authors. I handled all other aspects of the research. Prof.~Dr.~Wolfram~Burgard provided continuous technical supervision and contributed feedback on the manuscript.

\textbf{Chapter~\ref{chapter:arro} (ARRO)} reflects shared first authorship with Tobias~Jülg. This work was based on an idea I proposed. I defined the research question, developed the core solution, and designed the full ARRO pipeline. The robot and dataset collection infrastructure was developed by my coauthor Tobias~Jülg. I jointly developed the diffusion policy approach (ARRO, masked and vanilla variants) with Florian~Walter, and the data collection was provided by Tobias~Jülg. I conducted the experimental evaluation of three variants of the diffusion policy in real-world scenarios under visual domain shift and distractor objects. Tobias~Jülg developed real-to-sim, cross-embodiment, and generalist VLA policy integration and applied ARRO on them. The manuscript reflects shared first authorship with substantial contributions from all coauthors. Prof.~Dr.~Wolfram~Burgard provided general consultation and helped in writing the manuscript.

A signed version of the contribution statements is maintained separately and available upon request.

%% file: publications/fmloc.tex
\chapter{FM-Loc: Using Foundation Models for Improved Vision-based Localization}
\chaptermark{FM-Loc: Vision-based Localization}
\label{chapter:fm-loc}

The work presented in this chapter has been published in~\cite{fm-loc}:

\vspace{1cm}%
\hspace*{1cm}%
\begin{minipage}{.9\textwidth}%
R. Mirjalili, M. Krawez, and W. Burgard. 

\textbf{FM-Loc: Using Foundation Models for Improved Vision-Based Localization}

\textit{IEEE International Conference on Intelligent Robots and Systems (IROS), 2023.}

DOI: \url{https://doi.org/10.1109/IROS55552.2023.10342439}

\end{minipage}%

\vspace{1cm}%

\newpage

\section*{Abstract}
\begin{adjustwidth}{1.2cm}{1.2cm} 
\small 
\textbf{
Visual place recognition is essential for vision-based robot localization and SLAM. Despite the tremendous progress made in recent years, place recognition in changing environments remains challenging. A promising approach to cope with appearance variations is to leverage high-level semantic features like objects or place categories. In this chapter, we propose FM-Loc which is a novel image-based localization approach based on foundation models that uses the Large Language Model GPT-3 in combination with the Visual-Language Model CLIP to construct a semantic image descriptor that is robust to severe changes in scene geometry and camera viewpoint. We deploy CLIP to detect objects in an image, GPT-3 to suggest potential room labels based on the detected objects, and CLIP again to propose the most likely location label. The object labels and the scene label constitute an image descriptor that we use to calculate a similarity score between the query and database images. We validate our approach on real-world data that exhibit significant changes in camera viewpoints and object placement between the database and query trajectories. The experimental results demonstrate that our method is applicable to a wide range of indoor scenarios without the need for training or fine-tuning.}
\end{adjustwidth}

\section{Introduction}
Robust place recognition is of utmost importance for robot navigation as it supports highly relevant tasks including localization and SLAM. To efficiently perform navigation tasks and to build consistent maps of their environment, robots need the ability to determine their location given a map or a set of previously recorded observations. In vision-based place recognition, one relies on camera images and typically matches a query image to a set of reference or previously recorded images, which is akin to the problem of image retrieval. A popular approach to vision-based place recognition is to use local image features, While such feature-based methods provide good results in many applications, they typically are less robust when the scene appearance undergoes substantial changes.  In indoor environments, such changes can come either from lighting variations or from the addition, removal, or rearrangement of objects within the scene. Further problems for feature-based methods are caused by substantial viewpoint changes between the individual observations. A recent trend to tackle vision-based place recognition under such dynamic changes is to incorporate high-level semantic information, e.g., through the introduction of objects~\cite{bowman2017probabilistic, yu2018ds, gomez2020object}. 

\begin{figure}[t]
\vspace{1ex}
\centering
\includegraphics[clip,trim=2cm 0cm 1cm 0cm,scale=0.4]{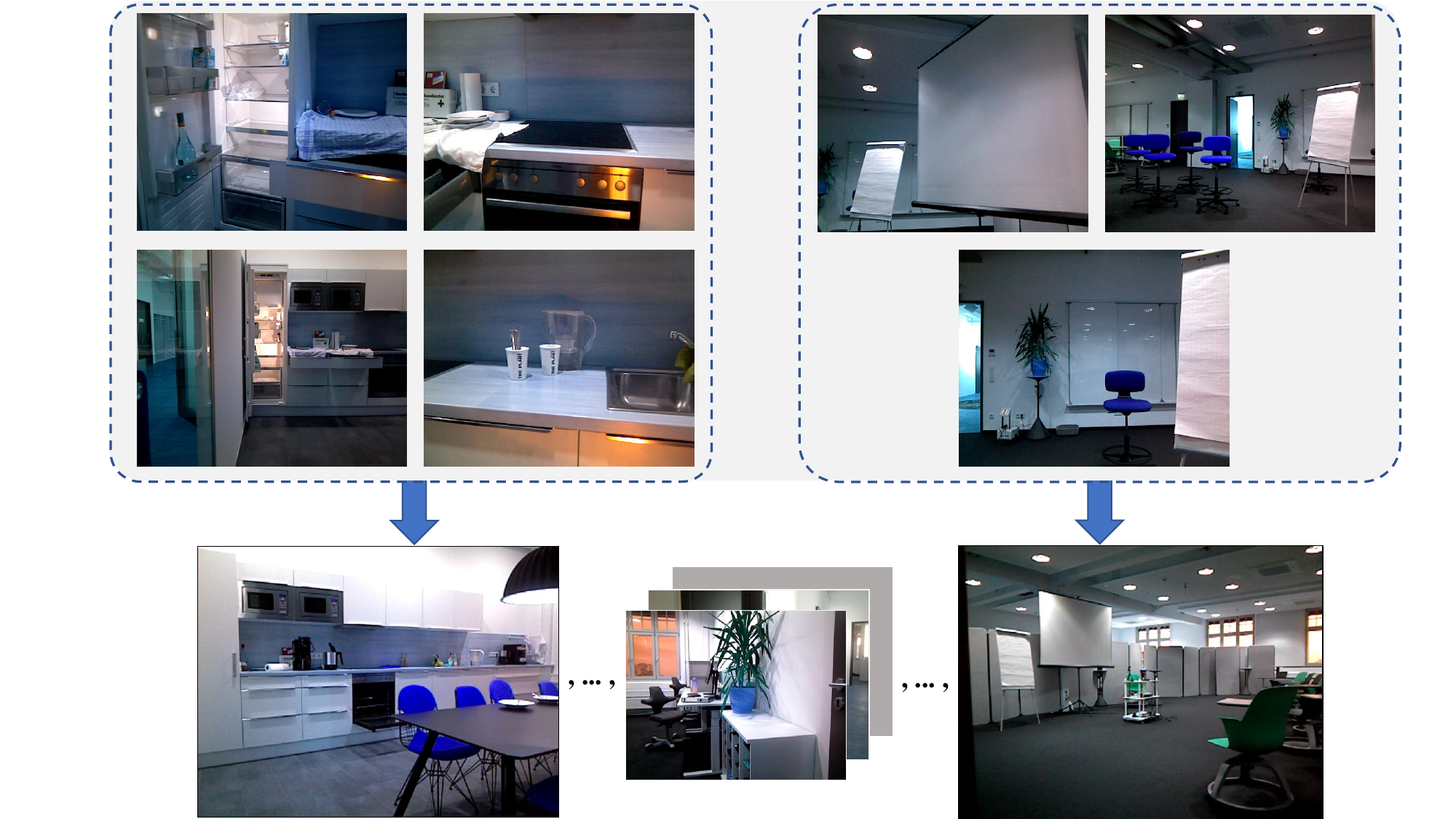}  
\caption[Visual localization under viewpoint and object placement changes]{Our localization method matches query images (top) to a set of reference images (bottom). Despite considerable differences in viewpoint and object placement between reference and query image sets, our approach correctly recognizes the locations.}
\label{fig:covergirl}
\end{figure}

In this chapter, we propose FM-Loc which is a novel method based on foundation models for vision-based localization in changing indoor environments. We leverage Large Language Models (LLMs) and Visual-Language Models (VLMs) to derive a semantic image descriptor that is robust to extreme scene rearrangements and viewpoint changes. Our method uses the VLM CLIP to detect objects in query and reference images and the LLM GPT-3 to classify the location based on the extracted object labels. We then use the object and location labels to form a semantic image descriptor that is robust even to extreme scene rearrangements and viewpoint changes, as shown in \autoref{fig:covergirl}. 

Typically, LLMs like BERT~\cite{devlin2018bert} or GPT-3~\cite{brown2020language} and VLMs like CLIP~\cite{CLIP} are referred to as foundation models~\cite{bommasani2021opportunities} since their scale and the generality of the training data enables downstream applications to a wide field of tasks. In conjunction, LLMs and VLMs allow for the grounding of natural language commands or descriptions in the real world, which is an important precondition for many robotics tasks including manipulation and human-robot interaction. While foundation models were already applied successfully in the field of robotics to perform navigation, object detection, and manipulation tasks, their utilization for SLAM and localization has received less attention. In the  localization method proposed in this chapter, we employ foundation models to ground object descriptions in images and to classify the depicted locations.

In particular, we deploy CLIP to match an image and a list of object labels and retrieve the objects with the highest matching score. Using the object labels as the sole image descriptor has several caveats. Static and unique objects are good global landmarks, however, other objects might be specific to a room but not static, and yet other objects do not provide any useful information about the location. For example, in an office building chairs can be found at working places, in meeting rooms, or in a kitchen. A set of dynamic objects like cups, plates, and cutlery can indicate a specific room, e.g., a kitchen, but not be present in the same configuration in query and reference images. To robustify the descriptor, we aggregate over the object labels by prompting GPT-3 to provide likely room types that contain these objects. We again use CLIP to ground this candidate list in the image and to select the final room label. The room label and the labels of detected landmark objects then constitute the image descriptor.  

The fact that our approach relies on foundation models makes it applicable to a wide range of environments without the need to re-train or fine-tune the underlying models. Further, LLMs and VLMs permit open-vocabulary lists of object and room labels which can be either drawn from existing taxonomies or defined manually. Thus, labels of descriptive landmarks specific to the target environment can be easily added to the list by the user and our approach does not require any pretrained classifier to perform the place recognition task. 

\textbf{In summary, this chapter includes the following contributions : }
\begin{enumerate}
    \item We present a novel approach based on foundation models to improve vision-based localization.
    \item We demonstrate that our  approach  can easily be extended to novel landmarks without additional training.
    \item We provide an approach to extract the landmarks that are most relevant for the localization task.
\end{enumerate}
 We furthermore present real-world experiments in dynamic environments to  demonstrate these capabilities.


\section{Related Work}
\label{sec:related_work}
Large-scale language models have recently become more common in the field of robotics and recent works demonstrate their potential to simplify robotic problems like navigation or human-robot interaction. The
LM-Nav system~\cite{shah2022robotic} leverages pretrained models to navigate a robot in a topological map following natural language instructions. The approach first extracts a sequence of landmark labels from the instruction using GPT-3. It then uses CLIP to match  landmark labels and map images to determine likely landmark locations. They employ a graph search algorithm to find a path that visits the nodes containing the requested landmarks. Finally, they employ ViNG~\cite{shah2021ving} to navigate the robot between the graph nodes. 
Huang \etal~\cite{huang2022visual} also tackle the problem of following instructions in robot navigation. They use a grid map annotated with CLIP features extracted from input images. The metric map representation enables the user to include spatial descriptions of the goal location in the instructions.
CLIP-Nav~\cite{dorbala2022clip} uses GPT-3 to break down complex navigation instructions into command keyphrases. It selects the next intermediate goal  using CLIP to ground a keyphrase in the panoramic image of the current location.
Gadre \etal~\cite{gadre2022clip} utilize CLIP for zero-shot, open-vocabulary object detection. Given a target object label, the robot explores its vicinity and generates a top-down map from RGB-D images. The algorithm monitors the CLIP similarity score between the current image and the object label and ends the search if that score is above a certain threshold.

LLMs were further applied in robotic manipulation. Brohan \etal~\cite{brohan2022can} present a scheme for grounding language commands in robot actions. In particular, they use RL to learn an affordance function  that estimates for a set of actions their execution success in a given context. To complete a task at hand, the LLM computes a score for the next best action which is then combined with the affordance function to ensure the  feasibility of the action. Chen \etal~\cite{chen2022open} extend the latter work to scene-level tasks with open vocabulary.

The approaches discussed so far mostly use the notion of objects to ground natural language instructions in the physical world, but do not explicitly address the problem of robot localization.
However, the objects detected by a robot can provide valuable clues about its location. For instance, observing a landmark that is unique in the environment strongly reduces the pose hypothesis space. Yet other, non-unique objects might be specific to a certain room type. Thus, several works use semantic object labels and a language-based classifier to leverage localization or scene recognition. 

Heikel and Espinosa-Leal~\cite{heikel2022indoor} deploy YOLO~\cite{redmon2016you} to retrieve object labels from an input image. They transform these labels into language-based features and employ a random forest classifier to predict the room label. Similarly, Chen \etal~\cite{chen2019scene} obtain object labels from YOLO and use a learned taxonomy to refine the results of a CNN-based scene classifier. Our approach follows the same strategy of deriving the room class from object labels. In contrast to the approaches mentioned above, we employ foundation models for both, object detection and scene classification. This allows us to perform both tasks in a zero-shot fashion and further grants greater flexibility on landmark and environment labels.

Appearance-based localization is closely related to image retrieval and place recognition. Most state-of-the-art approaches in these fields are able to match query images to a reference database if the query and reference images were captured under similar conditions. However, making image retrieval robust to drastic viewpoint and appearance changes is still an ongoing research problem.
A line of work by Vysotska \etal~\cite{vysotska2015efficient, vysotska2016lazy, vysotska2017relocalization} explores image sequence matching under seasonal or day-and-night appearance variations. Tomit{\u{a}} \etal~\cite{tomitua2021convsequential} also exploit sequential information to increase the robustness of place recognition. In single-image retrieval, data-driven approaches dominate the field.
NetVLAD~\cite{arandjelovic2016netvlad} is an end-to-end CNN model that is the de facto standard for place recognition. Patch NetVLAD~\cite{patchnetvlad} improves the results of the basic NetVLAD method by local feature matching. That approach displays significant robustness to viewpoint and appearance variations, although it can still fail under drastic changes. Garg, Suenderhauf, and Milford~\cite{garg2022semantic} show that incorporating semantics can improve opposing view matching in cases of changing scene appearance. Similar in spirit, we propose an approach to combine the Patch NetVLAD score with our language-based semantic descriptor and demonstrate that the matching performance is improved under extreme appearance and viewpoint variations.

\begin{figure*}
\centering
\includegraphics[clip,trim=0 1cm 0 1cm, width=\textwidth]{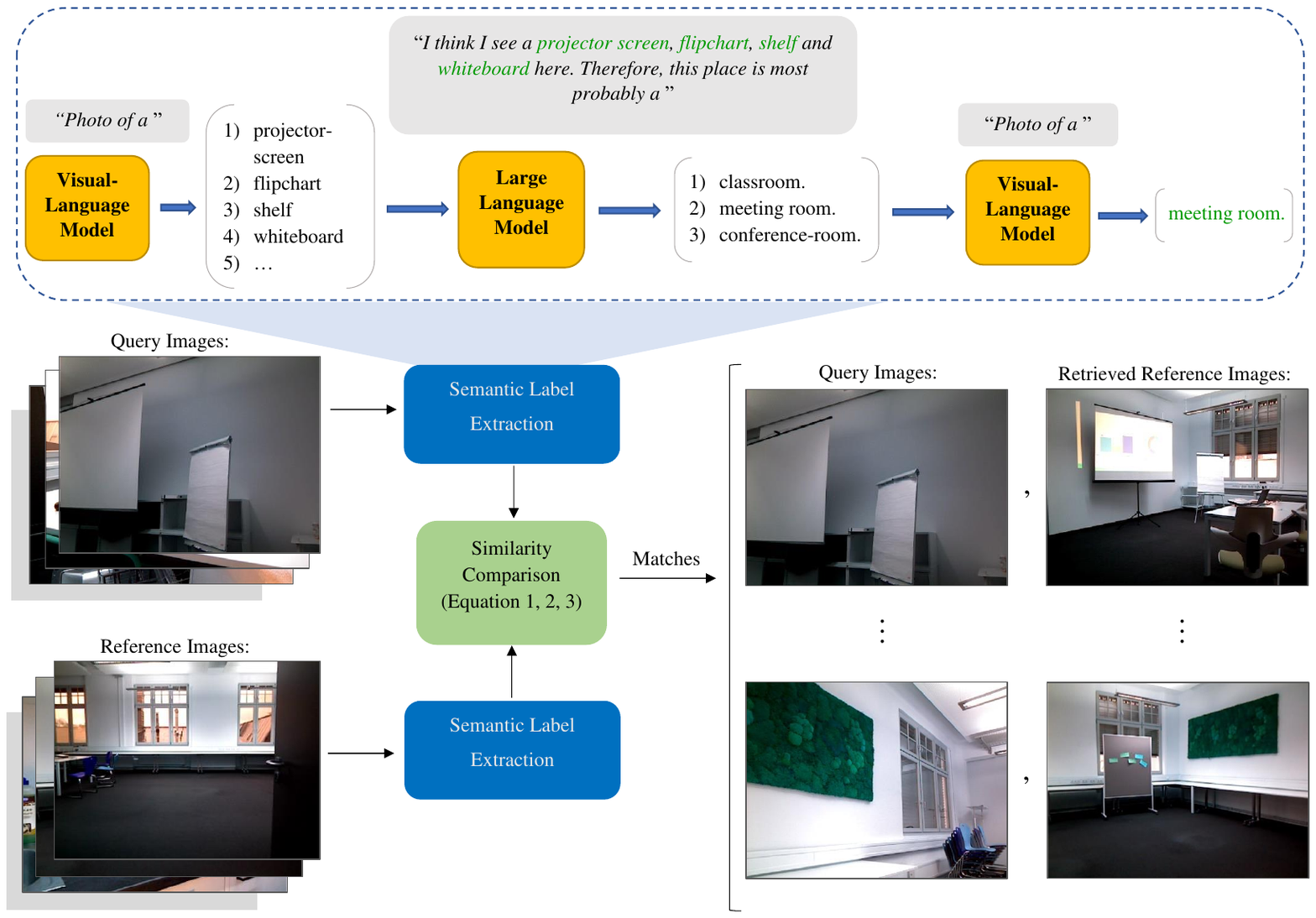}
\caption[FM-Loc: Overview of our localization method]{FM-Loc in a nutshell: We use a Visual-Language Model (VLM) and a Large Language Model (LLM) to achieve robust localization under severe viewpoint and scene changes. For each query image, the VLM detects objects and ranks them by grounding score. The top-k object labels form a prompt for the LLM, which predicts potential room categories. These room labels are re-ranked by the VLM to identify the most likely location. We use this semantic information to compare each query image with reference images and select the one with the highest similarity score.}
\label{fig:fmoverview}
\end{figure*}

\section{Approach}
\label{sec:approach}
In this section, we formally describe our localization approach. The  map of the robot corresponds to the sets of reference images $R$ and the corresponding camera poses $P_R$. The query image set $Q$ depicts the same location at a different time point whereby the poses $P_Q$ are unknown. Our method targets indoor environments which can be clearly separated into distinct rooms or areas, e.g., \textit{``hallway''}, \textit{``meeting room''}, or \textit{``kitchen''}. The environment can change between the recordings of $R$ and $Q$ in terms of illumination or object dynamics, and the robot can move along different trajectories. However, we assume that all rooms the robot visits in $Q$ are also present in the reference trajectory. With that, our goal is for each $I_i \in Q$ to find the best match $I_j \in R$ so that $I_i$ and $I_j$ show the same room and the pose difference between $p_i \in P_Q$ and $p_j \in P_R$ is minimal. Our approach is depicted in \autoref{fig:fmoverview}. 


\subsection{Language-Based Descriptor}
\label{sec:language_descriptor}
We compute the semantic image descriptor in three steps: First, we utilize CLIP to detect the objects contained in the image. Based on the object labels, GPT-3 generates several proposals for room or area labels. Finally, we again use CLIP to ground the room or area labels in the image and select the one with the highest score. 

CLIP does not directly map an image to a set of object categories but performs an image-caption pairing task. It converts the image and a natural language prompt to a common embedded space where both representations are compared. For an image $I_i$ and a label $l$, $s_i(l)$ denotes the comparison score between $I_i$ and the prompt \emph{``a photo of a [$l$]''}. To detect objects with CLIP, we, therefore, need a pre-defined list $\mathcal{L}^{obj}$ of object labels. Due to the large-scale nature of the VLM, there are few restrictions on what labels can be included in that list. We take the object class labels from the MS COCO dataset~\cite{cocodataset} as a foundation that covers objects commonly found in indoor spaces. Additionally, environment-specific object categories can be suggested by the user. For each $l \in \mathcal{L}^{obj}$, we compute $s_i(l)$ and consider the five best labels as the set of detected objects $L_i^{obj}$.

In the next step, we deploy GPT-3 for classifying the scene. To that end, we plug the labels $L_i^{obj}$ into the prompt \emph{``I think I see a $[l_1,\ldots, l_5]$ here. Therefore, this place is most probably a ''} and retrieve multiple answers from the LLM. We finally select the best room label candidate $l_i^{room}$ according to the CLIP score, whereby the prompt has the same shape as for object grounding. The object labels $L_i^{obj}$ and room label $l_i^{room}$ constitute the image descriptor.

There are two reasons for the multi-step place classification described above. First, automating the room label generation by GPT-3 removes the need to pre-define the labels beforehand. Given the large training set of the LLM, this approach can provide more specific room classes (e.g., a furniture store) in addition to common ones (e.g., office, kitchen, or bedroom). This provides greater flexibility to our method and enables localization in a broad set of environments. Second, following the idea of Zheng \etal~\cite{socratic}, we use the exchange between two foundation models, trained on different datasets, to achieve a more robust room categorization.

\subsection{Similarity Comparison}
\label{sec:semantic_similarity}
We use both, the detected object categories and room labels, to compare a query image $I_i$ and the reference image $I_j$. Room labels provide a high-level yet robust estimation of the robot's location. However, most robotic applications require more accurate localization. Therefore, it is essential to use the object information in addition to the room labels to provide the algorithm with a means to localize \emph{inside} the rooms. Thus, we define the semantic similarity score $S_{i,j}^{sem}$ as the sum of object and room similarity terms:
\begin{equation}
\label{eq:sim_lang}
S_{i,j}^{\mathit{sem}} =  S_{i,j}^{\mathit{obj}} + S_{i,j}^{\mathit{\mathit{\mathit{room}}}}
\end{equation}
To calculate $S_{i,j}^{\mathit{obj}}$, we first compute the set $L_{i,j}^{\mathit{obj}}$ of object labels appearing in both images. Thus, the object similarity score is
\begin{equation}
\label{eq:sim_obj}
    S_{i,j}^{\mathit{obj}} = \sum_{l \in L_{i,j}^{\mathit{obj}}} \frac{\min(s_i(l),s_j(l))}{\max(s_i(l),s_j(l))}.
\end{equation}
For a single label $l$, the fraction term is maximized if the magnitude of $s_i(l)$ and $s_j(l)$ are similar. We choose the above formulation because the magnitude of the CLIP detection score depends on how large the object appears in the image. Therefore, $S_{i,j}^{obj}$ is higher for image pairs showing the same set of objects, where the objects have a similar distance to the camera. This helps retrieve a reference image that has a viewpoint similar to that of the query image.

The room similarity score is given by
\begin{equation}
\label{eq:sim_room}
    S_{i,j}^{\mathit{room}} = f(e(l_i^{\mathit{room}})\cdot e(l_j^{\mathit{room}})).
\end{equation}
Here, $e(l_i^{room})$ and $e(l_j^{room})$ are the normalized GPT-3 embedding vectors for the two room labels that are compared using the dot product. We do not compare the label strings directly since they are generated by the LLM and prone to noise. Often, GPT-3 finds synonymous labels for the same location, e.g., \emph{``corridor''} and \emph{``hallway''}, which, however, have a similar embedded representation. The function $f(\cdot)$ is
\begin{equation}
    f(x) = 
    \begin{cases}
    \frac{x-\theta}{1-\theta} &  x>\theta \\
    0 & \text{else}
    \end{cases}.
\end{equation}
It sets $S_{i,j}^{room}$ to zero for embeddings with a similarity below a certain threshold (we use $\theta=0.75$) and re-normalizes other values to the range $[0,1]$.

\section{Experimental Evaluation}
\label{sec:results}
In this section, we present the experimental evaluation of the proposed method. We apply our approach to real-world data that exhibits substantial environmental changes and compare the results to state-of-the-art appearance-based localization methods. The experiments are designed to demonstrate that our approach 
\begin{enumerate}
    \item can be used to improve the robustness and accuracy of vision-based localization, 
\item can easily be  extended to incorporate novel objects without the need for additional training of models, and 
\item can be utilized to identify landmarks that are relevant to the localization task.
\end{enumerate}

\begin{figure}[t]
\vspace{1ex}
\centering
\includegraphics[clip, scale=0.5]{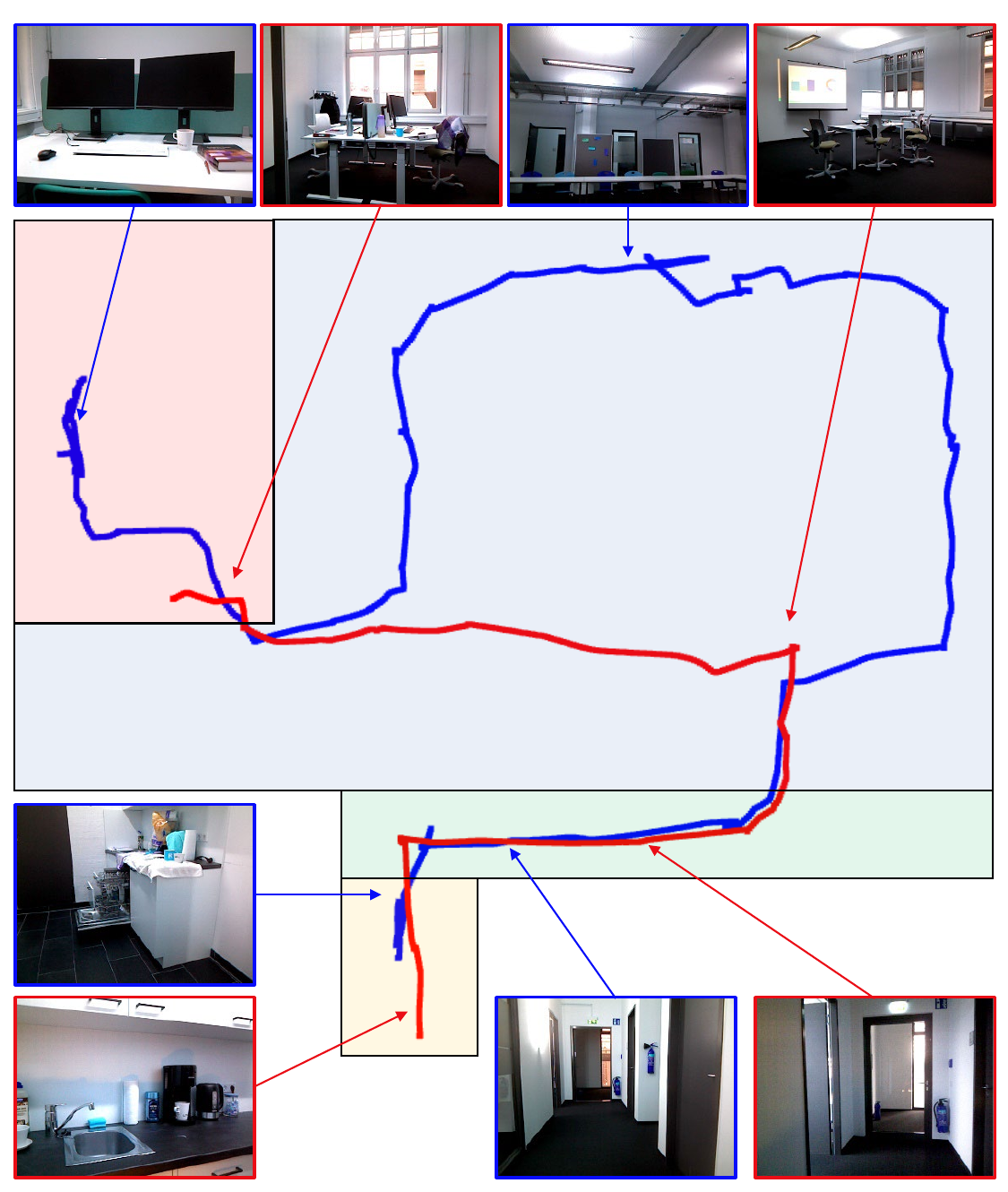}  
\caption[Query and reference trajectories across four room types]{The reference (red) and query (blue) trajectories in dataset~1. The environment consists of four rooms, namely a kitchen (yellow), a hallway (green), a conference room (blue), and an office (red). The images depict these rooms for query and reference trajectories.}
\label{fig:ds1}
\end{figure}

\subsection{Datasets}
For evaluation, we captured two datasets each containing a reference and a query image set in two different floors of an office building. The first dataset, shown in \autoref{fig:ds1}, contains $111$ images for query and $226$ for reference while the second dataset consists of 101 images for both query and reference sets. All trajectories cover four different locations, namely a kitchen, a hallway, a large conference room, and an office. The query trajectories exhibit changes in object and furniture arrangement and have different lighting conditions. Further, the query trajectories deviate from the reference trajectories in terms of position and camera viewpoints. We used the Toyota HSR robot for recording the trajectories and the images. We aligned the two trajectories by picking image pairs from the query and reference sets that contain the same static scene from a similar view. Next, we computed the camera transformation between the images using ORB features and RANSAC. We then used the aligned trajectories for calculating the translation errors of the considered approaches. For calculating the ratio of the correct room detections, we generated the ground truth room labels by manual annotation.

\begin{figure}[htbp]
\centering
\includegraphics[clip,trim=0cm 4cm 7cm 0cm, width=\textwidth]{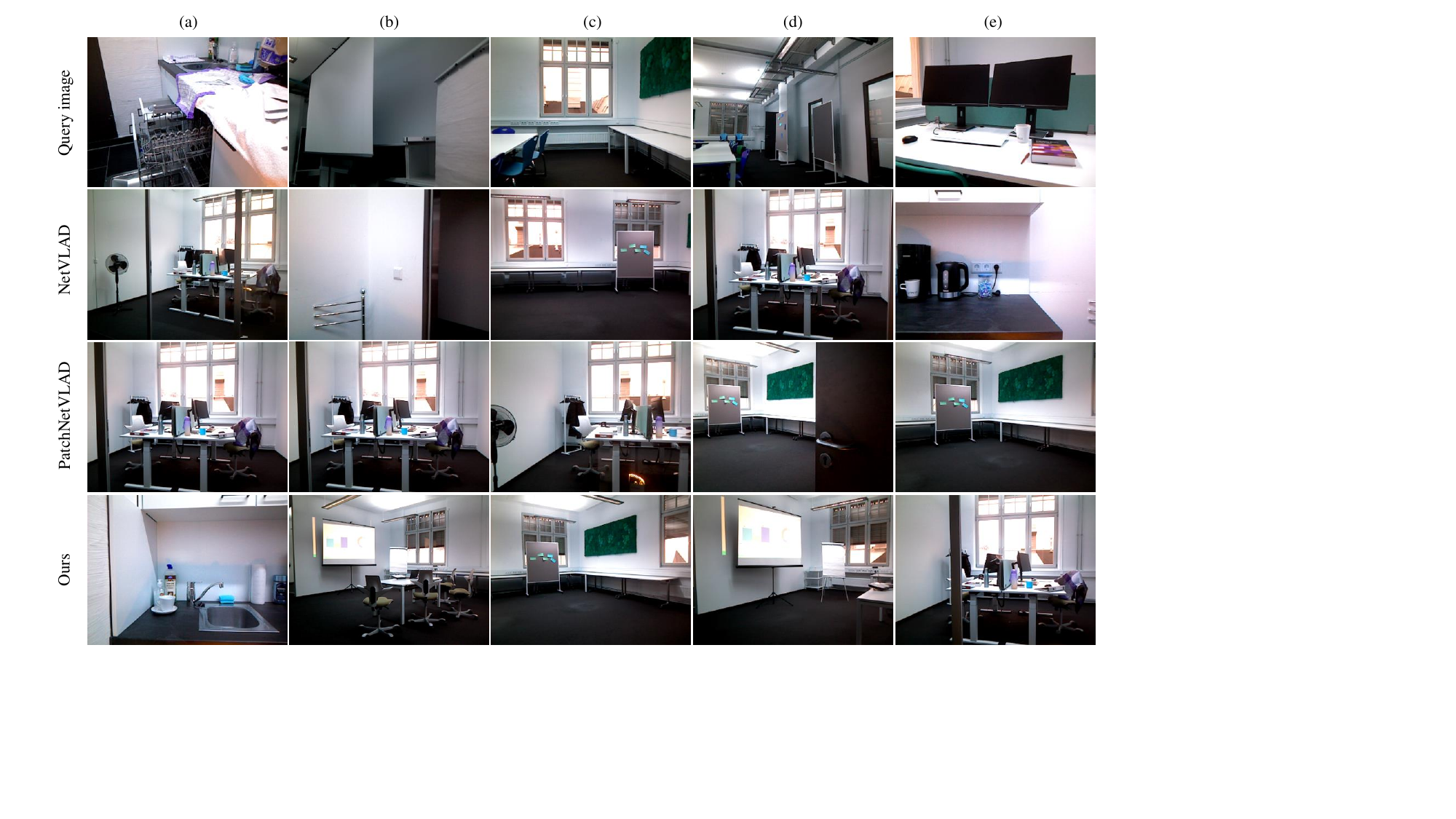}
\caption[Comparison of query and retrieved images for different methods]{Comparison of query and retrieved images for different methods.}
\label{floor1_picsMatch}
\end{figure}


We compare the performance of the proposed method to two state-of-the-art appearance-based localization approaches, NetVLAD~\cite{arandjelovic2016netvlad} and Patch NetVLAD~\cite{patchnetvlad}.
In our approach, we use the COCO dataset class labels as $\mathcal{L}^{obj}$ and extend it with labels of multiple objects found in our office environment, e.g., \emph{``living wall portrait''}. 

\autoref{floor1_picsMatch} shows qualitative results as a subset of the query images and their corresponding matches according to various approaches. The depicted query images emphasize different appearance variations in our dataset. Columns (a), (b), and (e) represent a combination of viewpoint changes and object re-arrangements. These appearance changes are difficult to handle for the baselines, whereas the foundation-model-based method robustly retrieves the correct references. 
The query in (c) includes an object manually added to $\mathcal{L}^{obj}$, a \emph{``living wall portrait''}. We found that this object helps our algorithm to localize within the room and to obtain a reference image with a similar viewpoint. In contrast, here NetVLAD and Patch NetVLAD focus on the distinct features of the window frame.
Lastly, the query image in (d) shows a view that is not represented in the reference trajectory. However, since the room is correctly categorized as the ``conference room'', the language-based method can successfully match this image to a reference frame in the same room. Note that the capability to easily add objects to the vocabulary is one of the strengths of our method. Given that we rely on foundation models, we generally do not need to train specific classifiers and can instead rely on the generative capabilities of the foundation models. We find this very convincing as it enables users to improve their robot's navigation performance by just providing them with a text label. 

For quantitative evaluation, we use two metrics. First, we compute the average translation error between the query image camera poses and the poses of the retrieved reference images. Second, we use the detected room labels to calculate the percentage of correct room detections. The numerical results are summarized in \autoref{1stfloor_results} and \autoref{3rdfloor_results} and are reported separately for each room and for the complete query trajectories. Our approach outperforms all baselines in the kitchen and office for both datasets. In the conference room of dataset~1, it is second to NetVLAD regarding the translation error but better on the room detection metric. In the hallway, however, our approach is behind the baselines due to the lack of distinct objects in that area. Averaged over the full trajectory of dataset~1, the proposed method has the lowest translation error with $4.98$\,meters which is $45.21\%$
lower compared to the second-best. We further achieve the highest room detection rate of $88.99\%$ on dataset~1 whereby the next best baseline achieves $54.13\%$. The results follow a similar trend for dataset~2 with an overall error of $3.96$\,meters, 
$7.26\%$ lower than the second-based, and an overall room detection rate of $84.16\%$.


\begin{table*}[h]
\caption[Comparison between language-based localization and visual feature-based methods for dataset~1]{Comparison between language-based localization and visual feature-based methods for dataset~1.}
\begin{center}
\resizebox{\textwidth}{!}{%
\begin{tabular}{ccccccccccc}
\hline
 & \multicolumn{2}{c}{Kitchen} & \multicolumn{2}{c}{Hallway} & \multicolumn{2}{c}{Conference room} & \multicolumn{2}{c}{Office} & \multicolumn{2}{c}{Total} \\
\hline
 & mean error & room detection & mean error & room detection & mean error & room detection & mean error & room detection & mean error & room detection \\
 & [m] & [\%] & [m] & [\%] & [m] & [\%] & [m] & [\%] & [m] & [\%] \\
\hline
NetVLAD & 4.08 & 18.18 & 0.42 & 100 & \textbf{4.19} & 22.97 & 7.73 & 37.50 & 9.09 & 39.10 \\
PatchNetVLAD & 3.86 & 9.09 & \textbf{0.20} & \textbf{100} & 5.95 & 59.46 & 9.96 & 12.50 & 10.96 & 54.13 \\
FM-Loc & \textbf{2.59} & \textbf{77.27} & 1.16 & 95.83 & 5.71 & \textbf{85.25} & \textbf{3.97} & \textbf{95.83} & \textbf{4.98} & \textbf{88.99} \\
\hline
\end{tabular}}
\end{center}
\label{1stfloor_results}
\end{table*}

\begin{table*}[h]
\caption[Comparison between language-based localization and visual feature-based methods for dataset~2]{Comparison between language-based localization and visual feature-based methods for dataset~2.}
\begin{center}
\resizebox{\textwidth}{!}{\begin{tabular}{rrrrrrrrrrr}
\label{3rdfloor_results}

\begin{tabular}{ccccccccccc}
\hline
   & \multicolumn{ 2}{c}{Kitchen} & \multicolumn{ 2}{c}{Hallway} &  \multicolumn{ 2}{c}{Conference room}  & \multicolumn{ 2}{c}{Office} &  \multicolumn{ 2}{c}{Total}\\
\hline
  & mean error & room detection & mean error & room detection & mean error & room detection & mean error & room detection  & mean error & room detection \\
   &  [m]& [\%]& [m] & [\%]& [m]&[\%]& [m]&[\%]& [m]&[\%]\\

\hline
NetVLAD & 6.49 & 47.83  & \textbf{1.10} & \textbf{100} & 7.61 & 16.67 & 4.3 & 0 & 4.27 & 59.41\\
PatchNetVLAD & 5.48 & 56.52  & 1.55 & 88.89 & 7.18 & 22.22 & 5.45 & 38.46 & 4.15 & 67.33\\
FM-Loc & \textbf{3.19} & \textbf{91.3}  & 2.145 & 92.6 & \textbf{7.17} & \textbf{44.44} & \textbf{2.05} & \textbf{100} & \textbf{3.96} & \textbf{84.16}\\
\hline
\end{tabular}
\end{tabular}}
\end{center}
\end{table*}

\subsection{Landmark Learning}
Not all detected objects are equally informative for localization, some objects are frequently displaced (e.g., cups) or are repetitive (e.g., windows). For robust localization, it is therefore beneficial to consider only static and unique landmarks. One possibility to select landmark candidates from a general object list is to prompt an LLM with the object list and the above criteria of a landmark. However, the LLM response is not grounded in the environment of the robot. Another approach is to learn the landmarks for a concrete location from data. To find the most reliable landmarks in our dataset, we repeatedly perform localization on the same data, each time eliminating one object and storing the mean translation error for that run.
We consider all objects that reduce the error by at least $0.1$\,meters as landmarks $\mathcal{L}^{lm}$. \autoref{landmarks_combined} show the impact of some objects on the mean error for the two different environments represented by our datasets. 

The landmark set contains the most informative objects in the considered rooms: chair, desk, and computer monitor in the office; flipchart, projector screen, desk, conference room table, and living wall portrait in the meeting room; and fire extinguisher, door, and mirror in the hallway. 
Objects frequently appearing in different rooms, e.g., dishes, increase the mean error and are therefore excluded from the landmark set. Filtering the detected object labels $L_i^{obj}$ by $\mathcal{L}^{lm}$ in the image descriptor decreases the mean localization error over the complete query trajectory to $4.71$\,meters for dataset~1 and $3.64$\,meters for dataset~2.

\begin{table}[h]
\caption[Learned landmarks for dataset~1 and dataset~2 and their effect on reducing the mean error]{Learned landmarks for dataset~1 and dataset~2 and their effect on reducing the mean error.}
\label{landmarks_combined}
\centering
\footnotesize
\begin{tabular}{lc lc}
\toprule
\multicolumn{2}{c}{\textbf{Dataset 1}} & \multicolumn{2}{c}{\textbf{Dataset 2}} \\
\cmidrule(r){1-2} \cmidrule(l){3-4}
\textbf{Object} & \textbf{Error Reduction [m]} & \textbf{Object} & \textbf{Error Reduction [m]} \\
\midrule
living wall portrait   & 0.71 & desk                & 0.57 \\
projector screen       & 0.66 & shelf               & 0.47 \\
conference room table  & 0.51 & exit sign           & 0.44 \\
computer monitor       & 0.24 & projector screen    & 0.17 \\
fire extinguisher      & 0.19 & dishwasher machine  & 0.16 \\
exit sign              & 0.16 & pinboard            & 0.10 \\
\bottomrule
\end{tabular}
\end{table}

\subsection{Joint Semantic and Local Feature Comparison}
\label{sec:joint_score}
The language-based semantic descriptor focuses on high-level concepts like object and room classes, which makes it robust to scene dynamics and camera view variations. However, it neglects salient image gradients that are not part of an object and can therefore fail in locations with sparse object occurrences, e.g., in hallways. We thus investigate if combining the semantic matching score with the score of Patch NetVLAD improves localization. 

For each query $I_i$, we retrieve the reference $I_p$ and corresponding score $S_{i,p}^{\mathit{pat}}$ with Patch NetVLAD and then normalize these scores by dividing them by the highest score in the query trajectory, i.e., $\hat{S}_{i,p}^{\mathit{pat}} = S_{i,p}^{\mathit{pat}}/\max_j(S_{j,p}^{\mathit{pat}})$. Next, we compute a reference image $I_s$ and score $S_{i,s}^{\mathit{\mathit{sem}}}$ with the semantic approach and also the scores $\hat{S}_{i,s}^{\mathit{pat}}$ and $S_{i,p}^{\mathit{sem}}$. We choose $I_s$ as the final match if $S_{i,s}^{\mathit{sem}} + \hat{S}_{i,s}^{\mathit{pat}} > S_{i,p}^{\mathit{sem}} + \hat{S}_{i,p}^{\mathit{pat}}$ and $I_p$ otherwise. Using this joint approach improves the results on both datasets. For dataset~1 the overall mean error is reduced to $4.64$\,m and for dataset~2 it is reduced to $3.1$\,m. As expected, these improvements mainly happen in the hallway for both datasets.

\section{Conclusion}

In this chapter, we presented FM-Loc, a novel localization system that leverages foundation models—specifically CLIP and GPT-3—to construct robust and interpretable semantic descriptors from visual observations. By grounding object and room labels and exploiting their semantic relationships, FM-Loc achieves accurate localization even under severe viewpoint changes and significant scene rearrangements. Unlike traditional visual feature-based methods, it requires no training or fine-tuning, yet delivers substantial improvements in both room classification and localization accuracy across diverse real-world environments.

Beyond its strong performance, FM-Loc identifies informative semantic landmarks that contribute most to localization, making the system not only effective but also interpretable and adaptable to specific settings. These results highlight the potential of foundation models to enrich robotic perception with high-level semantic understanding.

Having addressed localization in this chapter, we next turn to the domain of manipulation—another central challenge in robotics. In the following chapter, we explore how foundation models can be used to guide robots in grasping objects not only successfully, but also meaningfully. We investigate how semantics and contextual understanding can be integrated into the manipulation pipeline using vision-language and language models, further extending the capabilities of AI in embodied systems.

%% file: publications/langrasp.tex
\chapter{Lan-grasp: Using Large Language Models for Semantic Object Grasping}
\label{chapter:langrasp}
\chaptermark{Lan-grasp: Semantic Object Grasping}

The work presented in this chapter has been published in~\cite{Lan-grasp}:

\vspace{1cm}%
\hspace*{1cm}%
\begin{minipage}{.9\textwidth}%
R. Mirjalili, M. Krawez, S. Silenzi, Y. Blei and W. Burgard.

\textbf{Lan-grasp: Using Large Language Models for Semantic Object Grasping}

\textit{International Symposium on Robotics Research (ISRR), 2024.}


\end{minipage}%

\vspace{1cm}%

\newpage

\section*{Abstract}
\begin{adjustwidth}{1.2cm}{1.2cm} 
\small 
\textbf{
In this chapter, we propose Lan-grasp, a novel approach towards more appropriate semantic grasping. We use foundation models to provide the robot with a deeper understanding of the objects, the right place to grasp an object, or even the parts to avoid. This allows our robot to grasp and utilize objects in a more meaningful and safe manner. 
We leverage the combination of a Large Language Model, a Vision Language Model, and a traditional grasp planner to generate grasps demonstrating a deeper semantic understanding of the objects. We first prompt the Large Language Model about which object part is appropriate for grasping. Next, the Vision Language Model identifies the corresponding part in the object image. Finally, we generate grasp proposals in the region proposed by the Vision Language Model. Building on foundation models provides us with a zero-shot grasp method that can handle a wide range of objects without the need for further training or fine-tuning.
We evaluated our method in real-world experiments on a custom object data set. We present the results of a survey that asks the participants to choose an object part appropriate for grasping. 
The results show that the grasps generated by our method are consistently ranked higher by the participants than those generated by a conventional grasping planner and a recent semantic grasping approach. In addition, we propose a Visual Chain-of-Thought feedback loop to assess grasp feasibility in complex scenarios. This mechanism enables dynamic reasoning and generates alternative grasp strategies when needed, ensuring safer and more effective grasping outcomes.\footnote[1]{Video available at 
\href{https://tinyurl.com/5bnwpkuc}{https://tinyurl.com/5bnwpkuc}.}}
\end{adjustwidth}

\section{Introduction}
To function effectively in human environments, robots must be able to grasp and manipulate a wide variety of objects in a manner that is both contextually appropriate and physically sound. Objects found in household environments often require a specific way of interaction. For artificial objects, such as tools, the deployment mode can be implied by their design which ensures functionality and user safety. For instance, a knife should be typically held by the grip and not the blade. Similarly, a mug with hot tea is best held by the handle and not the rim of the mug. Incorrect handling can also impair the object itself, e.g. trying to carry a plant by the leaves would most likely lead to damage. Through experience, humans develop an intuitive understanding of objects, their parts, and their proper usage. As robots are increasingly involved in human living environments, it is crucial to provide them with the same kind of semantic knowledge.

\begin{figure}[ht!]
\centering
\includegraphics[clip,trim=3.5cm 4cm 3.7cm 3cm,width=0.6\linewidth]{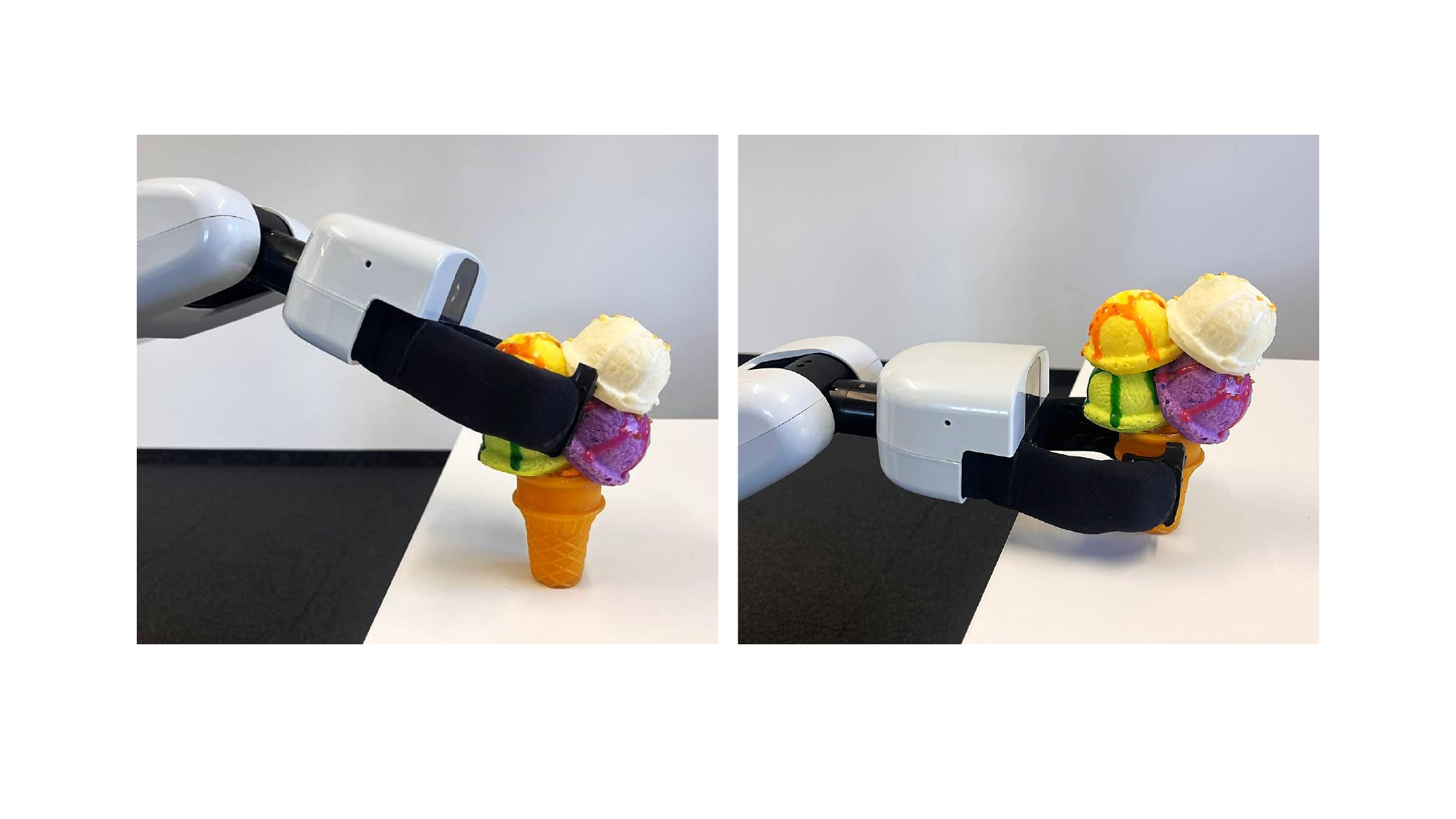}  
\caption[Robot performing the command of \emph{``Pick up the ice cream please''}]{Robot performing the command of \emph{``Pick up the ice cream please''}. The grasp on the left is generated without including semantic information while the grasp on the right is performed using our method leveraging a deeper understanding of the task and the object provided by Large Language Models. }
\label{fig:coverpic}
\end{figure}

Traditional approaches to robotic grasping~\cite{bicchi1995closure,miller2000graspit,ten2017grasp} only analyze the object geometry and aim to optimize the grasp stability.  Without a deeper understanding of semantic aspects as described above, this can limit the usability of tools or result in object or robot impairment. Recent data-driven approaches~\cite{jang2017end,kwak2022semantic,wu2023learning} also account for the object class and can generate grasps appropriate for the specific object type. Several works~\cite{murali2021same,tang2023task,tang2023graspgpt} tackle the problem of task-specific grasping where the object is grasped differently depending on the action at hand. However, most of these methods require substantial computational resources for training and can fail to generalize to unseen object categories. Our objective is an approach for object-specific grasping that ensures tool usability and safety without any training.


We proceed towards this goal by introducing Lan-grasp, a zero-shot method built on foundation models. The scale of these models and the massive size and generality of their training data allow us to reason about a large variety of objects without further training or fine-tuning. In particular, Lan-grasp uses a Large Language Model (LLM) to understand which part of an object is suitable for grasping. Next, this information is grounded in the object image by leveraging a Vision Language Model (VLM). Our method uses GPT-4 as the LLM and the OWL-ViT~\cite{owl-vit} as the VLM. However, due to the modular structure of Lan-grasp, it can easily be adapted to use other LLMs or VLMs. Finally, we use an off-the-shelf grasp proposal tool~\cite{miller2000graspit} to plan the grasps in accordance with the admissible parts of the object detected by the deployed foundation models.

\textbf{In summary, this chapter includes the following contributions: }
\begin{enumerate}
    \item We propose a novel approach using foundation models for zero-shot semantic object grasping.
    \item We demonstrate that the presented approach can work with a wide variety of day-to-day objects without the need for additional training.
    \item We evaluate our approach by asking human participants to choose the appropriate grasps.
    \item We employ a feedback mechanism using Visual Chain-of-Thought prompting to assess grasp feasibility and dynamically generate alternative grasp strategies when needed.
\end{enumerate}


\section{Related Work}
\label{sec:related_work}

Traditional grasping algorithms~\cite{bicchi1995closure,miller2000graspit,ten2017grasp,zapata2019fast,myers2015affordance} analyze the geometry of the object and the gripper to propose and evaluate a grasping pose. Building on decades of development, these methods are fast and reliable off-the-shelf tools. However, they do not incorporate semantic information and operate based on object shape only. Also, such methods rely on a precise object model and thus suffer from partial or noisy geometry. Data-driven approaches regress grasping candidates from either single view RGB images~\cite{jang2017end,redmon2015real} or point clouds~\cite{zhao2021regnet,alliegro2022end}, thus mitigating the need for a complete object model. Further, a network can learn a more natural grasping policy if human-like grasps are included in the training data, where such grasps are either created manually~\cite{corona2020ganhand} or learned through imitation~\cite{wu2023learning}.

Our work is closely related to task-oriented grasping (TOG) and affordance detection. TOG methods restrict the grasp candidates to a specific object part or area, conditioned on the action at hand. Murali~\etal~\cite{murali2021same} create a data set with a large number of objects and tasks and manually annotate task-specific grasp poses. Then, the authors use that data to train a grasp evaluation network. Kwak~\etal~\cite{kwak2022semantic} deploy a knowledge graph to select the gripper type and gripping force appropriate for the given object. Chen~\etal~\cite{chen2021joint} propose a network that jointly detects an object and generates a grasping pose according to a natural language command. However, the training requires object, grasp, and command ground truth data. Fang~\etal~\cite{fang2020learning} introduce TOG-Net, which optimizes task-oriented grasps and manipulation policies using simulated self-supervision.

Similarly to TOG, affordance detection is the problem of identifying objects or object parts that accommodate a certain action. Do~\etal~\cite{do2018affordancenet} propose an end-to-end trained network that detects object instances in an image and assigns pixel-wise affordance masks to object parts. Liu~\etal~\cite{liu2020cage} build on the previous work as a backbone for affordance detection and, in addition, infer the material of object parts to further facilitate semantic grasping. Monica and Aleotti~\cite{monica2020point} propose a system that decomposes an object point cloud into meaningful parts which then serve as grasping targets. However, the part the robot has to grasp is provided by the user whereas in our method the part is suggested by an LLM. Bohg~\etal~\cite{bbohg2013data} survey data-driven approaches to grasp synthesis, focusing on methods that sample and rank candidate grasps for both familiar and unknown objects, highlighting the role of feature extraction in these approaches. Nasiriany~\etal~\cite{nasiriany2024pivot} introduce a prompting framework for VLMs (PIVOT) that refines candidate actions iteratively, showing potential for spatial tasks such as grasping, but their focus is broader, addressing both navigation and manipulation tasks. Wei~\etal~\cite{wei2024grasp} propose a novel dataset, DexGYSNet, and utilize it to train a model for dexterous grasp generation based on language guidance. Jian~\etal~\cite{jian2023affordpose} introduce AffordPose, a large-scale dataset for affordance-driven hand-object interactions, focusing on part-level affordance labeling to guide the generation of hand-object interactions in fine detail. Zhu~\etal~\cite{zhu2021toward} propose a framework for human-like dexterous grasping, using semantic touch codes and object functional areas to guide grasps. Ren~\etal~\cite{ren2023leveraging} introduce ATLA, a meta-learning framework that uses LLMs to accelerate tool manipulation by combining language-based policies with affordances, focusing on general tool use rather than grasping specific object parts.

Foundation models have recently attracted a lot of attention in different sub-fields of robotics~\cite{mirjalili2023fm-loc,huang2023audio,cui2023no} and
have been also applied to boost TOG and affordance detection. Ngyen \etal~\cite{ngyen2023open} train an open-vocabulary affordance detector for point clouds whereby CLIP is deployed to encode the affordance labels. Similarly, Tang \etal~\cite{tang2023task} use CLIP to facilitate task-specific grasping from RGB images and language instructions. The authors propose to utilize CLIP embeddings from intermediate CLIP layers to allow their affordance detector to reason about fine-grained object parts. Gao \etal~\cite{gao2023physically} annotate a large object data set with physical object properties like mass or fragility and fine-tune a VLM on it to improve manipulation planning. Other methods integrate LLMs for encoding tasks or object parts from natural language. Song \etal~\cite{song2023learning} use BERT as the language back-end and train a network that grounds object parts in a point cloud from a user instruction. Here, however, the part label is explicitly referred to in the user input. The approach of Tang \etal~\cite{tang2023graspgpt} lifts this limitation by prompting an LLM to describe the shape and parts of an object. The LLM response is then processed by a Transformer-based grasp evaluation network. Our method also relies on an LLM for deciding what object part should be grasped. The crucial difference to the above works is that our approach relies solely on foundation models and does not require any training. Thus, once more powerful foundation models are available, the performance of our approach is easily improved by switching to a novel LLM or VLM. Newbury ~\etal~\cite{newbury2023deep} conduct a systematic review of deep learning methods for six-DoF grasp synthesis, highlighting sampling-based, direct regression, and reinforcement learning methods to generate grasp poses. Wu~\etal~\cite{wu2024see} propose an approach to enhance LMMs’ robustness in vision tasks, introducing reasoning capabilities to correct false premises, which improves reasoning for affordance-based grasping tasks. In a similar line of thought, Huang ~\etal~\cite{huang2023voxposer} propose VoxPoser, a framework that generates 3D value maps to guide robotic manipulation using affordances extracted from LLMs and visual grounding, however their focus is on manipulation tasks rather than detecting the specific grasping part of an object.
Finally, embodied vision-language models like PALM-E~\cite{driess2023palm} aim to close the gap between language, vision, and robot actions by training the network jointly on these modalities. The recent RT-2 model~\cite{brohan2023rt} shows remarkable capabilities of generating robot controls from language instructions. However, these models are computationally expensive, and despite their zero-shot capabilities, they often require fine-tuning to perform well in novel environments.

\section{Approach}

\begin{figure*}[ht]
\centering
\includegraphics[clip,trim=0cm 3.5cm 0cm 2cm,width=1\linewidth]{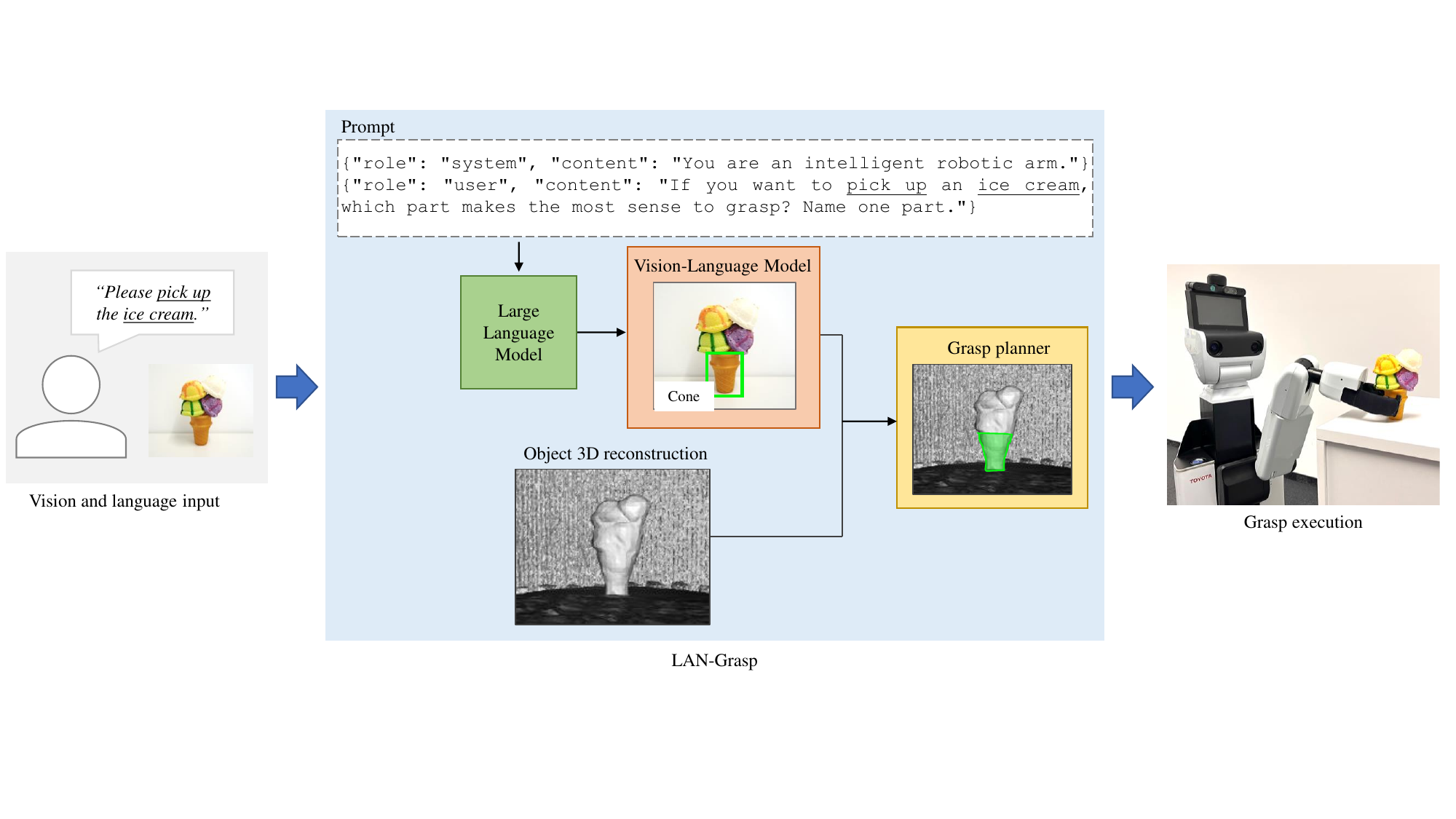}  
\caption[Lan-grasp in a nutshell]{Lan-grasp in a nutshell: The command from the user is turned into a prompt suitable for the Large Language Model (LLM). With this prompt as an input, the large language model outputs the proper part for grasping the object, which in this example is the \emph{cone}. This word is then grounded to the object image using a Vision-Language Model (VLM). The grounded grasp part is integrated to the 3D reconstruction model of the object to generate the proper grasp.}
\label{fig:langraspoverview}
\end{figure*}

In this section, we explain the details of our approach. Lan-grasp generates a grasping pose from an object label, a camera image showing that object, and the corresponding object geometry. The method consists of two parts. In the language module, an LLM first decides what object part to grasp, which is then grounded in the image by a VLM. The resulting bounding box is projected onto the object geometry to mark the grasp target in the grasp planning module.  Thanks to the modular structure of Lan-grasp, it is easy to enhance the pipeline by leveraging more advanced models as they emerge. The pipeline of Lan-grasp is depicted in \autoref{fig:langraspoverview}.

\subsection{Language Module}
\label{sec:method_language}
In the first step, the object label {\tt<object>} provided by the user is transferred into a LLM prompt. The scheme of the prompt is chosen to be compatible with GPT-4 which is the LLM that we used in the pipeline~\cite{openai2023gpt4}. We included the last sentence to prevent the LLM from giving extra explanations and thus only output the desired object part.
We use OWL-ViT~\cite{owl-vit} as the VLM for grounding the object part label in the image. 
It builds on the Vision Transformer Architecture, first presented by Dosovitskiy~\etal~\cite{dosovitskiy2020image}. 
The authors then pre-train the model using contrastive learning \cite{zhai2022lit} on a large image-text data set \cite{jia2021scaling}.
Afterward, the authors fine-tune the model on publicly available detection data sets. OWL-ViT then detects and marks the desired object part with a bounding box which is projected on the object 3D model.

\subsection{Grasp Planning Module}
\label{sec:grasp_planning}
We deploy the GraspIt! simulator~\cite{miller2004graspit} as our grasp proposal generator. It is a standard tool that operates on geometric models and evaluates grasps according to physical constraints. Thus, the first step for grasp planning is to create a dense 3D mesh model of the object. In our setup, we use two fixed RGB-D sensors and a turning table for object scanning. We acquire the camera poses from an Aruco board and integrate the depth images via KinectFusion~\cite{newcombe2011kinectfusion}. However, we note that any other suitable reconstruction approach could be used here. 

The possible poses for grasping the object are generated by sampling. The initial gripper position is chosen based on object geometry, after that the gripper is iteratively brought closer to the object while avoiding obstacles~\cite{miller2003automatic}. In this regard, GraspIt! splits the scene into object and obstacle geometry, and we exploit this mechanism by marking the mesh parts that project into the VLM-generated bounding box as object and the rest as obstacle. This enforces grasping only at the desired object part. The resulting grasp proposals are ranked based on grasp efficiency and finger friction.


We want to point out that our approach is agnostic about the grasp planner and could be potentially replaced by other tools that do not require a complete object model, e.g., the method of Alliegro~\etal~\cite{alliegro2022end}. In this case, the reconstruction step could be skipped entirely and the grasp candidates could be computed on a point cloud acquired from the robot's sensors.

\section{Experimental Evaluation}
\label{sec:results}
In this section, we present the details of our experiments and results. Our goal is to demonstrate that our method proposes to grasp object parts that are preferred by humans on a variety of objects. We argue that humans generally choose grasps that enable correct tool usage and ensure safety and that a robot retains these desirable qualities by executing similar grasps. To that end, we first collect a data set of typical household objects. Next, we apply our approach to these objects and execute the grasping on a real robot. Finally, we show that our grasping strategy is similar to human preferences obtained through a survey and that our approach outperforms two baselines based on this similarity metric. In the following, we describe our data set, provide details on the baseline approaches and the performed experiments, and discuss the results. In \autoref{sec:feasibility_feedback} we present an extension to the main algorithm that reasons about the feasibility of a grasp in complex scenarios. Finally, in \autoref{sec:ablation} we perform an ablation study on several components of the pipeline.


\subsection{Dataset}
\label{sec:dataset_original}
We collect a data set containing $22$ different objects commonly found in household environments. We chose these objects to cover a wide range of situations where semantic knowledge is required for proper grasping. Our first objective was to showcase grasping on functional objects like tools or kitchen supplies, e.g., \emph{shovel}, \emph{hand brush}, and \emph{knife}. Further, we included delicate objects that might be damaged with an improper grasp, for instance, \emph{rose}, \emph{cupcake}, and \emph{ice cream}. For other objects, a wrong grasp can cause a dangerous situation, e.g., \emph{candle}. Finally, we include objects where an improper grasp might not necessarily be harmful but is rather unnatural to a human observer, for instance, \emph{doll}, \emph{bag}, and \emph{wine glass}. The objects in the dataset are shown in \autoref{fig:tile1}, \autoref{fig:tile2}, and \autoref{fig:narrow}.


\subsection{Experimental Setup and Baselines}
For $11$ objects from our data set, listed in \autoref{tab:similarity}, we perform real-world experiments using the Human Support Robot (HSR)~\cite{yamamoto2019development}. we first scan each object and than apply GraspIt! to the resulting 3D model as detailed in \autoref{sec:grasp_planning}. From the top-$20$ grasp proposals we randomly pick one and execute it using the proprietary HSR motion planner. 

Our first baseline is the plain GraspIt! simulator. Here we use the same 3D models as for our approach but do not restrict grasping to the object part selected by the language model. For each object, we evaluate the top-$20$ grasp proposals and carry two of them out on the HSR, as shown in \autoref{fig:tile1} and \autoref{fig:tile2}.

The second baseline is GraspGPT~\cite{tang2023graspgpt}, a recent approach to task-oriented grasping. This method requires as input an object point cloud and a natural language prompt describing the object, the object class, and the task. We generate the point clouds from the object meshes reconstructed as above and use an object-specific activity as the task label. Again, we retrieve $20$ grasps per object but do not carry them out on the robot.


\subsection{Qualitative Results}
\begin{figure*}[htbp]
\centering

\includegraphics[clip,trim=1cm 8.07cm 0 1cm,,width=1\linewidth]{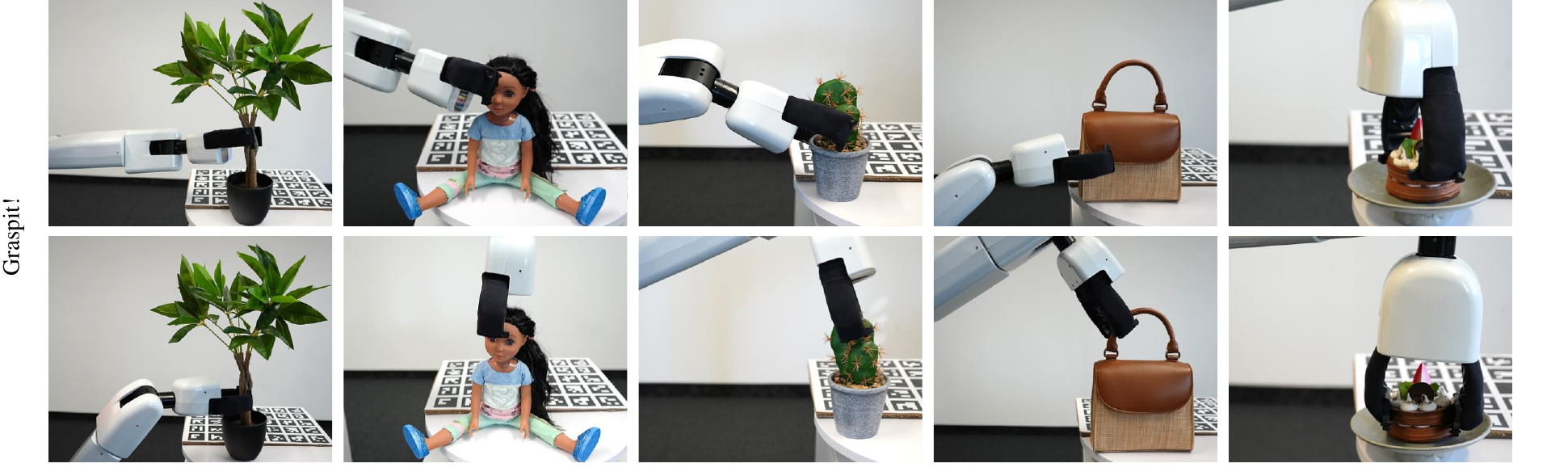}  
\includegraphics[clip,trim=1cm 1cm 0 0cm,,width=1\linewidth]{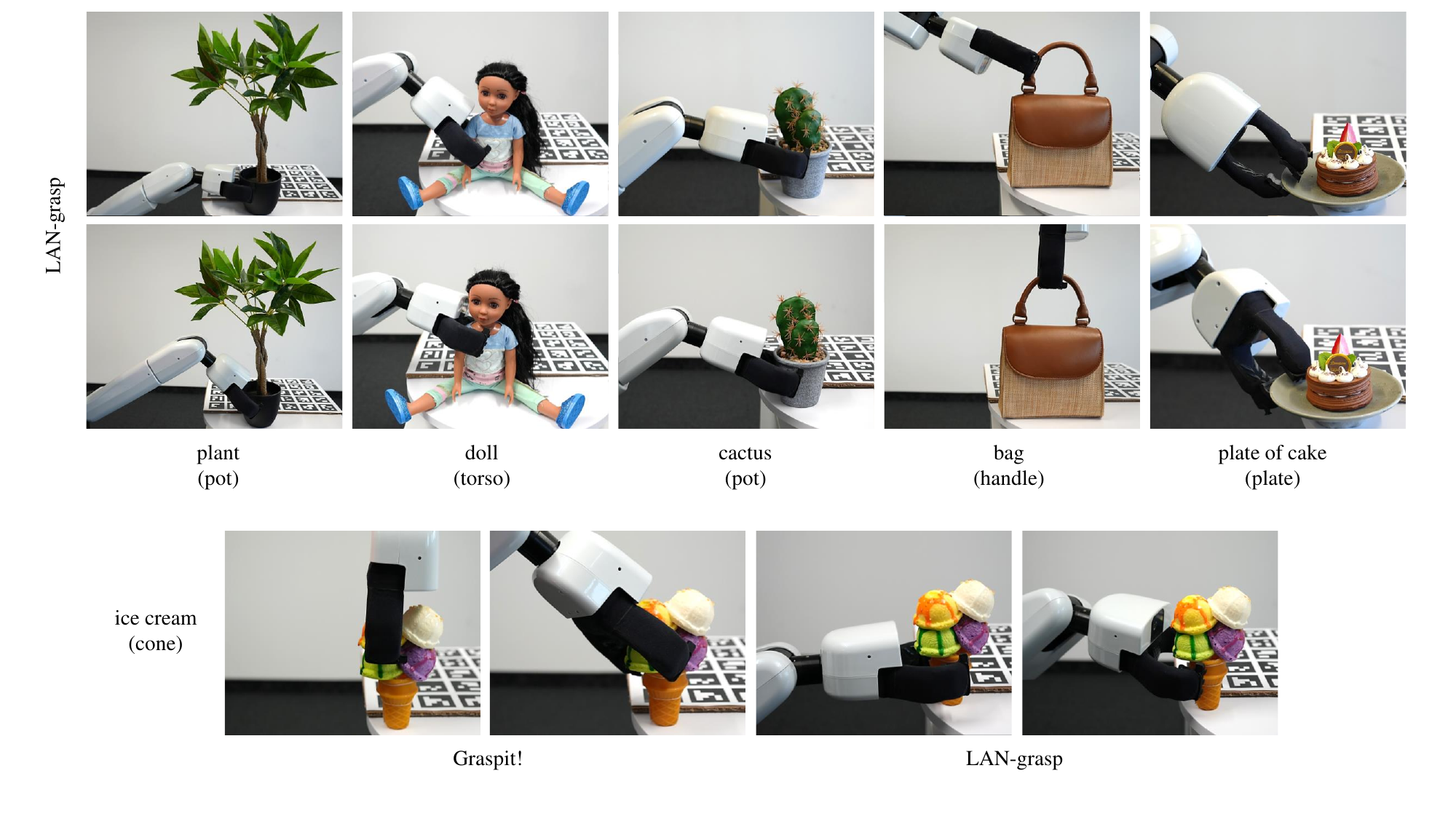}  
\caption[The performed grasps by the HSR robot]{The performed grasps by the HSR robot: Each column presents the grasps for one object. The first two rows for each object, show the grasps generated without semantic knowledge about the objects, while the third and fourth rows show the grasps generated by Lan-grasp.}
\label{fig:tile1}
\end{figure*}

\begin{figure*}[htbp]
\centering
\includegraphics[clip,trim=1cm 8.02cm 0 1cm,width=1\linewidth]{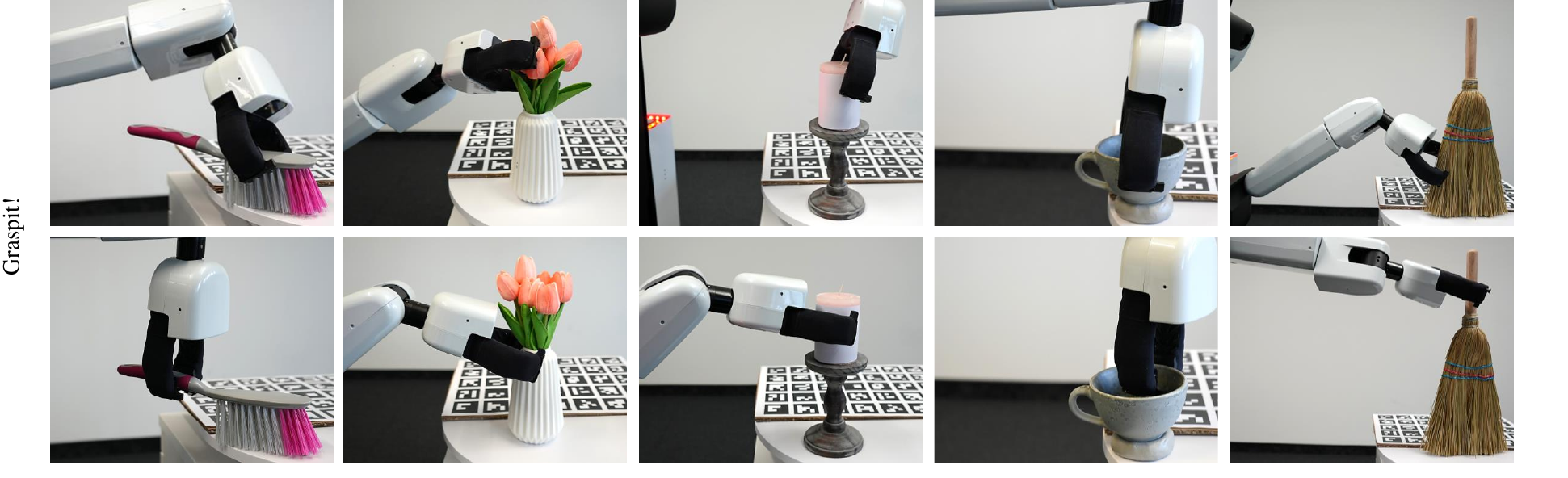}  
\includegraphics[clip,trim=1cm 6.5cm 0 0.7cm,width=1\linewidth]{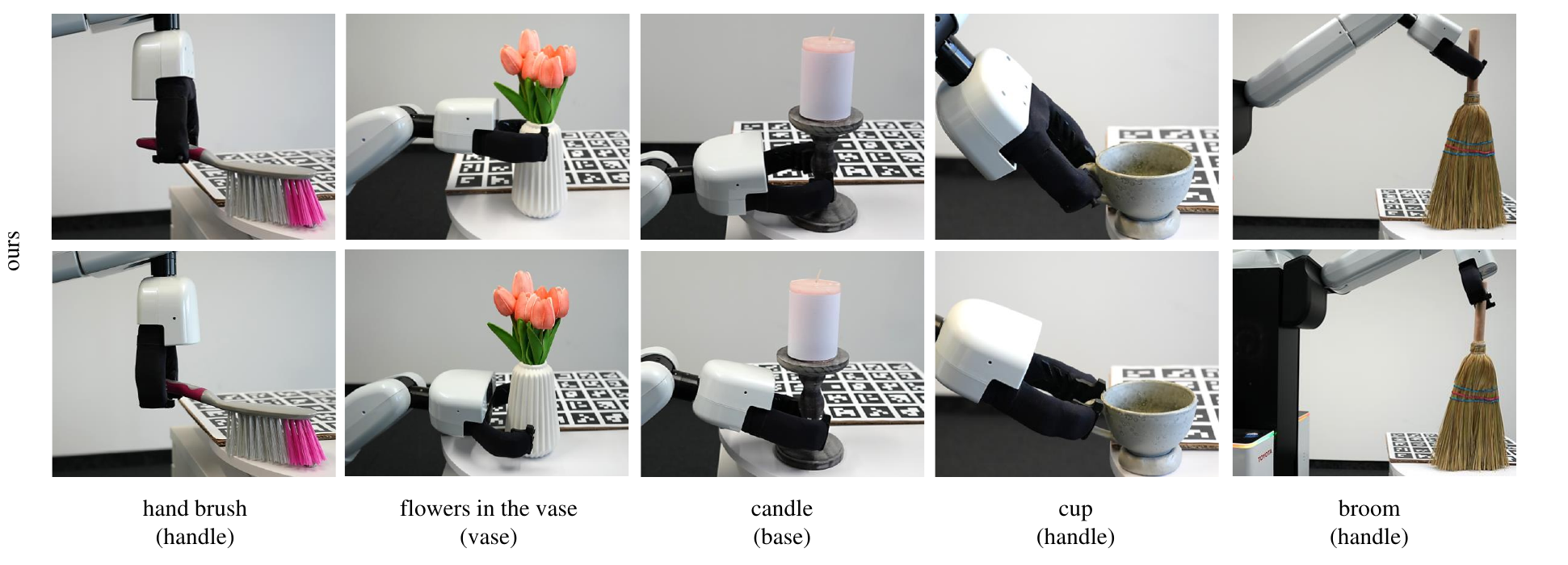}  
\caption[The performed grasps by the HSR robot]{The performed grasps by the HSR robot: Each column presents the grasps for one object. The first two rows for each object, show the grasps generated without semantic knowledge about the objects, while the third and fourth rows show the grasps generated by Lan-grasp.}
\label{fig:tile2}
\end{figure*}

\begin{figure*}[htbp]
\centering
\includegraphics[clip,trim=1cm 9cm 9cm 0.5cm,width=1\linewidth]{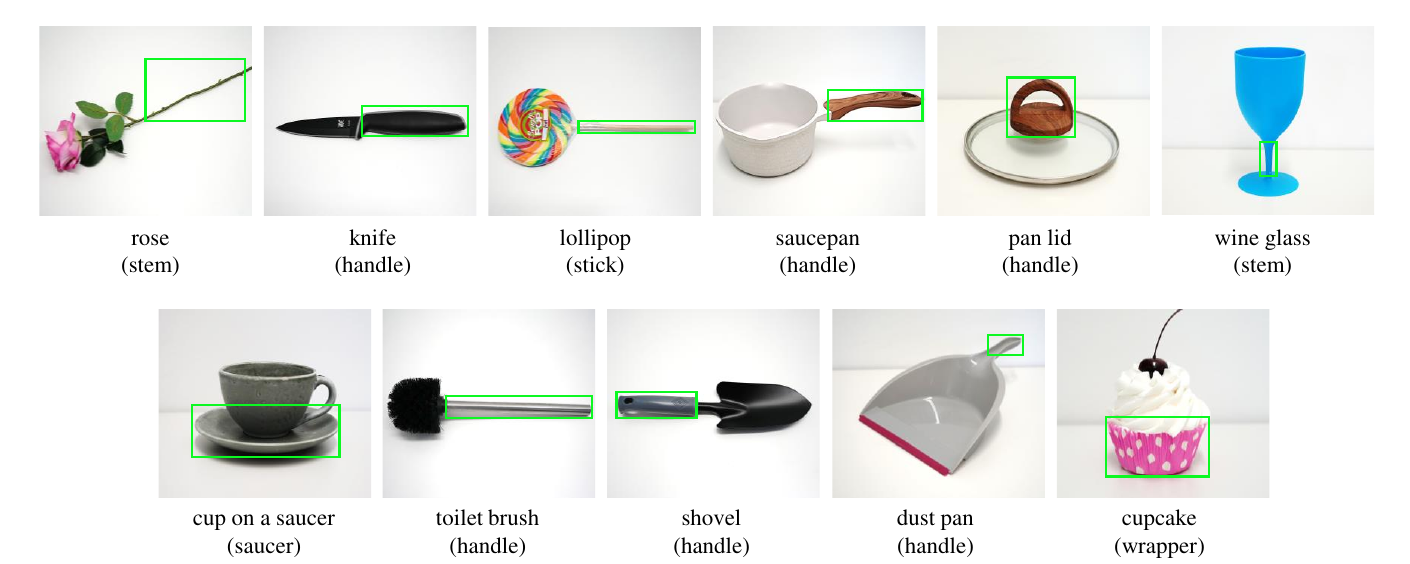}  
\caption[The results of Lan-grasp on a set of common household objects]{The results of Lan-grasp on a set of common household objects. The green bounding box shows the area to grasp suggested by the method. }
\label{fig:narrow}
\end{figure*}

In this section, we present and discuss the grasping results of Lan-grasp and the GraspIt! baseline. The grasps executed on the robot are shown in \autoref{fig:tile1} and \autoref{fig:tile2}. For the rest of the objects, the grasping area suggested by our method is presented in \autoref{fig:narrow}. 

The results suggest that Lan-grasp proposes grasps suitable for the usage of the respective object.
For instance, grasping the \emph{handle} for \emph{shovel} and \emph{broom} corresponds to the intended use of these items. For \emph{lollipop} and \emph{cupcake}, the grasp is placed away from the edible part at the \emph{stick} and the \emph{wrapper}, respectively. It is noteworthy that our method is able to understand the relation between stacked objects, e.g., ~\emph{flowers in a vase} or \emph{plate of cake}. Also, for a single \emph{cup}, Lan-grasp suggests grasping the \emph{handle} while for the \emph{cup on a saucer} the grasp proposal is the \emph{saucer}. Other objects, e.g., \emph{doll}, \emph{bag}, or \emph{wine glass}, do not possess a critical area where grasping would cause harm or directly interfere with the functionality. However, our method is able to generate grasps that are closer to how a human would handle these items. In contrast to Lan-grasp, the areas suggested by GraspIt! are expectantly random and do not consider semantic intricacies.

\newpage

\subsection{Quantitative Results}
\label{sec:results_quantitative}

To support the claim that our approach proposes grasps similar to human preferences, we designed a questionnaire on grasping choices. A group of $83$ participants were presented with images of all objects used in the experiments and were asked where they would grasp them. For each object, the participants could choose between two parts marked by bounding boxes in the image. The survey results are summarized in \autoref{tab:similarity}. Per object, we state the preferred part and the percentage of participants that selected it.

For the proposed approach and the baselines, we want to evaluate how similar the generated grasps are compared to the ones suggested by human users. Given that an object is segmented into parts $a$ and $b$, let $p_a \in [0,1]$ be the empirical probability that a method grasps at part $a$ and $p_b = 1 - p_a$ that part $b$ is grasped. Further, let $p_a^h$ be the human grasping frequency at $a$ according to the survey results and $p_a^x$ the corresponding frequency produced by one of the considered methods. To compute $p_a^x$ for the baselines, we obtained $20$ grasp proposals from each algorithm and counted the grasps falling into region $a$. Lan-grasp restricts the grasps to the object part selected by the LLM, which in our experiments robustly proposed the same part for a given object. Thus, the values of $p_a^x$ were here either $1$ or $0$. Finally, we computed a per-object similarity score for each method $x$ as $sim_x = 1-|p_a^h-p_a^x|$. These scores are shown in \autoref{tab:similarity} along with the average similarity scores over all objects.

\begin{table}[h]
\centering
\footnotesize
\caption[Similarity of grasping area preferences compared to a human user]{Similarity of grasping area preferences compared to a human user. The left half of the table lists the objects and the object part the majority of survey participants suggested for grasping, with the corresponding percentage of users. The right half shows similarity scores for the two baselines and the proposed method.}
\label{tab:similarity}
\begin{tabular}{llccc}
\toprule
\textbf{Object} & \textbf{Preferred Part (\%)} & \textbf{GraspIt!} & \textbf{GraspGPT} & \textbf{Lan-grasp} \\
\midrule
doll                 & torso 92.1       & 0.28 & 0.48 & \textbf{0.92} \\
ice cream            & cone 100.0       & 0.05 & 0.40 & \textbf{1.00} \\
candle               & base 93.1        & 0.22 & 0.57 & \textbf{0.93} \\
flowers in the vase  & vase 93.2        & 0.32 & 0.73 & \textbf{0.93} \\
bag                  & handle 91.1      & 0.79 & 0.69 & \textbf{0.91} \\
plant                & pot 94.3         & 0.16 & 0.56 & \textbf{0.94} \\
hand brush           & handle 95.4      & 0.65 & \textbf{0.95} & \textbf{0.95} \\
toilet brush         & handle 97.6      & 0.42 & 0.52 & \textbf{0.98} \\
cactus               & pot 98.8         & 0.26 & \textbf{0.99} & \textbf{0.99} \\
cupcake              & wrapper 100.0    & 0.10 & 0.40 & \textbf{1.00} \\
cup on a saucer      & saucer 81.2      & 0.24 & 0.59 & \textbf{0.81} \\
plate of cake        & plate 98.8       & 0.11 & 0.51 & \textbf{0.99} \\
mug                  & handle 77.1      & 0.28 & 0.73 & \textbf{0.77} \\
saucepan             & handle 94.3      & 0.36 & \textbf{0.94} & \textbf{0.94} \\
broom                & handle 97.6      & 0.42 & \textbf{0.98} & \textbf{0.98} \\
\midrule
\textbf{Average}     & --               & 0.31 & 0.67 & \textbf{0.94} \\
\bottomrule
\end{tabular}
\end{table}

Our method consistently outperforms the baselines on the similarity score and ties only in four cases with GraspGPT. The average similarity score of Lan-grasp is considerably higher with the value of $0.94$ compared to $0.31$ achieved by GraspIt! and $0.67$ achieved by GraspGPT. We further note that in all cases, the object part choice of Lan-grasp coincides with the majority vote of the survey participants. The low score of GraspIt! is not surprising since it only considers geometric and not semantic aspects of the object. Thus, whether the object is grasped in a particular region is pure chance, and we expect the similarity score to be closer to $0.5$ for a larger data set. GraspGPT exhibits a better performance compared to GraspIt! due to leveraging semantic concepts and LLMs. However, it was tuned on a data set mostly containing tools and house supplies and thus does not perform well on objects like a \emph{doll} or an \emph{ice cream}.




\begin{figure*}
\centering
\includegraphics[clip,trim=0.25cm 3cm 0.25cm 0cm,width=1\linewidth]{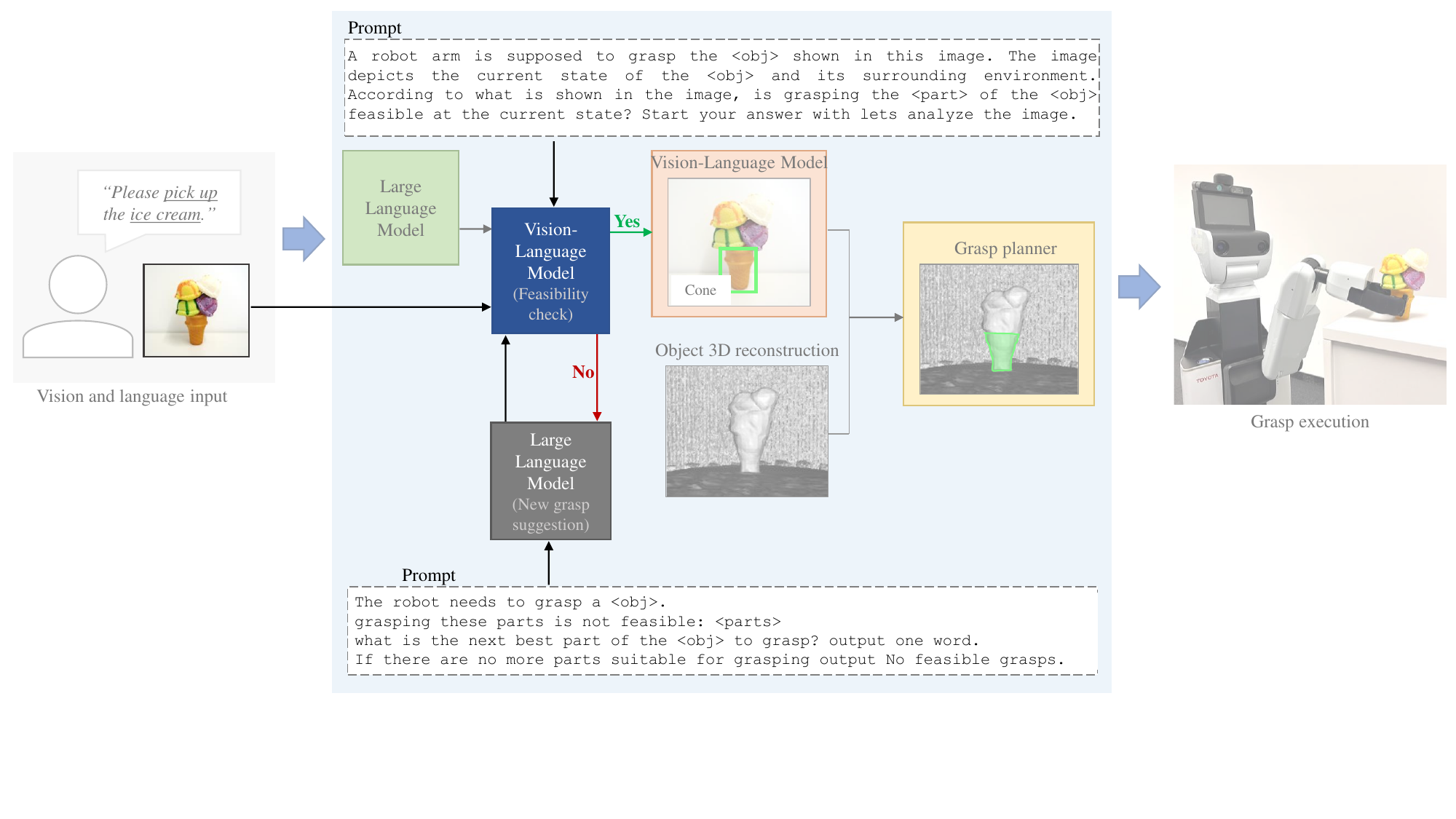}  
\caption[Schematic diagram of the feasibility feedback loop added to the core pipeline of Lan-grasp]{Schematic diagram of the feasibility feedback loop added to the core pipeline of Lan-grasp.}
\label{fig:main_feedback}
\end{figure*}

\begin{figure}[ht!]
\centering
\includegraphics[clip,trim=1cm 0.5cm 1cm 0cm,width=1\linewidth]{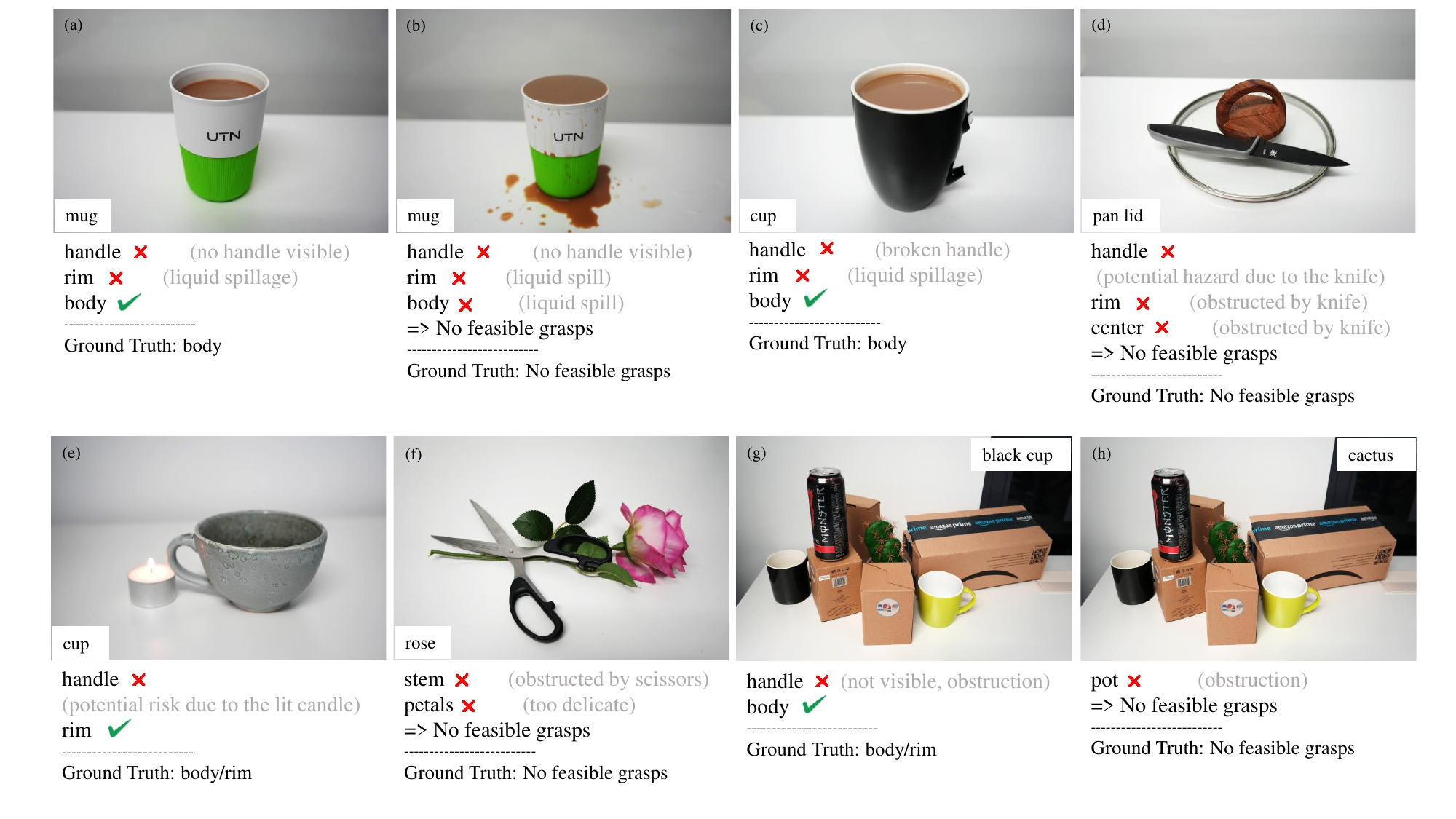}  
\caption[Qualitative results of Lan-grasp on complex grasping scenarios]{Qualitative results of our method on complex grasping scenarios. The object part labels below each image show the suggested grasp part and whether it was considered feasible or not by our feedback algorithm.}
\label{fig:feedback}
\end{figure}

\begin{figure}[ht!]
\centering
\includegraphics[clip,trim=1.5cm 8cm 1.5cm 0cm,width=1\linewidth]{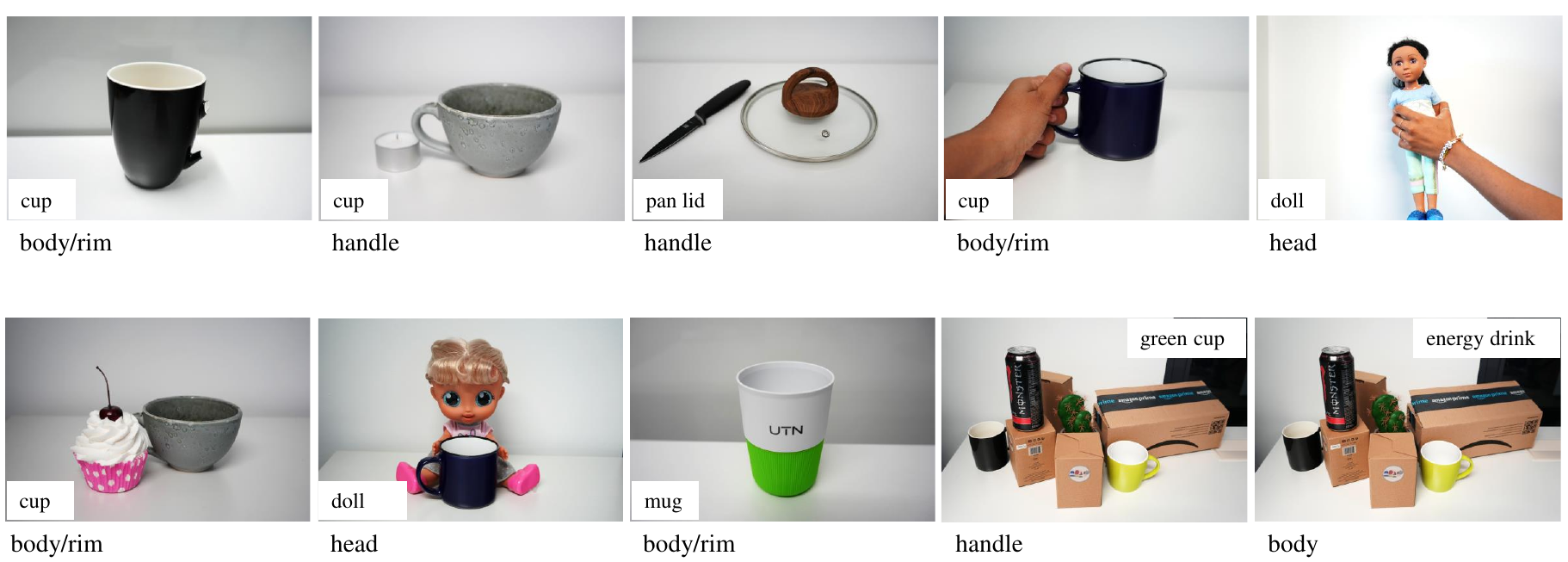}  
\caption[Instances of our dataset for complex grasp scenarios]{Instances of our dataset for complex grasp scenarios.}
\label{fig:feedback-dataset}
\end{figure}

\subsection{Grasp Feasibility Feedback}
\label{sec:feasibility_feedback}
So far, we assumed that the grasp suggested by the LLM is feasible, however, this is not always the case. For instance, the referenced object part might not be visible in the image, broken, or occluded by other objects. Further, the execution of the grasp could lead to undesirable outcomes. 
In order to mitigate this issue, we propose a feedback loop consisting of a VLM and an LLM that communicate with each other to find a feasible grasp. The approach and the used prompts are shown in \autoref{fig:main_feedback}.
As in our original pipeline, the LLM first suggests an object part based on the object label. 
Next, the VLM analyzes the image and evaluates the feasibility of the grasping part. Crucially, we follow the idea of Chain-of-Thought (CoT) prompting~\cite{kojima2022large} and include the sentence \noindent {{\tt ``Start your answer with lets analyze the image.''}} into the prompt. If feasible, we proceed as before. If not, the LLM receives a list of all object parts rejected so far and is asked to propose another grasp. This loop repeats until either a feasible grasp part is found or there are no more suitable parts of the object left for grasping. In our implementation, the roles of LLM and VLM are carried out by the same model (GPT-4o).

To evaluate this approach we gathered a dataset of $18$ challenging scenarios, shown in \autoref{fig:feedback-dataset} and \autoref{fig:feedback}, where the initial grasp suggestion is not feasible and requires reasoning to find the right grasp part. We defined the ground truth manually to evaluate the results. We ran the algorithm $5$ times to obtain an average success rate of $91.14\%$ over all scenes. The qualitative results are shown for part of the dataset in \autoref{fig:feedback}. Our algorithm can take into account different criteria, e.g., the risk due to the proximity of the lit candle, potential mess due to the overfilled cup, or occlusions by nearby objects. We further analyze the effects of VLM choice and CoT in the following section.


\subsection{Ablation Study}
\label{sec:ablation}
In this section, we evaluate the influence of different algorithm components on the above results, specifically the choice of the LLM and VLM and the effect of CoT on the feasibility feedback. First, we consider the pipeline without feedback as described in \autoref{sec:method_language}. Here we consider GPT-3.5-turbo and GPT-4 as LLMs. We further investigate whether replacing the LLM with a VLM improves the performance. To that end, we deploy GPT-4o, GPT-4o-mini, and the open-source LLaVA-1.5 7B VLM and provide them with an image of the object to grasp.

Here, we run our pipeline only until detecting the grasping area in the image, without executing the grasp nor generating a grasping pose. Therefore, we use different metrics than in \autoref{sec:results_quantitative}. First, we count the exact matches between the LLM-generated and GT object part labels and report the average success rate. Second, we compute the Intersection-over-Union (IoU) for the proposed grasping regions and the ground truth. All algorithm versions are evaluated on data from \autoref{sec:dataset_original} and the results are summarized in \autoref{tab:ablation_single_step}.

\begin{table}[b]
\centering
\footnotesize
\caption[Ablation of LLMs and VLMs in the main Lan-grasp pipeline]{Ablation of LLMs and VLMs in the main Lan-grasp pipeline.}
\label{tab:ablation_single_step}
\begin{tabular}{cccccc}
\toprule
\multirow{2}{*}{\textbf{Method}} & \multicolumn{2}{c}{\textbf{Text only}} & \multicolumn{3}{c}{\textbf{Text + Image}} \\
\cmidrule(lr){2-3} \cmidrule(lr){4-6}
 & GPT-3.5-turbo & GPT-4 & GPT-4o & GPT-4o-mini & LLaVA \\
\midrule
Success rate (\%) & 81.8 & 81.8 & 86.3 & 86.3 & 45.5 \\
IoU               & 0.63 & 0.64 & 0.65 & 0.64 & 0.38 \\
\bottomrule
\end{tabular}
\end{table}

Both text-only GPT versions perform equally with an $81.8\%$ success rate and an IoU score of around $0.63$. The VLM variants perform slightly better with a success rate of $86.3\%$. We note that there is no difference between the flagship GPT-4o and the downsized GPT-4o-mini model. LLaVA performed significantly worse with $45.5\%$ success rate and $0.38$ IoU score. Analyzing the object parts suggested by LLaVA showed that the model was correct for objects possessing a handle, for other objects the answer was either wrong or referred to generic image locations, e.g., ``bottom'' or ``top''. A straightforward explanation could be simply the smaller model size. However, another reason might be that we used the same prompt for LLaVA as for the GPT models, and better results could be achieved with further prompt engineering specifically targeting LLaVA.

For the feasibility feedback algorithm, we compare GPT-4o and GPT-4o-mini. Further, we experiment with two prompt variants, the first with CoT (zero-shot-CoT) and the second without (zero-shot). In the latter, we omit the sentence \noindent {{\tt ``Start your answer with lets analyze the image.''}} from the prompt. The experiments were performed in the same fashion as in \autoref{sec:feasibility_feedback} and the results are reported in \autoref{tab:ablation_feasibility}. First, we consider the CoT variants. With $62.75\%$ success rate GPT-4o-mini performed significantly worse than the larger GPT-4o ($91.14\%$), which indicates that model size is an important factor for complex reasoning. Without CoT the performance dropped to $65.33\%$ for GPT-4o and $52.78\%$ for GPT-4o-mini. That result demonstrates that our algorithm, in fact, benefits from the CoT approach.

\begin{table}
\centering
\footnotesize
\caption[Ablation of VLMs and prompting strategies in feasibility feedback]{Ablation of VLMs and prompting strategies in feasibility feedback.}
\label{tab:ablation_feasibility}
\begin{tabular}{ccccc}
\toprule
\multirow{2}{*}{\textbf{Method}} & \multicolumn{2}{c}{\textbf{GPT-4o}} & \multicolumn{2}{c}{\textbf{GPT-4o mini}} \\
\cmidrule(lr){2-3} \cmidrule(lr){4-5}
 & Zero-shot-CoT & Zero-shot & Zero-shot-CoT & Zero-shot \\
\midrule
Success rate (\%) & \textbf{91.14} & 65.33 & 62.75 & 52.78 \\
\bottomrule
\end{tabular}
\end{table}

\subsection{Conclusion}
In this chapter, we presented Lan-grasp, a novel approach to semantic object grasping. By leveraging foundation models, we provide our approach with a deep understanding of the objects and their intended use in a zero-shot manner.
Through extensive experiments, we showed that for a wide range of objects Lan-grasp is generating grasps that are preferred by humans and also ensure safety and object usability.
In particular, the proposed grasps were compared to human preferences gathered through a questionnaire. The evaluations showed that Lan-grasp performs consistently better on that metric than the baseline methods. We also proposed a feedback loop approach that reasons about grasp feasibility in complex scenarios. 
In future work, we plan to test Lan-grasp in more complex, and cluttered environments to evaluate its robustness. Additionally, we aim to enhance the feedback loop by introducing mechanisms for when no feasible grasp is detected. For instance, the robot could ask a human for assistance or employ more sophisticated reasoning strategies to modify the environment to facilitate grasping. This would make our algorithm capable of handling more complex real-world scenarios.
Inspired by these results, in the future we plan to further exploit Large Language Models to not only decide where to grasp an object but also how to grasp and hold it according to a specific task. 
As an example, we would expect a robot operating in daily environments to hold a knife vertically and downwards when the task is to carry the knife around rather than holding the knife in a horizontal pose. This would be the next step towards more meaningful grasps that help robots with object manipulation and task execution in day-to-day environments.

In the following chapter, we transition from manipulation tasks to exploring how foundation models can enhance the reasoning capabilities of domestic robots, such as vacuum cleaners. The objective is to enable these systems to operate autonomously in real-world environments and dynamically adapt to changes in their surroundings. To realize this goal, we introduce a pipeline that integrates the semantic understanding of foundation models with knowledge distillation and continual learning strategies. This approach facilitates efficient, real-time performance, making it suitable for deployment in practical, everyday scenarios.







%% file: publications/vlmvac.tex
\chapter{VLM-Vac: Enhancing Smart Vacuums through VLM Knowledge Distillation and Language-Guided Experience Replay}
\chaptermark{VLM-Vac: Leveraging VLMs for Autonomous Vacuuming}
\label{chapter:vacuum}

The work presented in this chapter has been published in~\cite{vlmvac}:

\vspace{1cm}%
\hspace*{1cm}%
\begin{minipage}{.9\textwidth}%
R. Mirjalili, M. Krawez, F. Walter and W. Burgard.

\textbf{VLM-Vac: Enhancing Smart Vacuums through VLM Knowledge Distillation and Language-Guided Experience Replay}

\textit{IEEE International Conference on Robotics and Automation (ICRA), 2025.}


\end{minipage}%

\vspace{1cm}%

\newpage

\section*{Abstract}
\begin{adjustwidth}{1.2cm}{1.2cm} 
\small 
\textbf{
In this chapter, we propose VLM-Vac, a novel framework designed to enhance the autonomy of smart robot vacuum cleaners. Our approach integrates the zero-shot object detection capabilities of a Vision-Language Model (VLM) with a Knowledge Distillation (KD) strategy. By leveraging the VLM, the robot can categorize objects into actionable classes---either to avoid or to suck---across diverse backgrounds. However, frequently querying the VLM is computationally expensive and impractical for real-world deployment. To address this issue, we implement a KD process that gradually transfers the essential knowledge of the VLM to a smaller, more efficient model. Our real-world experiments demonstrate that this smaller model progressively learns from the VLM and requires significantly fewer queries over time. Additionally, we tackle the challenge of continual learning in dynamic home environments by exploiting a novel experience replay method based on language-guided sampling.
Our results show that this approach not only reduces energy consumption by 53\% compared to cumulative learning but also surpasses conventional vision-based clustering methods, particularly in detecting small objects across diverse backgrounds.\footnote[1]{Video available at 
\href{https://youtu.be/QmZp12rQaFU}{https://youtu.be/QmZp12rQaFU}.}}
\end{adjustwidth}

\section{Introduction}
Robotic systems are increasingly integrated into our everyday lives, playing critical roles in various domains, including manufacturing, logistics, transportation, entertainment, and home automation. Despite their broad adoption, domestic robots, in particular, face persistent challenges due to the complexity and diversity of everyday environments and tasks. While advanced algorithms and sensors enhance these robots in terms of navigation and operation, reliable real-world perception and decision-making are important research topics in domestic robotics.
Smart robot vacuum cleaners, for instance, encounter various challenges in this regard. Naively covering the entire floor area can be problematic as it does not guarantee cleaning, might spread liquid or sticky substances, and risks sucking up valuable items.
Recent advances in computer vision and machine learning have aimed to address these challenges. Data-driven methods for dirt detection have been proposed in several works ~\cite{bormann2020dirtnet, canedo2021deep, yun2022deep}. However, these approaches rely heavily on manually annotated datasets, which can be costly and limit their practical applicability in real-world scenarios.

\begin{figure}[t!]
\centering
\includegraphics[clip,trim=0cm 8cm 10cm 0cm,width=0.8\linewidth]{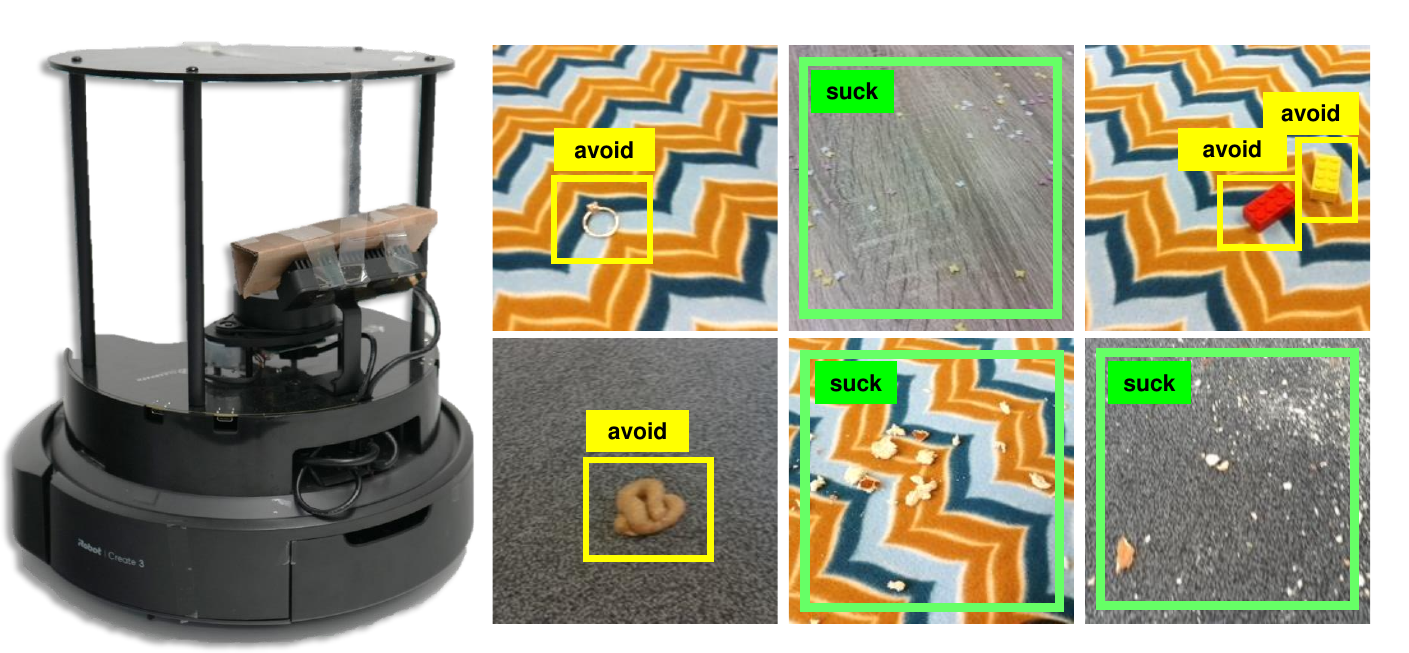}  
\caption[Overview of our smart robot vacuum cleaner system]{Overview of our smart robot vacuum cleaner system. The \mbox{TurtleBot 4} platform, shown on the left, is used in our experiments and resembles a robot vacuum cleaner. The images on the right, captured by the robot's camera, illustrate our system's real-time detection of ``suck'' or ``avoid'' actions.}
\label{fig:coverpic}
\end{figure}

Large Language Models (LLMs) and Vision-language models (VLMs) have shown impressive capabilities in image understanding and common sense reasoning, benefiting from extensive training data.
However, the practical deployment of these models is hindered by their high inference costs, whether due to energy consumption for self-hosted models or expenses related to commercial services. 
Addressing these challenges requires innovative methods that balance performance and efficiency.

In this chapter, we introduce VLM-Vac, a novel approach that leverages multimodal foundation models to enable smart vacuum cleaners to autonomously decide whether to clean or avoid an area. Our system integrates knowledge distillation (KD) with language-based continual learning to enhance functionality. Specifically, we fine-tune YOLOv8n \cite{yolov8_ultralytics} using bounding boxes labeled with actionable categories: ``suck'' or ``avoid''. These bounding boxes and labels are generated through the combination of a VLM and an open-vocabulary object detection model. During operation, when the robot encounters unfamiliar objects or flooring patterns, it queries the VLM. This new data instance is then stored and used to fine-tune the action-based object classification model. Over time, this approach reduces the number of queries to the VLM, thus improving system efficiency.

\textbf{In summary, this chapter includes the following contributions: }
\begin{enumerate} 
    \item We propose VLM-Vac, a novel framework that enhances the functionality of smart vacuum cleaners by leveraging VLMs.
    \item We implement a knowledge distillation process to transfer essential knowledge from a computationally expensive VLM to a smaller, more efficient model.
    \item We introduce language-guided experience replay for continual learning in dynamic environments.
    \item We gather a new dataset of images captured by a TurtleBot~4 robot, representing a vacuum cleaner in a household setting, and thoroughly evaluate our approach with real robot data.

\end{enumerate}

\section{Related Work}
\label{sec:related_work}
Detecting and classifying dirt on floors is an active research topic in cleaning robotics, with earlier approaches relying on saliency detection~\cite{bormann2013autonomous}, spectral analysis~\cite{milinda2017mud}, or regular pattern analysis for floor-waste separation~\cite{ramalingam2018vision}. Later works employ data-driven methods~\cite{bormann2020dirtnet, canedo2021deep, yun2022deep, singh2023vision, guan2022dirt}, with many of them utilizing a version of the YOLO-architecture. These models are trained on dedicated datasets, which are small compared to modern VLM training data sizes and thus do not support detection in an open-world scenario. Novelty detection using a CNN has been proposed as a method for floor inspection~\cite{grunwald2018optical}. However, it does not classify detected items or stains.
Recent advancements include the work by Xu~\etal~\cite{xu2024sweepmm}, who propose a multi-modal dataset tailored for household sweeping robots and use it to fine-tune an LLM for several cleaning-related downstream tasks. The model supports open-world object detection and provides cleaning recommendations according to user recommendations. However, the authors do not target model size reduction, and the model does not adapt to the environment over time.

Continual learning becomes crucial for robots functioning in dynamic real-world environments.  
Techniques such as Elastic Weight Consolidation (EWC)~\cite{kirkpatrick2017overcoming} address catastrophic forgetting by penalizing changes in model parameters that are important for the previously learned task. Peng~\mbox{\etal}~\cite{peng2023diode} observe that the performance of regularization methods for continual learning like EWC deteriorates with more new tasks learned by a model. As one possible cause, the authors identify the progressively decreasing number of free parameters that are not used for previously learned tasks. To elevate this problem, they propose to extend the model with new parameters for each new task. Liu~\mbox{\etal}~\cite{liu2020incdet} link poor performance of EWC on continual object detection to exploding gradients caused by the quadratic term in the EWC loss and missing bounding boxes of old classes in new training data. The authors propose to use Huber regularization to stabilize training and to deploy the previous model version to generate bounding boxes of old classes.
Liu~\etal~\cite{liu2020multi} address continual learning in object detection by combining a parameter regularization method with experience replay. In particular, they add a so-called Attentive Feature Distillation term to the loss function, consisting of a bottom-up and a top-down part. The bottom-up term selects and preserves parameters relevant for previous tasks. The top-down attention is computed from the intersection of ground truth bounding boxes and region proposals of the model and aims at distilling foreground objects. Finally, the authors propose a memory buffer sampling scheme which accounts for the number of bounding boxes per training image.
Several studies analyze different continual learning approaches in various scenarios.
Wang \etal~\cite{wang2021wanderlust} create a dataset and benchmark for online continual object detection. They further evaluate several continual learning baselines on the benchmark, including EWC~\cite{kirkpatrick2017overcoming} and iCaRL~\cite{rebuffi2017icarl}. The latter shows the best performance on the baseline, though lagging behind a model trained offline on all object categories. 
Kalb \etal~\cite{kalb2021continual} evaluate several continual learning baselines in the field of semantic segmentation. Their findings suggest that KD approaches that employ a previously trained model as a teacher for the new model perform better in class-incremental tasks, while replay-based methods are better suited for domain-incremental learning. In our work, we rely on experience replay, since we expect possible domain changes like newly discovered flooring patterns.
Building on these advancements, integrating language-based descriptors and experience replay methods can significantly enhance continual learning in dynamic environments. For instance, Stephan~\etal\ \cite{stephan2024text} argue that language can serve as a robust image descriptor and propose a text-guided image clustering method. Similarly, El Banani, Desai, and Jonson~\cite{el2023learning} propose sampling training image pairs based on their caption similarity which captures semantic instead of visual similarities.   

With the recent emergence of foundation models, VLMs opened the door for open-vocabulary object detection and boosted visual perception \cite{Lan-grasp}, \cite{fm-loc}.
However, the size of these models is a prohibiting factor for many applications. Therefore significant efforts~\cite{xu2024survey} were put into KD, i.e., training approaches that use a large expert model to guide the training of a small and efficient one. Gu \etal~\cite{guopen} train a Mask R-CNN architecture as an open-vocabulary detector by aligning the object proposal embeddings with the embeddings of the teacher model. KD was also used by Ma~\mbox{\etal}~\cite{ma2022open} to transfer class-level and instance-level knowledge from a teacher to an efficient network for open-vocabulary object detection.
A number of works applied KD to train a model for a specific downstream task.
Sumers~\mbox{\etal}~\cite{sumers2023distilling} leverage a VLM to label trajectories and goal states to teach a robotic agent object categories and their attributes. Yang \etal~\cite{yang2024embodied} use a pretrained LLM to guide the training of a VLM-based agent. Building upon these existing methods, our approach integrates KD with language-based continual learning to enhance the adaptability and efficiency of smart vacuum cleaner robots in diverse and dynamic environments.

\section{Approach}
\label{sec:vac_approach}
In this section, we introduce the details of VLM-Vac. Our approach consists of two main components. We first describe the distillation process from the VLM to the smaller model. We then proceed to detail our continual learning method, which utilizes language-based experience replay. 

\begin{figure*}[ht]
\centering
\vspace{5.2pt}
\includegraphics[clip,trim=0cm 3.4cm 0cm 0.6cm,width=1\linewidth]{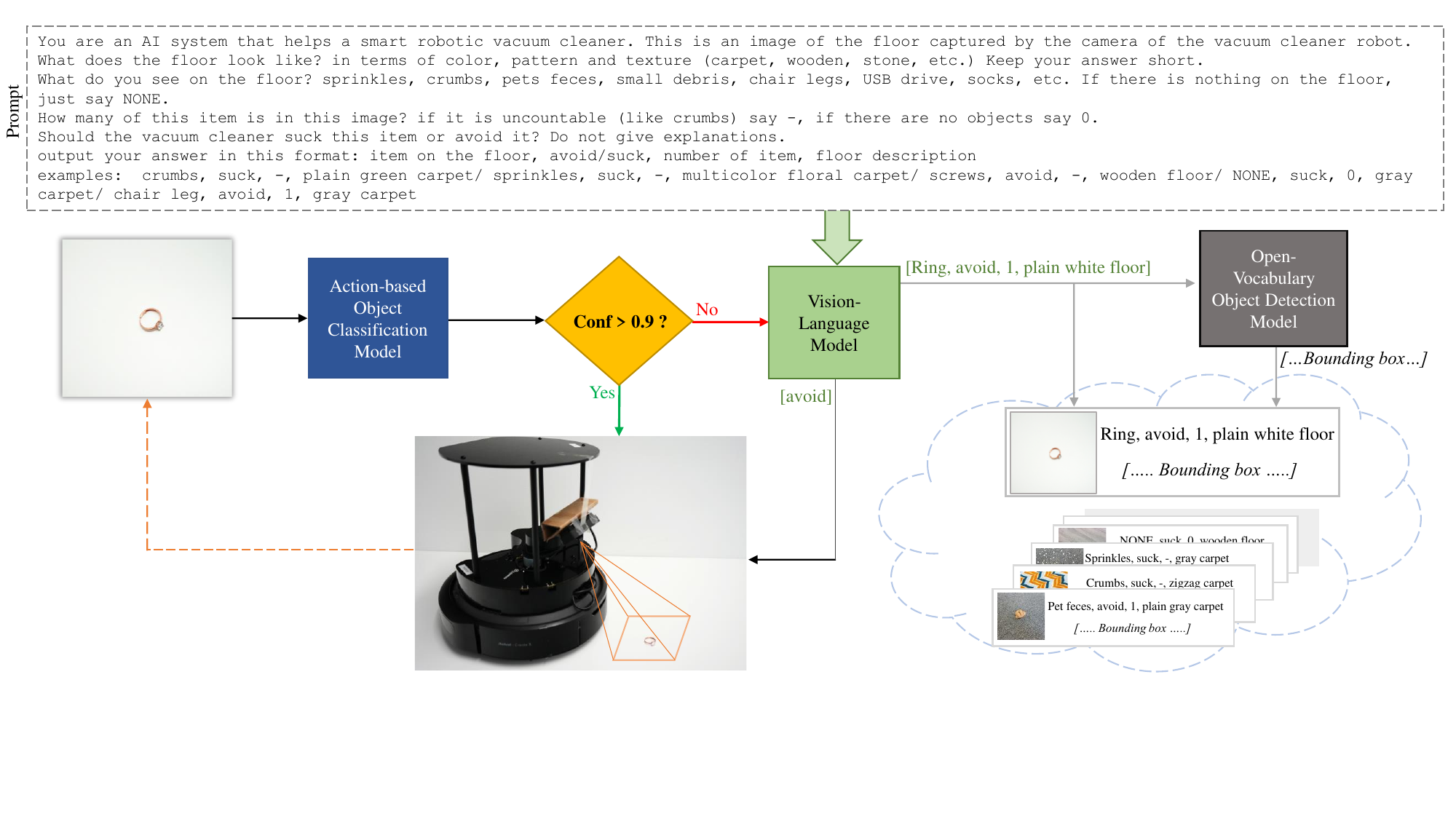}  
\caption[VLM-Vac in a nutshell]{VLM-Vac in a nutshell: We distill relevant knowledge from a Vision-Language Model (VLM) into a compact action-based object detector. The smaller model queries the VLM whenever it encounters an unfamiliar object or background. The new image, its text description from the VLM, and corresponding bounding boxes from the open vocabulary object detector are stored in the experience pool and later used for training the smaller model. Over time, the smaller model learns from these interactions, adapting to its specific environment and thus reducing the need for VLM queries. }
\label{fig:overview}
\end{figure*}

\subsection{Knowledge Distillation for Action-Based Classification}
VLMs are ideal tools to extract semantic understanding about items and flooring patterns in a zero-shot manner.
However, since continuously querying a VLM is computationally expensive and thus impractical for real-world applications, we employ a KD technique to transfer the relevant knowledge from the VLM to a smaller, more efficient object detection model, specifically YOLOv8n \cite{yolov8_ultralytics} in our approach.
Based on this technique, when YOLOv8n encounters an unfamiliar object that it cannot classify with a defined confidence threshold, it queries the VLM for guidance and gradually learns from these interactions. 
The general process, along with the prompt \(\mathrm{p}\), is depicted in \autoref{fig:overview}: The image \( I \) captured by the robot's camera is fed to the GPT-4o VLM along with the specified prompt to get the text description \(\mathrm{t}(c, q, a, f)\) which provides information on the item category \( c \), quantity \( q \), action class \( a \) and floor type \( f \):
\begin{equation}
\mathrm{t}(c, q, a, f) = \text{VLM}(\mathrm{p}, I)
\end{equation}
Additionally, we use OWL-ViT \cite{owl-vit} for grounding the item label detected by GPT-4o in the image through bounding boxes. A new image \(I^{new}\) is stored in what we call an \textit{experience pool} \( \mathcal{E} \) along with its text description \(\mathrm{t}^{new}\) and corresponding bounding boxes \( \mathrm{B}^{new} \). After each day of operation, YOLOv8n is trained on an \textit{experience buffer} \(\mathcal{B}\):
\begin{equation}
\mathcal{B}  = \{(I_1, a_1, \mathrm{B}_1),\dots, (I_N, a_N, \mathrm{B}_N) \}
\end{equation}
\(\mathcal{B}\) is sampled from the experience pool \( \mathcal{E} \) based on a selection strategy that focuses on choosing a balanced mix of image instances along with their corresponding bounding boxes and action classes. This selection process is crucial, and we will discuss it in greater detail in the following subsection.

Using this approach, the frequency of VLM queries is expected to decrease over time due to the effective distillation of knowledge into YOLOv8n. Consequently,  the vacuum cleaner will gradually adapt to its surrounding environment and, as a result, become more efficient and responsive.

\subsection{Continual Learning with Language-Based Experience Replay}
As previously discussed, a critical aspect of our approach is the data selection strategy for fine-tuning YOLOv8n. One straightforward method is \textit{cumulative training}, which involves training on all the accumulated data.
While this method exhibits a strong performance, it is computationally expensive and thus not suitable for practical resource-constrained applications like mobile robots.
An alternative approach is \textit{naive fine-tuning} of the model solely on new data, e.g. all samples acquired throughout the preceding day.
Although this is more cost-effective than cumulative learning, it presents significant challenges: Vacuum cleaner robots operate in dynamic domestic environments, where the collected data can vary significantly over time. It is, therefore, critical to retain previously learned knowledge while adapting to changes to improve over time. For instance, the robot may encounter specific items, such as pet waste, infrequently or operate in different rooms with varying flooring patterns on different days. Learning only from the most recent experiences can lead to what is known as \emph{catastrophic forgetting} \cite{mccloskey1989}, where previously acquired knowledge is eroded, posing a substantial challenge in continual learning. 

To tackle this issue, we take inspiration from Stephan~\etal~\cite{stephan2024text} and exploit a novel \textit{language-based experience replay} method. As mentioned earlier, whenever the robot encounters an unfamiliar object with low confidence score, it queries the VLM and stores the new image \( I^{new} \), along with its text description \( \mathrm{t}^{new} \), quantity \( q^{new} \), action class \( a^{new} \) and flooring pattern \( f^{new} \) in the experience pool \( \mathcal{E} \). We exploit language-based experience replay to group similar items---considering both the background and the item category---into the same cluster. 
To achieve this, we extract a language embedding \( e^{new} \) based on the item category \( c^{new} \), quantity \( q^{new} \), action class \( a^{new} \) and flooring pattern \( f^{new} \):
\begin{equation}
e^{new} = \text{Embedding}(c^{new}, q^{new}, a^{new}, f^{new})
\end{equation}
We then apply \(K\)-means clustering to group similar experiences together based on these embeddings:
\begin{equation}
\argmin_{\mathcal{C}} \sum_{k=1}^{K} \sum_{e \in \mathcal{C}_k} \|e - \mu_k\|^2
\end{equation}
where \( \mathcal{C}_k \) represents the \( k \)-th cluster, \( \mu_k \) is the centroid of \( \mathcal{C}_k \), and \( \|e - \mu_k\|^2 \) is the squared Euclidean distance between the embedding \( e \) and the cluster centroid \( \mu_k \). We then randomly select an identical number of images within each cluster, to form a balanced experience replay buffer \( \mathcal{B} \). Fine-tuning the model on this buffer \( \mathcal{B} \) ensures that the model is trained on a balanced data mix to avoid catastrophic forgetting.

\section{Experimental Results}
\label{sec:results}

In this section, we detail the experimental evaluation of VLM-Vac. Our goal is to demonstrate that within the proposed framework, a small, efficient object detection model for vacuum cleaner robots can effectively learn from a VLM and adapt to its operating environment over time without catastrophic forgetting. We start by creating a dataset of images representing a robot vacuum operating in a household setting and use this data to showcase the effectiveness of language-based clustering and its advantages over vision-based clustering. We then present the results of the KD process along with our language-based continual learning approach and analyze the energy consumption over 9 virtual days of operation with one training run per day. Finally, we report the queries made to the VLM during these days, highlighting a reduction in the number of queries.

\subsection{Setup and Dataset}

We employed a TurtleBot 4 Pro robot for our experiments, a research platform derived from the Roomba vacuum cleaning robot that is equipped with an OAK-D-PRO \mbox{RGB-D} camera. The camera angle was adjusted downward to provide a closer view of the floor, thereby enhancing the visibility and clarity of both objects and surfaces. Due to the platform's lack of a dedicated GPU, we executed our fine-tuned online detector on a workstation equipped with two NVIDIA RTX 6000 GPUs and an AMD Ryzen Threadripper PRO 64-core CPU. The YOLOv8n model, serving as the backbone for our online classifier, is reputedly capable of real-time performance on edge devices such as the NVIDIA Jetson~\cite{ultralyticsJetsonGuide}, making it a viable candidate for deployment on robot vacuum systems. Processing related to the LLM, the VLM, the open-vocabulary object detector, and the learning pipeline was carried out on the same workstation.

We used this setup to generate a dataset of 2,500 images capturing what a vacuum cleaner would typically observe in household environments. The dataset is balanced and comprises 3 distinct flooring patterns and 12 different item categories. Categories for rare items, such as a ring, were sampled from a single object instance, while more common categories, such as socks or Lego blocks, were sampled from a selection of different instances. For the repeatability of our experiments, we minimized other factors of variation, such as lighting. \autoref{fig:yolo}  shows example images from each category.

\subsection{Language-based Clustering}
To assess the effectiveness of the language-based clustering, we initiated our evaluation by randomly selecting a subset of data comprising 700 images. We utilized the \mbox{GPT-4o} VLM to obtain text descriptions and employed the \textit{text-embedding-ada-002} model for generating corresponding text embeddings. Subsequently, we applied \(K\)-means clustering to these embeddings, as detailed in \autoref{sec:vac_approach}, using visual features extracted from ResNet50.

\begin{figure*}
\centering
\vspace{5.2pt}
\includegraphics[clip,trim=0cm 6.6cm 0cm 0cm,width=1\linewidth]{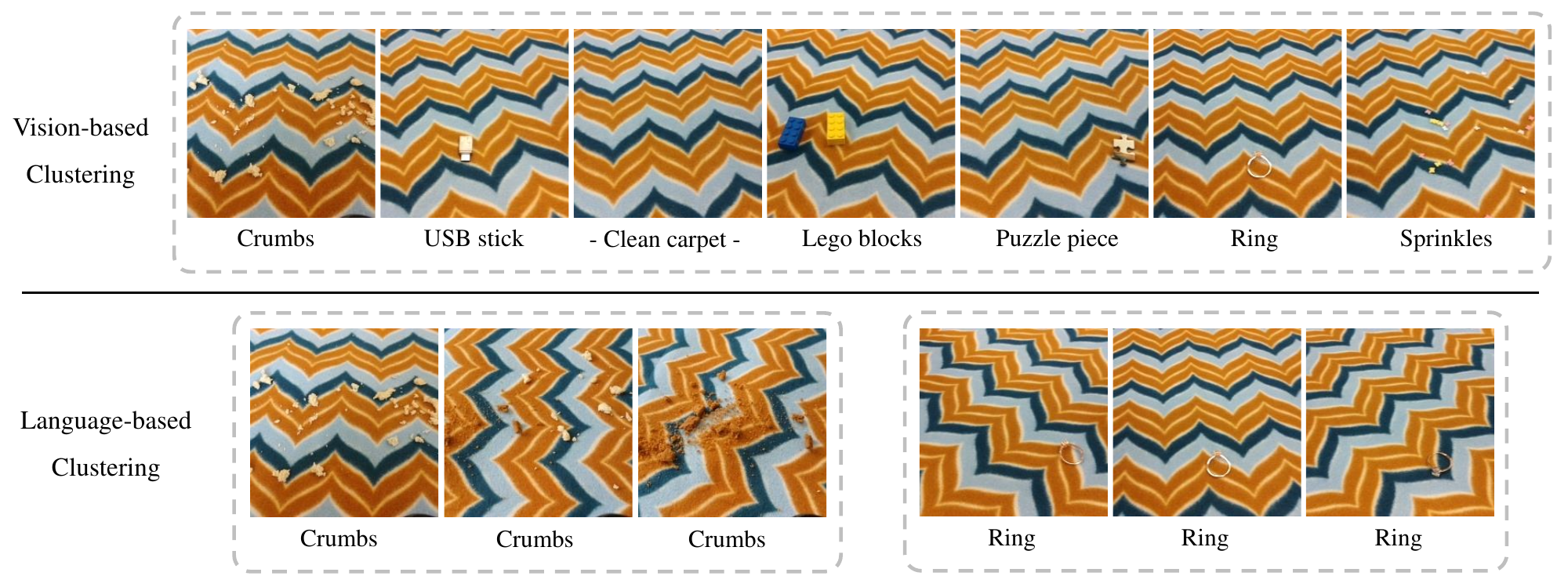}  
\caption[Sample clusters from \(K\)-means clustering]{Sample clusters from \(K\)-means clustering. The top row displays a sample cluster from the vision-based clustering approach, while the bottom row shows two sample clusters from the language-based clustering approach.}
\label{fig:clustering}
\end{figure*}

The clustering process was executed for 20 clusters, and the mean class purity was calculated for both the vision-based and language-based clustering approaches. We determined the mean class purity using a manually defined ground truth, where images were considered to belong to the same class if both the object and the background were categorized similarly. This metric allowed us to rigorously compare the performance of the two clustering methods.
The results demonstrate a clear distinction between the two approaches: the mean class purity wrt.~the ground truth for language-based sampling is $93.11\%$, which is significantly higher than the $74.12\%$ achieved by the vision-based clustering. \autoref{fig:clustering} provides examples of clusters generated by both approaches. As illustrated, the vision-based method relies heavily on background features, which leads to the incorrect grouping of small items (e.g., rings, crumbs, pet waste) into a single cluster. In contrast, language-based clustering effectively groups similar items together despite some variations in the generated labels, such as ``screws'' and ``nails'' or ``multi-color zigzag carpet'' and ``blue and orange chevron carpet''. This capability is crucial for applications like vacuuming, where the robot must differentiate between small objects (e.g., rings, crumbs) on diverse backgrounds. Based on these results, we proceed with the language-based sampling strategy for our experience replay approach.

\subsection{Knowledge Distillation and Continual Learning}
Building on these initial findings, we conducted experiments over $9$ consecutive days, with the vacuum cleaner robot visiting one of three rooms each day. Each room featured one of the three types of flooring—gray carpet, wooden floor, and zigzag-patterned carpet—and contained $7$ different item categories.
Training and clustering ran at the end of each day. In the future, this frequency could be adjusted based on the complexity of the environment.
Common household items and dirt (e.g., crumbs) were repeated in all rooms, while less common items (e.g., screws, rings, paperclips) were present in only one or two of the rooms. We randomly selected images from the categories in the dataset and distributed them across the $9$ days to ensure sufficient variation as well as repeated object occurrences. Additionally, we assumed the robot revisited each room every third day following the sequence of gray carpet, wooden floor, and zigzag carpet. This setup allowed us to investigate our algorithm's ability to adapt to new data domains while retaining previously acquired knowledge. By spacing out the visits to each room, we were able to assess how effectively the algorithm updates its model with new data without forgetting prior information—an essential factor for maintaining consistent performance in continuously changing environments.

We then conducted the experiments on the $9$ consecutive days, utilizing YOLOv8n as a student model. Each day, the model classified images into actionable categories of ``to avoid'' or ``to suck''. To minimize false detections, we set a high confidence threshold of $0.9$ for accepting the model's output, as the vacuum robot repeatedly operates in a confined area, and we, thus, expect overfitting. Following our method described in the \autoref{sec:vac_approach}, our framework queried the VLM for images where the model was uncertain, and the text descriptions, along with bounding boxes, were stored in the experience pool. Lower thresholds will decrease the number of VLM queries at the risk of undesired robot behavior.

We used GPT-4o as the VLM for generating image text descriptions and \mbox{OWL-ViT} for generating the bounding boxes. We engineered the prompts to maximize the accuracy of the models' predictions. To provide the VLM with more context, we included samples of common household objects in the prompt and found that the results improve when asking the model to first output the object categories before the actual action class. Setting the temperature of GPT-4o to zero ensured repeatability.

To benchmark our approach, we compared it against two baseline methods: cumulative learning and naive fine-tuning. In the cumulative learning approach, the YOLOv8n model was trained from scratch each day using all the data accumulated in the experience pool up to that point. For naive fine-tuning, the YOLOv8n model was trained only on the data acquired from the previous day, but with a warm start from earlier training sessions.

In our evaluations, we ignore the bounding box information returned by YOLOv8n and, thus, are left with a binary classification task. Accordingly, we focused on the $F_1$ score, which offers a balanced assessment of both precision and recall. Given that our emphasis is more on accurate classification than on the precision of the bounding boxes, the $F_1$ score provides a reliable metric for evaluating the overall performance of the model. It is computed on each day on the new data of that day before training.

\begin{figure}[t]
\centering
\includegraphics[clip,trim=0cm 0cm 0cm 0cm,width=0.8\linewidth]{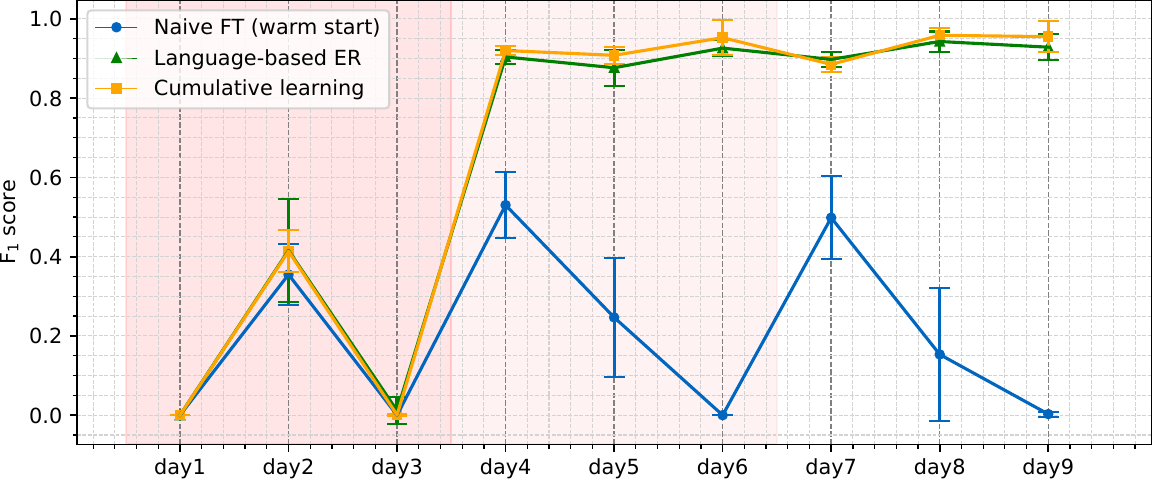}  
\caption[The $F_1$ score for naive fine-tuning, language-based experience replay and cumulative learning]{The $F_1$ score for naive fine-tuning, language-based experience replay and cumulative learning across $9$ consecutive days averaged over 10 runs with different random seeds. Background shadings represent 3-day intervals. The bars indicate standard deviations.}
\label{fig:F1}
\end{figure}

\autoref{fig:F1} presents the $F_1$ scores for the three methods averaged over 10 runs. We trained the model for $100$ epochs on each day. As can be seen, naive fine-tuning suffers from catastrophic forgetting, with performance periodically dropping. This behavior is expected as the robot encounters rooms with drastically different flooring backgrounds and items over different days. However, both the language-based experience replay and cumulative learning approaches demonstrate similar performance trends, with rapid improvement after the robot has visited each room once.

\begin{figure}
\centering
\includegraphics[clip,trim=0cm 0cm 0cm 0cm,width=0.8\linewidth]{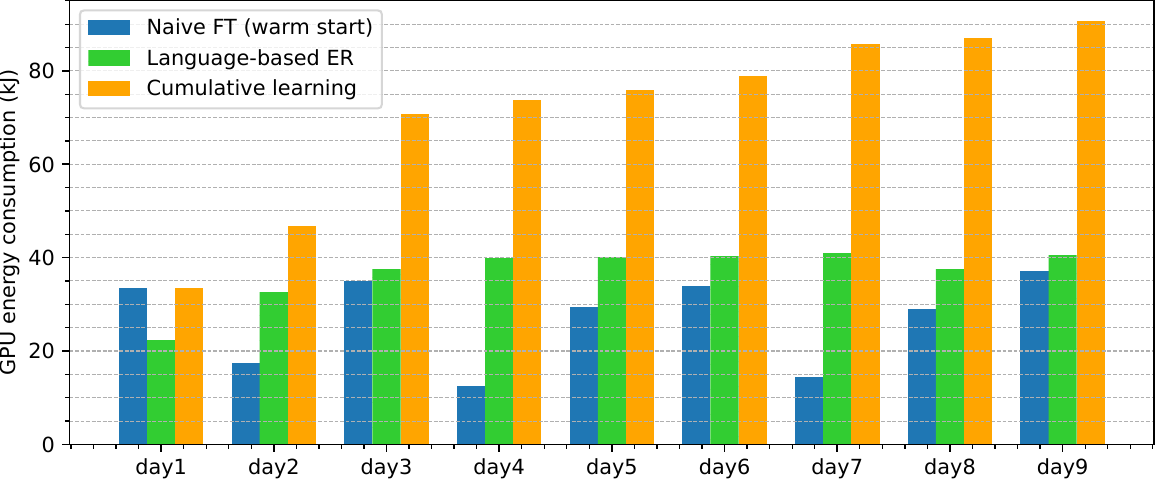}  
\caption[GPU energy consumption during training for naive fine-tuning, language-based experience replay and cumulative learning]{GPU energy consumption during training for naive fine-tuning, language-based experience replay and cumulative learning.}
\label{fig:energy}
\end{figure}

\begin{table}[t]
\centering
\footnotesize
\caption[Average mean $F_1$ score and GPU energy consumption]{Average mean $F_1$ score and GPU energy consumption.}
\label{table:summary}
\begin{tabular}{lcc}
\toprule
\textbf{Method} & \textbf{Mean $F_1$ Score} & \textbf{Mean GPU Energy (kJ)} \\
\midrule
Naive fine-tuning (warm start)       & 0.239          & \textbf{26.1} \\
Language-based experience replay     & \textbf{0.913} & \textbf{39.3} \\
Cumulative learning                  & \textbf{0.930} & 83.6 \\
\bottomrule
\end{tabular}
\end{table}

Next, we analyzed the GPU energy consumption based on the power draw reported by the GPU driver. The results in \autoref{fig:energy} indicate energy usage for each day across the different methods. While the reported numbers are specific to our hardware setup, the relative differences in energy consumption are expected to apply also to other platforms. As illustrated in the figure, cumulative learning consumes considerably more energy due to the increasing size of the training data each day. The mean $F_1$ performance score and GPU energy consumption from day $4$ onward over the 10 runs are presented in \autoref{table:summary}. As can be seen, the language-based approach results in an $F_1$ score of $0.913$ comparable to $0.930$ achieved by cumulative learning but reduces energy consumption by $53\%$.

\begin{figure}
\centering
\vspace{5.2pt}
\includegraphics[clip,trim=0cm 3.5cm 0cm 0cm,width=1\linewidth]{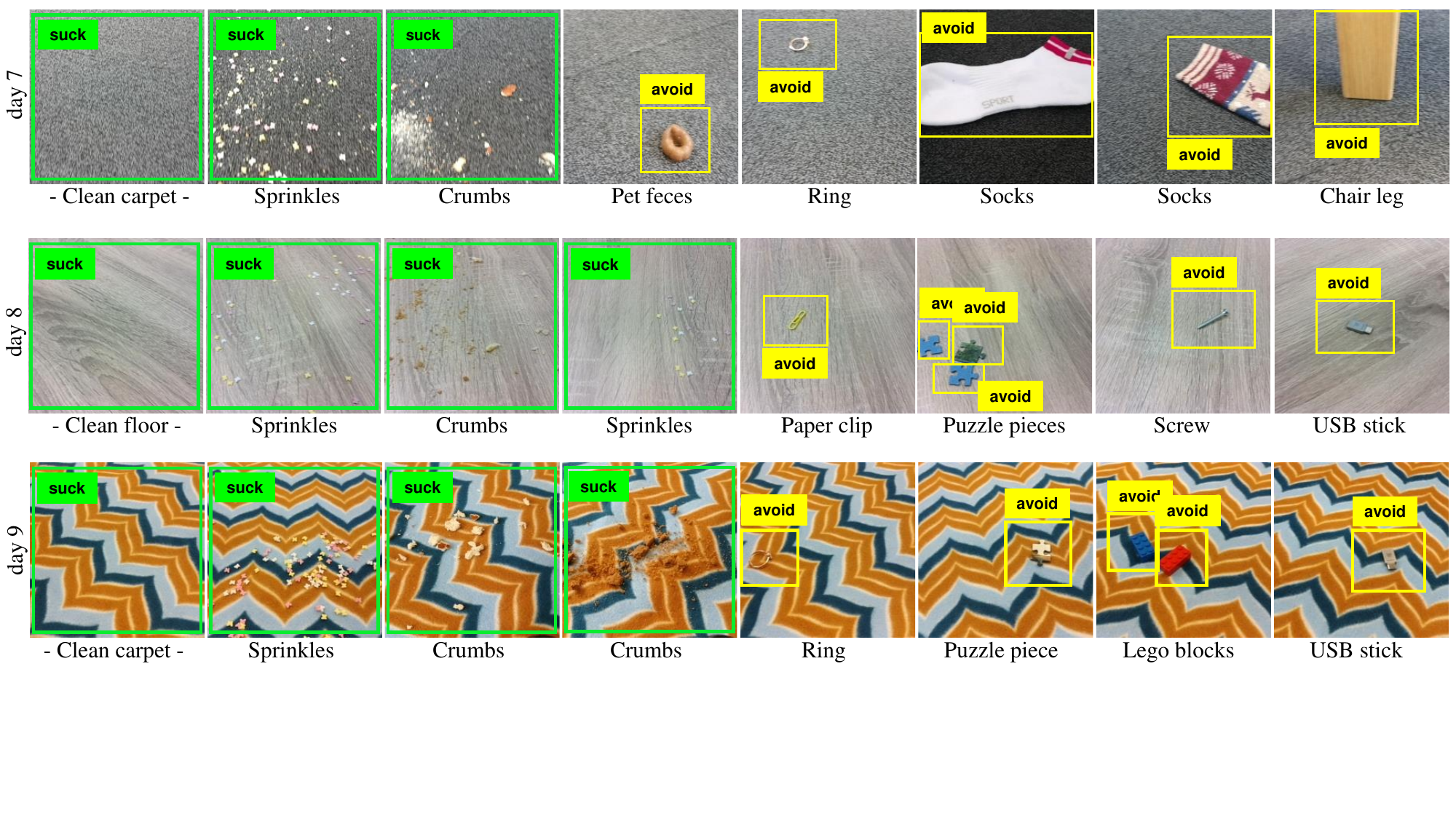}  
\caption[Results of action-based object classification with fine-tuned YOLOv8n on three consecutive days]{Results of action-based object classification with fine-tuned YOLOv8n on days $7$, $8$, and $9$. The model effectively classifies small objects even on complex backgrounds. To improve visualization, we enlarged the bounding boxes.}
\label{fig:yolo}
\end{figure}

\begin{figure}
\centering
\vspace{5.2pt}
\includegraphics[clip,trim=0cm 0cm 0cm 0cm,width=0.8\linewidth]{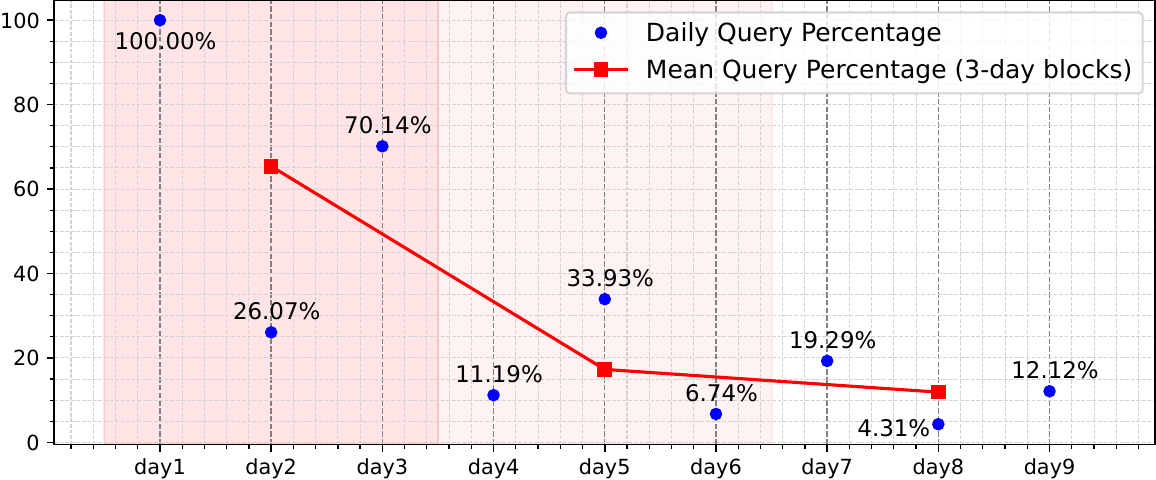}  
\caption[Percentage of images queried to the VLM over nine consecutive days, with the mean query percentage indicated for each three-day interval]{Percentage of images queried to the VLM over nine consecutive days, with the mean query percentage indicated for each three-day interval.}
\label{fig:queries}
\end{figure}

Finally, the object classification results for days $7$, $8$, and $9$, using the language-based experience replay approach, are depicted in \autoref{fig:yolo}. The bounding boxes show the class categories that the fine-tuned YOLO model detected for each item. As shown in the figure, our method effectively performs object detection, even for small items and on complex flooring patterns. Additionally, \autoref{fig:queries} illustrates the percentage of images queried to the VLM each day using the language-based experience replay method. The red line represents the mean query percentage over three-day blocks. The decreasing trend in the mean query percentage supports our claim that knowledge from the VLM is gradually being distilled into the smaller model, YOLOv8n.

\newpage
\section{Conclusion}
In this chapter, we proposed VLM-Vac, a novel framework for enhancing the capabilities of smart vacuums in dynamic real-world environments. By leveraging knowledge distillation, we transfer the knowledge from a VLM to a smaller, more efficient model, increasing the system's efficiency. We also introduce a novel language-based continual learning approach to mitigate catastrophic forgetting.

One interesting direction for future work is to investigate the long-term performance of our approach. According to recent studies \cite{plasticity}, the capacity of deep learning models declines when trained on new data over extended periods. Wang~\etal~\cite{plasticity}, proposed to tackle this phenomenon by continuously reinitializing a fraction of the less-active neurons. They report that this approach helps models to indefinitely maintain their plasticity and keep learning from new streams of data. This is particularly relevant to our approach, as a robot vacuum cleaner needs to operate in dynamic environments and continuously learn from new data. 

Another potential approach for future work is to deploy an out-of-distribution detection algorithm in our framework. This could prevent the vacuum cleaner from making confident wrong guesses when faced with unfamiliar objects or backgrounds that deviate from its training data. Finally, it would be interesting to include a broader range of objects and backgrounds in the experiments, as well as to conduct them over a longer period of time. This expansion could help provide more insights about the performance of our algorithm when faced with diverse real-world scenarios.

While this chapter focused on enabling smart vacuums to generalize across time and space, many of the challenges addressed—such as robustness to visual variation—are equally critical for manipulation tasks. In the next chapter, we introduce ARRO, a method that equips robots with an augmented view of the world by filtering out visual distractions. ARRO enhances visuomotor policy robustness under domain shifts by masking irrelevant parts of the scene and preserving only task-relevant visual information. This approach continues our exploration of using foundation models to make robotic systems more generalizable and robust in diverse and unstructured environments.

\label{sec:conclusion}



%% file: publications/ARRO.tex
\chapter{Augmented Reality for RObots (ARRO): Pointing Visuomotor Policies Towards Visual Robustness}
\chaptermark{ARRO: Augmented reality for Robots}
\label{chapter:arro}

The work presented in this chapter has been published in~\cite{mirjalili2025augmented}: 

\vspace{1cm}%
\hspace*{1cm}%
\begin{minipage}{.9\textwidth}%
R. Mirjalili\textsuperscript{*}, T. Jülg\textsuperscript{*}, F. Walter and W. Burgard

\textbf{Augmented Reality for RObots (ARRO): Pointing Visuomotor Policies Towards Visual Robustness}

\textsuperscript{*}Equal contribution

\textit{arXiv preprint} \href{https://arxiv.org/abs/2505.08627}{arXiv:2505.08627}

\end{minipage}%

\vspace{1cm}%

\newpage

\section*{Abstract}
\begin{adjustwidth}{1.2cm}{1.2cm} 
\small 
\textbf{
Recent advances in visuomotor policy learning show that training on human demonstrations enables models to perform effectively across a range of manipulation tasks. However, these policies often face difficulties when exposed to visual domain shifts—such as changes in background appearance or differences in robot embodiment—which limits their ability to generalize. To address this challenge, we introduce ARRO, a calibration-free visual preprocessing framework that improves robustness without requiring any additional training. ARRO leverages zero-shot open-vocabulary segmentation and object detection to isolate task-relevant elements, effectively suppressing irrelevant visual content. ARRO can be seamlessly integrated into both training and inference pipelines by masking distractors and introducing structured visual cues. This selective filtering enhances policy generalization while reducing the need for extensive data collection. We evaluate ARRO through comprehensive experiments using Diffusion Policy across a range of tabletop manipulation tasks in both simulated and real-world environments, and also demonstrate its compatibility with generalist models such as Octo and OpenVLA. ARRO consistently improves performance, supports targeted object-level masking, and remains effective even under challenging segmentation conditions.\footnote[1]{Videos illustrating our results available at: 
\href{https://augmented-reality-for-robots.github.io/}{augmented-reality-for-robots.github.io}}}

\end{adjustwidth}

\section{Introduction}

\begin{figure}[h]
\centering
\includegraphics[clip,trim=0cm 0cm 0cm 0cm,width=1\linewidth]{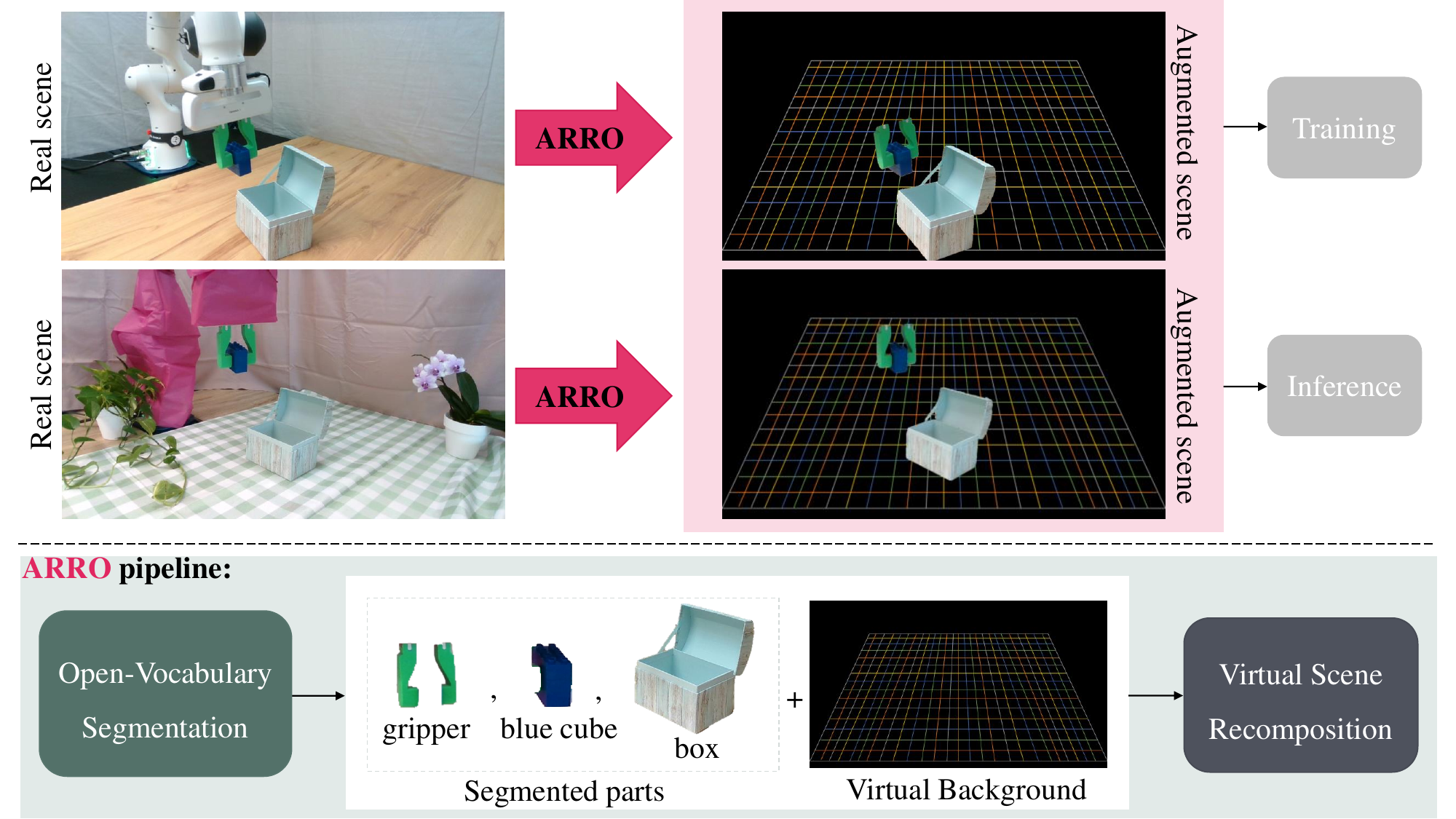}  
\caption[ARRO in a nutshell]{ARRO in a nutshell: Our approach employs open-vocabulary segmentation to isolate the robot gripper and task-relevant objects, which are then overlayed onto a consistent virtual background. This visual preprocessing is applied at both training and deployment stages, ensuring robustness to environmental variations and promoting stable policy behavior across diverse visual conditions.}
\label{fig:coverpic}
\end{figure}

Achieving autonomous operation in human-centered environments requires robots to effectively perceive their surroundings and translate visual information into coordinated motor actions. Visuomotor policy learning tackles this challenge by establishing a direct mapping from visual perception to control, enabling robots to carry out complex tasks in diverse and dynamic conditions. Recent years have seen significant progress in this area, particularly through the adoption of generative models, which have demonstrated impressive performance in learning scenarios~\citep{Vaswani.2017, Ho.2020, Dosovitskiy.2021}. Among these developments, imitation learning—where robots learn from expert human demonstrations—has proven especially effective across a variety of robotic systems, including both manipulators and mobile platforms~\citep{visuomotordiffusion, viola, dalal2023imitating}. These advances have been accompanied by the growth of large-scale robotics datasets~\citep{Walke.2023, Fu.2024, Bharadhwaj.2024, Fang.2024, droiddataset} and a growing emphasis on standardization and open collaboration within the research community~\cite{openx}. As a result, there is increasing interest in developing general-purpose visuomotor policies—often referred to as robotics foundation models—that aim to integrate perception and control across a broad range of environments, tasks, and robotic embodiments~\citep{openx, rt1, rt2, octo, openvla}.

Despite notable progress, a fundamental obstacle in deploying visuomotor policies lies in their limited robustness beyond the training distribution. Ideally, these policies should generalize reliably across variations in visual context.
In practice, however, real-world deployment frequently introduces domain shifts that were not encountered during training. Even seemingly minor changes—such as differences in background, the inclusion of distractor objects, or visual variations in the robot appearance—can lead to substantial performance degradation~\cite{decomposegengap, Li.2024}. This fragility poses a significant barrier to real-world applicability, where environmental variability is both routine and unpredictable. While increasing the diversity of the training dataset can help mitigate these issues, it is often prohibitively expensive and rarely sufficient to fully overcome the generalization gap.

To overcome these challenges, this work introduces an alternative approach inspired by recent progress in foundation models within computer vision. Rather than training visuomotor policies to directly accommodate the full diversity of visual environments, we propose a transformation of the input space into a standardized, task-relevant representation. By explicitly filtering out scene elements that are irrelevant to the manipulation task, this method improves policy robustness. 
This conceptual shift gives rise to the central question that guides our work:

\begin{quote}
\textit{Is it possible to equip robots with a task-aware perceptual interface—an augmented reality view that abstracts away distractors and selectively retains the visual information most critical for successful task execution?}
\end{quote}

This chapter introduces \textbf{ARRO} (\textbf{A}ugmented \textbf{R}eality for \textbf{RO}bots), a visual preprocessing framework that acts like \textit{augmented reality glasses} for robots—suppressing irrelevant scene elements, emphasizing key task components, and improving policy generalization without additional training. ARRO operates without the need for camera calibration and generates task-focused visual inputs by leveraging open-vocabulary segmentation and object detection. It isolates the robot’s gripper and relevant objects, removing irrelevant content and overlays the retained elements on a structured, virtual background. This transformation creates a consistent observation space that enhances the robustness of visuomotor policies across a wide range of environments and robotic platforms.

\textbf{In summary, this chapter includes the following contributions:} 

\begin{enumerate}
\item We propose ARRO, a calibration-free augmented reality preprocessing framework designed to enhance the visuomotor robustness. ARRO selectively retains only the task-relevant elements—specifically the robot gripper and target objects—while masking out irrelevant or distracting components from the visual scene.

\item ARRO integrates foundation models for open-vocabulary segmentation and object detection. Operating in a zero-shot manner, ARRO eliminates the need for camera calibration or task-specific training and remains broadly compatible with diverse visuomotor policy architectures.

\item We incorporate a structured virtual background with spatially consistent visual cues into the augmented observations. Our experiments show that this background design improves performance over simpler alternatives that lack visual cues.

\item We conduct extensive evaluations of ARRO across a range of manipulation tasks and policy backbones, including Diffusion Policy~\citep{visuomotordiffusion}, Octo~\citep{octo}, and OpenVLA~\citep{openvla}. Results consistently demonstrate improved robustness across changes in background appearance, robot embodiment, and the presence of visual distractors.
\end{enumerate}


\section{Related Work}
\label{sec:related}

Addressing domain shift has been a persistent challenge in visuomotor policy learning, particularly within imitation learning frameworks. One of the classic strategies for bridging the sim-to-real gap is domain randomization, which exposes the policy to diverse environmental variations during training~\citep{Tobin.2017, chebotar2019closing}. Imitation learning has increasingly leaned on large-scale datasets to improve generalization across heterogeneous robots and tasks. The Open X-Embodiment dataset~\citep{openx}, for example, integrates demonstrations from many robotic systems, enabling the training of versatile policies capable of operating across embodiments~\citep{octo, openvla}. However, collecting such large datasets remains expensive and time-consuming. To address this, some approaches incorporate human video demonstrations~\citep{chen2021learning, xiong2021learning, phantom} as a lower-cost alternative. Others mitigate embodiment variability by mounting cameras on the robot itself~\citep{polybot, umi}, ensuring a consistent egocentric view. However, such approaches cannot leverage abundant third-person data and may struggle with tasks requiring a broader understanding of the scene. By contrast, ARRO works directly on third-person observations and eliminates the need for data collection or policy adaptation during deployment.

Creating visual representations that remain stable under domain shift is a key concern in the development of robust visuomotor policies. ARRO follows a similar motivation to Object-centric methods~\citep{Devin.2018, viola}—focusing on what is essential in the scene—but achieves this directly in image space without additional training.
Other methods, like Transporter Networks~\citep{Zeng.2021}, further extend object-centricity by including language conditioning for actions~\citep{Shridhar.2022}. 
A separate class of techniques encodes manipulation-relevant features through learned or predefined keypoints~\citep{Sieb.2020, Manuelli.2022}, with recent work using foundation models to extract such features automatically~\citep{Huang.2024}. 
Meanwhile, vision-language approaches have shown that explicit keypoints are not always necessary; visual cues within the camera view can be sufficient for reasoning~\citep{Nasiriany.2024, fangandliu2024moka}. Keypoint-based methods, however, tend to assume fixed task structures and can lack flexibility. Broader generalization is often pursued through representation learning on large-scale datasets—e.g., by incorporating human videos~\citep{Nair.2022, Karamcheti.2023} or 3D data streams~\citep{Chen.2024}.

An increasing number of approaches aim to modify raw visual inputs to enhance policy robustness, either by augmenting training datasets or adapting observations at test time. Several methods~\citep{cacti, rosie, genaug, roboagent, genima, rovi-aug} leverage generative models to create synthetic variations of scenes, backgrounds, or robot embodiments during training. Other approaches focus on transforming real-world observations during deployment to better align with a known training distribution. For example, Zhang et al.~\cite{vr-goggles} apply style transfer to re-render camera frames in the visual domain of a simulation environment, facilitating sim-to-real policy transfer. Mirage~\citep{mirage} supports cross-embodiment transfer through a multi-step pipeline involving inpainting and overlaying rendered robot images, but depends on accurate URDF models and camera calibration, and is not suited for background variation. SHADOW~\citep{shadow} similarly overlays segmented robot masks to enable embodiment transfer but assumes fixed backgrounds, requires precise calibration, and necessitates retraining for each new embodiment. 
RoVi-Aug~\citep{rovi-aug} enriches training datasets by leveraging fine-tuned diffusion models to generate novel robot embodiments and alternative viewpoints. However, it assumes fixed backgrounds and its effectiveness depends on access to paired training data and requires additional trainings.

In contrast to these techniques, ARRO performs direct preprocessing of real camera observations by filtering out irrelevant visual content through open-vocabulary segmentation. It operates without requiring calibration, retraining, or additional data collection, yet achieves strong robustness to both background and embodiment variations.

\section{Approach}
\label{sec:approach}
\subsubsection{Problem Setup}

We study a visuomotor policy learning setup in which a robot observes a sequence of RGB frames \( I_{t-T_o}, \ldots, I_t \) and produces a sequence of actions \( a_t, \ldots, a_{t+T_a} \) using the conditional policy \( \pi(a_{t:t+T_a} \mid I_{t-T_o:t}) \).
\( T_o \) and \( T_a \) represent the length of observation and action horizon respectively.
The model is trained on a dataset \( \mathcal{D}_{\text{train}} \), composed of image-action pairs \((I_t, a_t)\) collected from expert rollouts in a source environment. Despite strong performance within this domain, visuomotor policies often degrade when transferred to new deployment settings \( \mathcal{D}_{\text{test}} \), where unseen visual distractors, background variations, or embodiment variations introduce domain shifts.

To address this, we introduce a calibration-free transformation \( \Phi(I_t) = \tilde{I}_t \), which maps the raw image \( I_t \) to an augmented view \( \tilde{I}_t \) that contains only the elements required for the task. 
Our transformation relies on recent advances in vision-language and segmentation models leveraging their zero-shot object detection and segmentation capabilities without requiring camera calibration or task-specific finetuning. As a result, the system generalizes effectively across environments, requiring little to no setup adjustment.

\subsubsection{Open-Vocabulary Visual Filtering}

To extract the robot gripper and task-relevant objects from a sequence, we adopt a two-step segmentation strategy. This approach begins with identifying and segmenting task-relevant parts in the initial frame, followed by the propagation of segmentation masks across subsequent frames to maintain temporal coherence. Specifically, at the beginning of each episode, we use an open-vocabulary object detection model—such as Grounding DINO~\citep{groundingdino}—prompted with a textual prompt corresponding to the target class \(c\). This produces an initial bounding box \(B_0\) on the first frame \(I_0\):

\begin{equation}
    B_0 = \text{Detect}(I_0, c)
\end{equation}
This bounding box \( B_0 \) is used to prompt a segmentation model (e.g., SAM 2~\citep{sam2}) to extract the object segmentation.

Unlike standard objects, robot grippers are often poorly represented in pretrained object detectors. Thus, we need to take another approach for extracting the gripper segmentation. 
In this regard, we apply class-agnostic segmentation on \( I_0 \) to generate candidate masks, using models such as SAM~\citep{sam}, to extract generic object proposals without requiring predefined categories:

\begin{equation}
    \{ S_0^i \}_{i=1}^{N} = \text{Segment}(I_0)
\end{equation}

Each candidate mask region is then visually annotated by placing numeric identifiers at its center, producing a modified image \( I_0^* \). This annotated image alonside a textual prompt is passed to a vision-language model such as GPT-4o~\citep{gpt-4o}, which returns the annotations corresponding to the gripper fingers:

\begin{equation}
    g_0 = \text{VLM}(I_0^*)
\end{equation}
where  \( g_0 = \{ (x_\ell, y_\ell), (x_r, y_r) \} \) denotes the identified keypoints corresponding to the left and right fingers of the robot gripper, as detected in the initial frame. This method also proves effective for objects with simple geometric structures—such as plain cubes—that can be precisely segmented without relying on bounding box annotations.

We then provide the object bounding box \( B_0 \) and the gripper keypoints \( g_0 \) as prompts to a memory-based segmentation model(e.g., SAM 2~\citep{sam2}), which outputs the corresponding segmented regions:

\begin{equation}
    S_0^{\text{obj}}, \quad S_0^{\text{gripper}} = \text{Segment}(I_0 \mid B_0, g_0)
\end{equation}

after establishing the initial segmented regions, segmentation is propagated forward over time, producing the binary segmentation masks \( S_t^{\text{obj}} \) and \( S_t^{\text{gripper}} \) for subsequent frames \(I_t\) for \(t > 0\), by maintaining and updating temporal memory from earlier frames ~\citep{sam2}.
This yields consistent tracking of relevant segmentation regions without additional prompts or supervision.

\subsubsection{Virtual Scene Reconstruction}

After obtaining the segmentation masks \( S_t^{\text{obj}} \) and \( S_t^{\text{gripper}} \), we take their union to define the complete task-relevant region:
\[
    S_t = S_t^{\text{obj}} \cup S_t^{\text{gripper}}
\]

Next, we overlay \( S_t \) onto a fixed background image \( I_B \).
We choose this background image \( I_B \) either as a plain black canvas or a hand-crafted grid, which is consistently reused across all sequences to ensure visual consistency.
We then compute the final augmented image, \( \tilde{I}_t \), as:

\begin{equation}
    \tilde{I}_t = S_t \odot I_t + (1 - S_t) \odot I_B
\end{equation}
where \( \odot \) represents element-wise multiplication.
This transformation is applied uniformly to training and test data. The pipeline runs in real time during inference and requires no task-specific tuning. This operation selectively preserves task-relevant regions of the image, while all other pixels are replaced by the structured virtual background, promoting consistency and reducing irrelevant visual variation.

The following pseudocodes (Algorithm~\ref{alg:arro_init} and Algorithm~\ref{alg:arro_mask}) present the summary of our pipeline.

\begin{algorithm}
\caption{ARRO Initialization}
\label{alg:arro_init}
\textbf{Input:}
\begin{tabular}[t]{@{}ll}
$ I $            & An RGB frame \\
$ p^o_1, \dots, p^o_n $ & Object prompts \\
$ p^t $          & A task prompt \\
\texttt{Detect}  & Open-vocabulary object detector (e.g., \texttt{GroundingDINO}) \\
\texttt{Segment} & Uninitialized segmentation model (e.g., \texttt{SAM2}) \\
\texttt{VLM}     & Vision-language model (e.g., \texttt{GPT-4o}) \\
\end{tabular}

\textbf{Output:} The initialized segmentation model $\texttt{Segment}_{0}$ \vspace{0.7em}

\begin{algorithmic}
\State \textit{// Get object bounding boxes for complex objects}
\vspace{0.3em}
\For{$p^o_i$ in $\{ p^o_1, \ldots, p^o_n \}$}
    \State $B_i = \texttt{Detect}(I, p^o_i)$
\EndFor
\vspace{0.7em}
\State \textit{// Get region masks to segment simple objects and gripper fingers}
\vspace{0.3em}
\State Run unprompted segmentation on $I$ to get region masks: $K_i \leftarrow \texttt{Segment}(I)$
\State Annotate keypoints in $I$ with numeric labels to retrieve $I^*$
\State Identify task-relevant keypoints in $I^*$: $\{K_0, \ldots, K_m\} \leftarrow \texttt{VLM}(I^*, p^t)$
\vspace{0.7em}
\State \textit{// Initialize and return the segmentation model}
\vspace{0.3em}
\State \Return $\texttt{Initialize}(\texttt{Segment}, I, B_0, \ldots, B_n, K_0, \ldots, K_m)$
\end{algorithmic}
\end{algorithm}

\begin{algorithm}
\caption{ARRO Masking}
\label{alg:arro_mask}
\textbf{Input:}
\begin{tabular}[t]{@{}ll}
$ I $           & RGB frame \\
$ I_B $         & RGB background image \\
$\texttt{Segment}_{t}$ & Initialized segmentation model (e.g., \texttt{SAM2}) \\
\end{tabular}

\textbf{Output:} A masked RGB frame $\tilde{I}$ and the updated segmentation model $\texttt{Segment}_{t+1}$
\vspace{0.7em}
\begin{algorithmic}
\State Track object and gripper masks: $S^{\text{obj}}, S^{\text{gripper}}, \texttt{Segment}_{t+1}(I) \leftarrow \texttt{Segment}_{t}(I)$
\State Combine masks: $S = S^{\text{obj}} \cup S^{\text{gripper}}$
\State Overlay $S$ on virtual background $I_B$ to recompose the scene into $\tilde{I}$
\vspace{0.7em}
\State \Return $\tilde{I}$, $\texttt{Segment}_{t+1}$
\end{algorithmic}
\end{algorithm}

\section{Experimental Results}
\label{sec:experiments}

\subsection{Real-World Experiments}

This section provides an in-depth empirical analysis of ARRO’s performance across a range of real-world robotic manipulation tasks. The central aim is to evaluate how effectively ARRO enhances the robustness of visuomotor policies when exposed to visual domain shifts and distractor objects. We evaluate both task-specific and generalist models under a unified evaluation framework. All experiments were carried out using a Franka Emika Research 3 (FR3) robotic arm, equipped with custom-designed 3D-printed gripper fingers, as depicted in Figure~\ref{fig:thumbnail}. The robot executed actions in a closed feedback loop, relying solely on RGB observations captured from a fixed third-person camera.

\begin{figure}[t]
\centering
\includegraphics[clip,trim=0cm 0cm 0cm 0cm,width=0.9\linewidth]{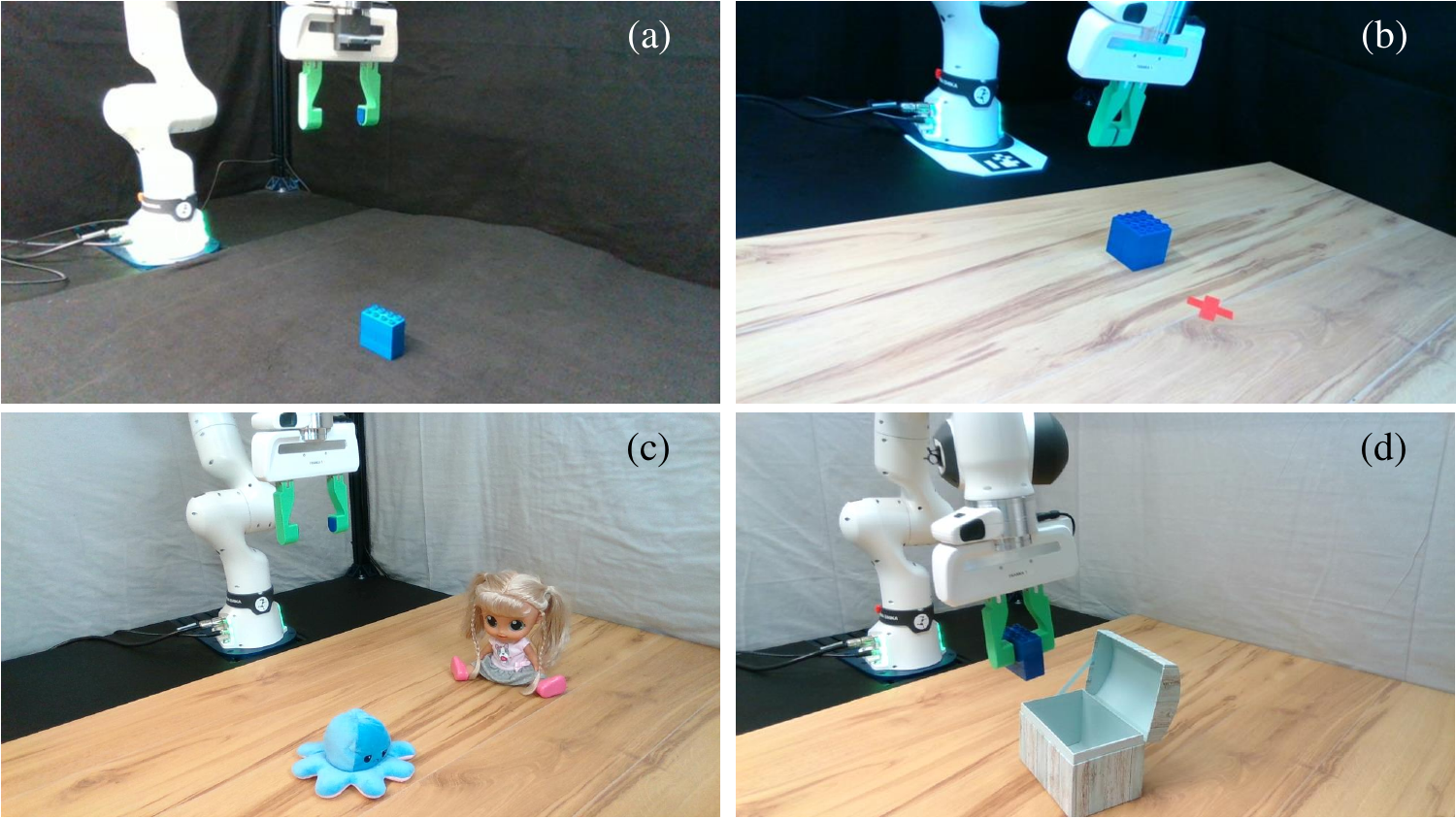}  
\caption[Tabletop manipulation task setups used in our real-world experiments]{Visual setups for (a) pick-v1, (b) push-v1, (c) doll-v1, and (d) box-v1 tasks used in our experiments.}
\label{fig:thumbnail}

\label{fig:thumbnail}
\end{figure}

\subsubsection{Robustness to Visual Domain Shifts}

This section investigates ARRO’s capacity to counteract the performance degradation often observed when visuomotor policies are deployed in unfamiliar visual settings. 
We evaluate ARRO on four tabletop manipulation tasks as depicted in Figure~\ref{fig:thumbnail}. We designed each task to reflect a different manipulation challenge under visually diverse conditions. These include: (a) picking up a single cube from the table (pick-v1), (b) pushing a cube to a red cross marker (push-v1), (c) placing a plush octopus adjacent to a doll (doll-v1), and (d) dropping a cube into a box and closing the lid (box-v1). We gather 90 human-provided demonstrations for each task, to facilitate both policy training and performance assessment.

\begin{figure}
\centering
\includegraphics[clip,trim=0cm 0cm 0cm 0cm,width=0.8\linewidth]{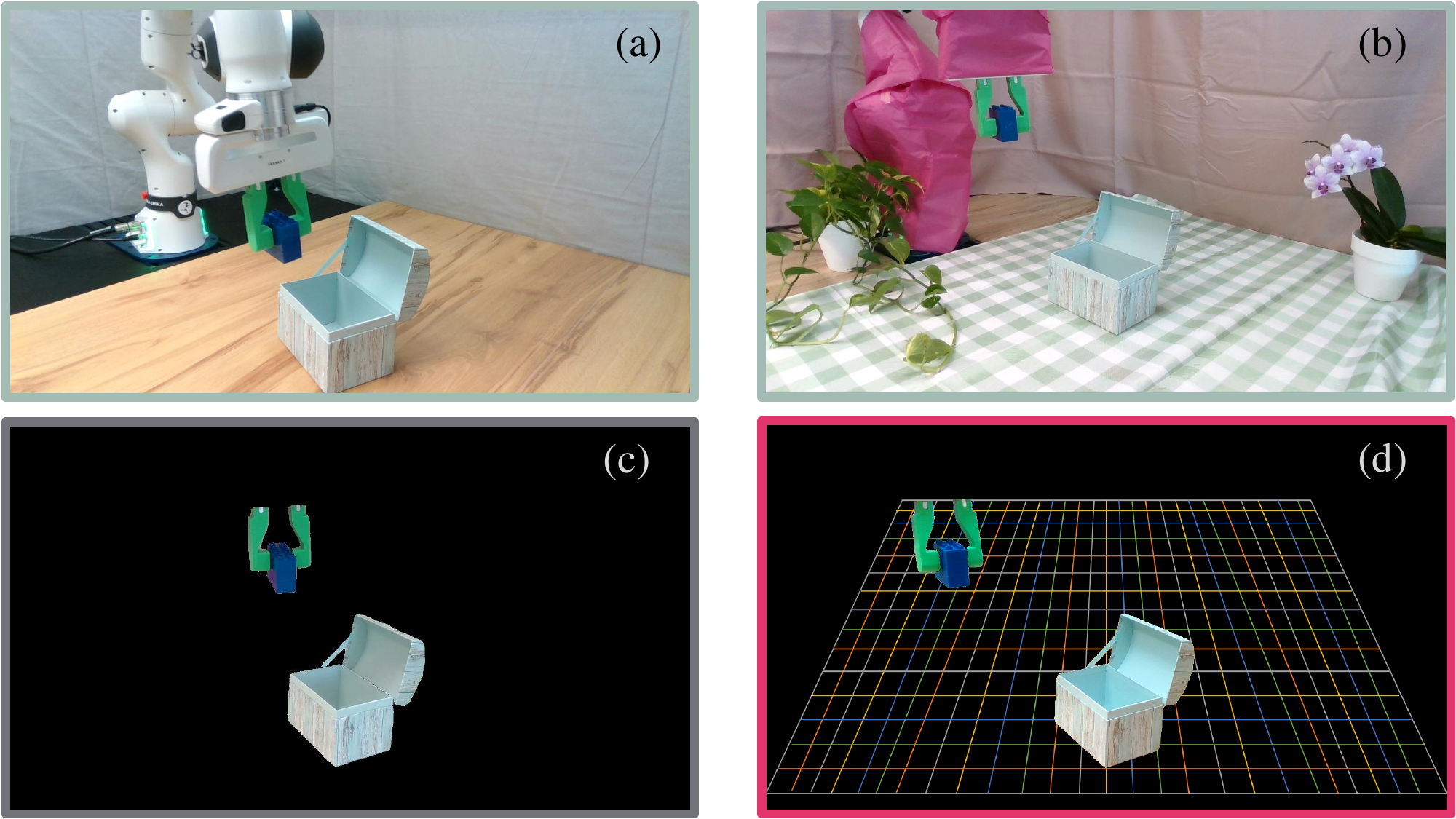}  
\caption[Examples of visual inputs used across training and evaluation]{Examples of input representations: (a) training-time scene, (b) evaluation-time scene with visual changes, (c) input used by the masked policy, and (d) ARRO’s augmented view.}
\label{fig:bar_plot_images}
\end{figure}

To assess the effectiveness of our approach, we consider three variants of Diffusion Policy on each task, each representing a distinct visual input strategy: 

\begin{enumerate}
    \item \textit{Vanilla Diffusion Policy}, which operates directly on unaltered RGB frames.
    \item \textit{Masked Diffusion Policy}, which isolates the robot gripper and target object via segmentation and places them against a plain black background.
    \item \textit{ARRO}, which similarly extracts the relevant components but overlays them onto a structured grid background designed to provide useful visual cues.
\end{enumerate}

Figure~\ref{fig:bar_plot_images} illustrates these input representations using the box-v1 task as an example. 
As depicted in Figure, we introduced a range of visual perturbations during inference to evaluate robustness under real-world domain shifts. These variations include changes in background and tabletop textures, modifications to the robot’s appearance using colored papers, and the placement of unrelated objects in the scene.

All variants of the Diffusion Policy operate within an absolute Cartesian coordinate frame, where each action is represented by a 6-degree-of-freedom (6-DoF) end-effector pose. This includes the position \((x, y, z)\), orientation \((r, p, y)\), and a discrete gripper command (open or closed).
We use a ResNet-18 encoder pretrained on ImageNet to process image observations. Input images are resized to \(320 \times 180\) and augmented with a 90\% random crop. Training is performed with a batch size of 16 for 1000 epochs. For evaluation, each policy is assessed over 10 rollout episodes per task.
The action sequence length is adjusted based on task complexity: 14 steps for pick-v1 and box-v1, and 6 steps for push-v1 and doll-v1. At each timestep, the policy receives both the current and previous image-action pairs as input, promoting temporal consistency in decision-making. Evaluation parameters for each real-world task are summarized in Table~\ref{tab:evaluation-params}.

\begin{table}[t]
\centering
\caption{Experimental settings for real-world evaluations with Diffusion Policy across manipulation tasks.}
\label{tab:evaluation-params}
\begin{tabular}{lccc}
\toprule
\textbf{Task} & \textbf{Action Horizon} (\( T_a \)) & \textbf{Observation Horizon} (\( T_o \)) & \textbf{Evaluation Trials} \\
\midrule
pick-v1   & 14 & 2 & 10 \\
push-v1   & 6  & 2 & 10 \\
box-v1    & 14 & 2 & 10 \\
doll-v1   & 6  & 2 & 10 \\
\bottomrule
\end{tabular}
\end{table}

\begin{figure}
\centering
\includegraphics[clip,trim=0cm 0cm 0cm 0cm,width=1\linewidth]{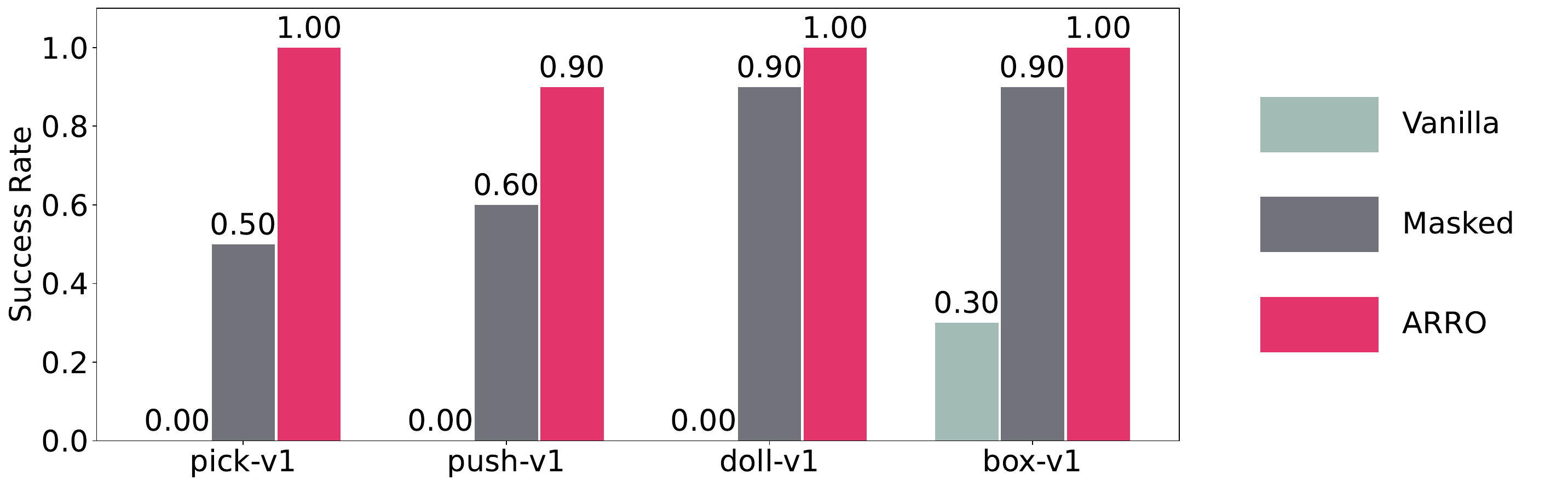}  
\caption[Task success rates for real-world manipulation under visual variation and domain shift]{Task success rates for pick-v1, push-v1, doll-v1, and box-v1 using Vanilla, Masked, and ARRO variants of Diffusion Policy. ARRO shows consistently higher performance under visual domain shifts.}
\label{fig:bar_plot}
\end{figure}

As illustrated in Figure~\ref{fig:bar_plot}, ARRO consistently achieves superior performance compared to both baseline methods across all tasks. The Vanilla Diffusion Policy, which operates directly on unprocessed RGB frames, exhibits significant performance degradation, primarily due to its sensitivity to visual variability. The Masked Diffusion Policy, which improves robustness by discarding visual distractions, offers moderate gains; however, its performance is constrained—likely because the removal of all contextual background information also eliminates spatial cues that could aid task execution. In contrast, ARRO selectively filters out irrelevant content while reintroducing structure through a repeatable grid-based background. This design enhances visual consistency across domains, preserves essential spatial information, and facilitates more reliable task execution under domain shifts.

\begin{figure}[h]
\centering
\includegraphics[clip,trim=0cm 6cm 0cm 0cm,width=1\linewidth]{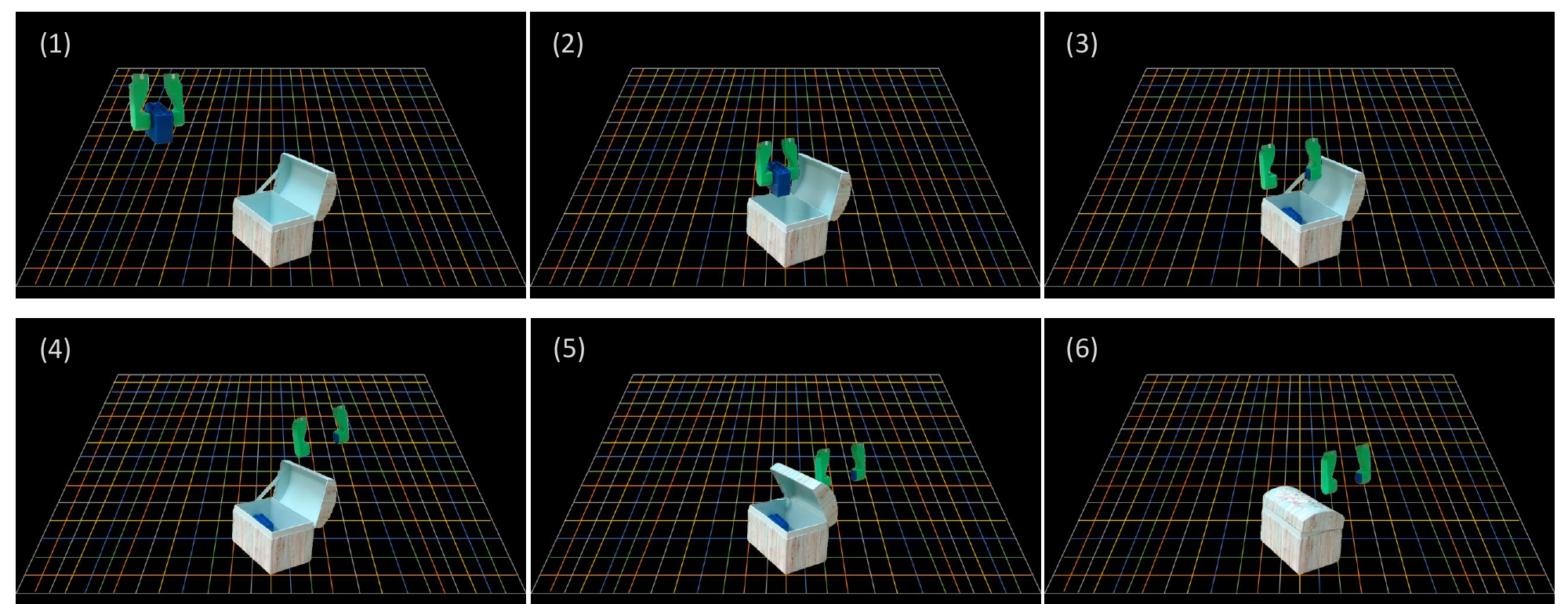}

\vspace{0.5em}  

\includegraphics[clip,trim=0cm 6cm 0cm 0cm,width=1\linewidth]{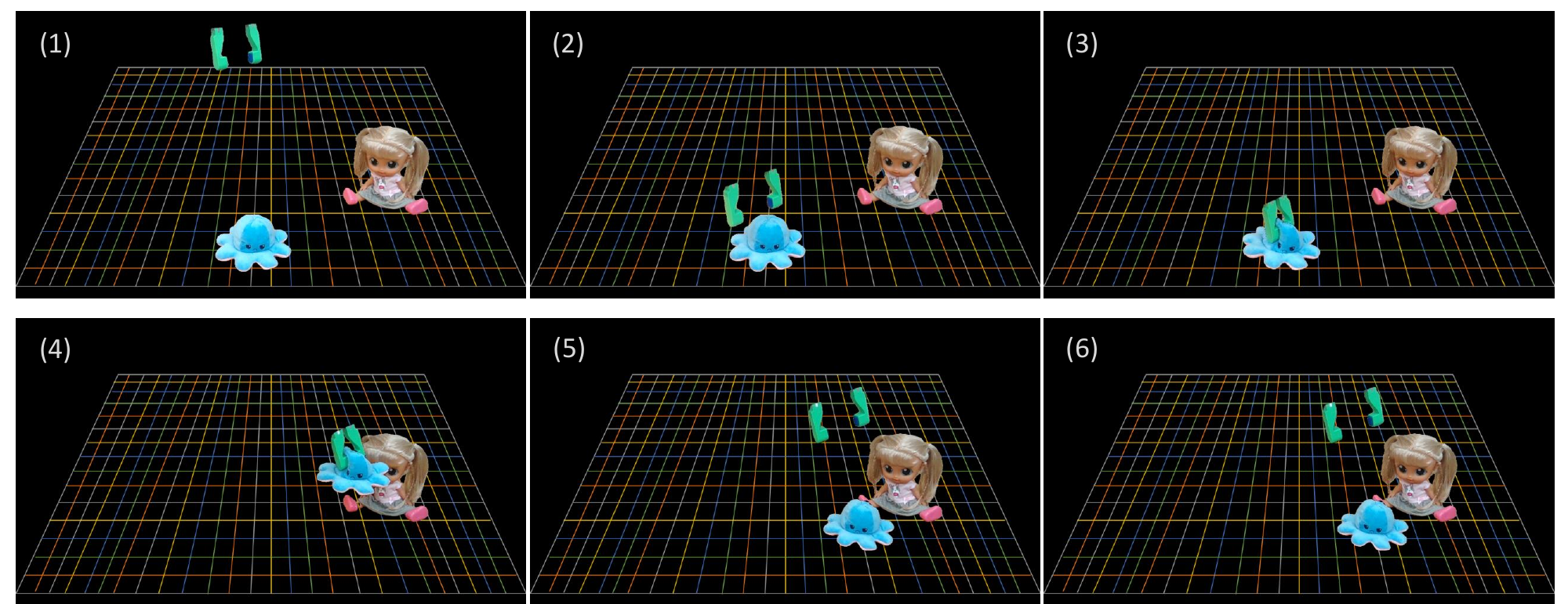}

\caption[Rollout examples for box-v1 and doll-v1 tasks using the ARRO pipeline]{Representative execution rollouts for box-v1 and doll-v1 using ARRO. The approach maintains reliable segmentation even under complex visual conditions, including rich textures and non-rigid object deformations. Task-relevant regions are overlaid onto a structured virtual background, neutralizing visual domain shifts.}
\label{fig:arro_sequence_combined}
\end{figure}

To provide additional qualitative insights, Figure~\ref{fig:arro_sequence_combined} presents sample sequences for the ARRO approach on box-v1 and doll-v1 tasks. The segmentation process demonstrates strong temporal consistency, even under visually challenging scenarios. Despite the presence of complex object shapes and rich textures, the segmentation remains robust across frames, enabling precise and reliable policy execution throughout the task sequence.

\begin{figure}
\centering
\includegraphics[clip,trim=0cm 8.9cm 0cm 0cm,width=1\linewidth]{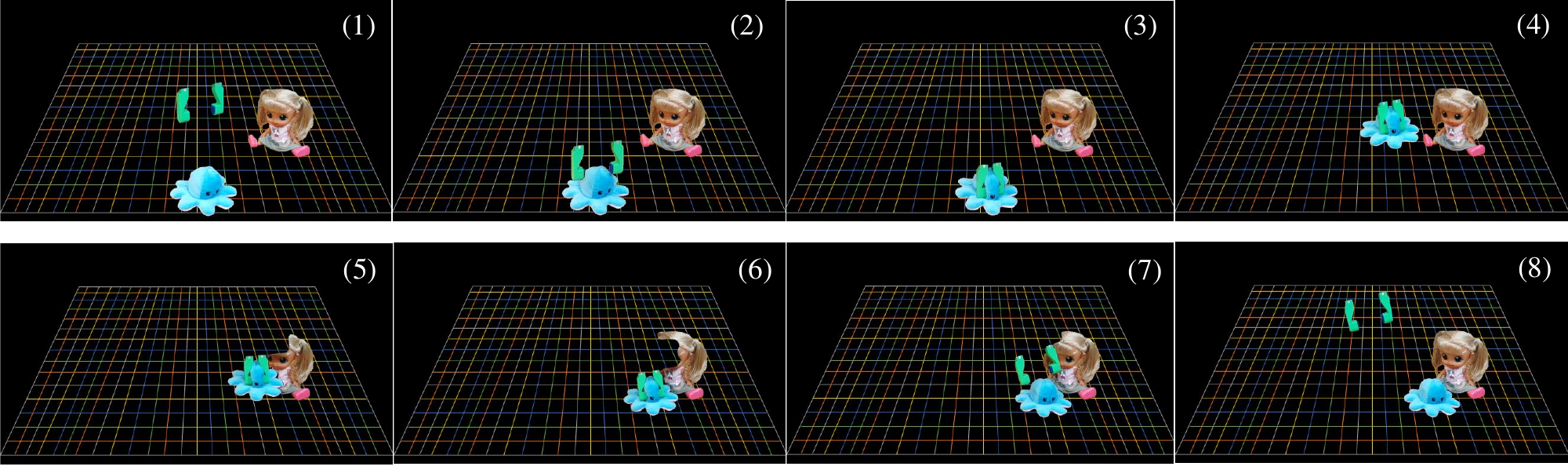}  
\caption[Robust segmentation under occlusion using ARRO.]{ARRO maintains segmentation accuracy despite temporary occlusions. In frames (5)–(7), the doll is partially obscured by the robot arm or the plush toy, yet the segmentation recovers automatically once visibility is restored—without requiring manual intervention.}

\label{fig:occlusion}
\end{figure}

Notably, ARRO also demonstrates robustness to partial occlusions—an important capability given that such occlusions frequently happen in both real-world deployments and collected datasets. They often arise due to the robot’s own movements or the presence of other objects in the scene. An example sequence from the doll-v1 task is shown in Figure~\ref{fig:occlusion}, where the target doll becomes temporarily hidden in frames (5), (6), and (7) due to obstruction by the robot arm and the octopus plush toy. 
Even in the presence of temporary occlusions, ARRO continues to deliver stable and accurate segmentation throughout the episode. When the obstructing objects move out of view, the segmentation of the doll seamlessly resumes—requiring no manual correction or reinitialization. This behavior highlights the reliability of our segmentation pipeline, which preserves focus on task-relevant regions across time. Such robustness to transient visual disruptions is critical for maintaining reliable visuomotor control in complex, cluttered, and continuously changing environments.

\subsubsection{Spatial Reasoning with Distractors}

\begin{figure}
\centering
\includegraphics[clip,trim=0cm 12cm 0cm 0cm,width=1\linewidth]{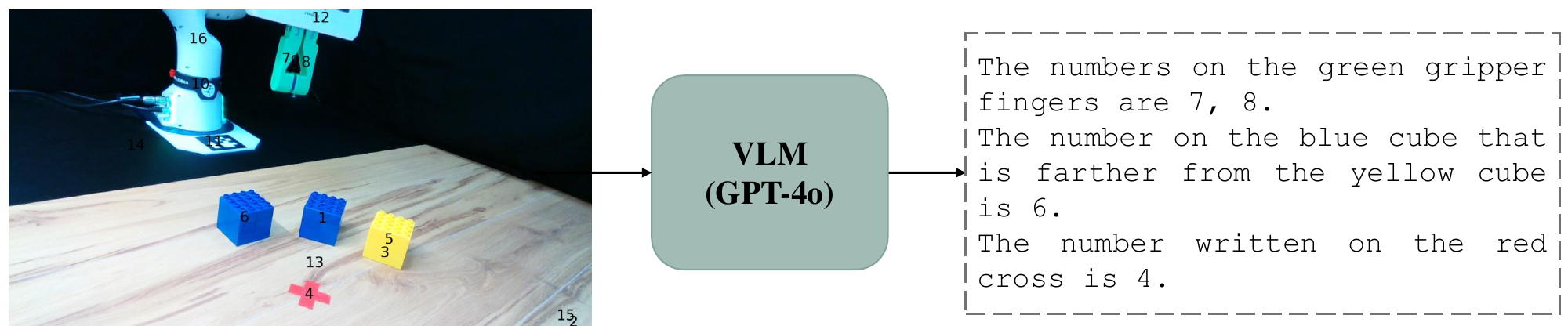}  
\caption[Initialization of segmentation using vision-language prompting]{Segmentation initialization using a vision-language model. The scene is partitioned into labeled regions, each marked with an enlarged numerical identifier for visibility. This annotated frame is then processed by GPT-4o, which is prompted to locate specific elements—such as the \textit{gripper fingers}, the \textit{red cross}, and the \textit{blue cube that is farther from the yellow cube}. }
\label{fig:annotated}
\end{figure}

As described in Section~\ref{sec:approach}, each episode begins with ARRO performing segmentation on the initial frame to isolate key visual elements, such as the robot’s gripper and task-relevant objects. Each identified segmentation region is assigned a distinct numeric label, as illustrated in Figure~\ref{fig:annotated}, to facilitate clear reference and interpretation. This annotated frame is then passed to a vision-language model, such as GPT-4o, which processes natural language prompts to identify and reason about the relevant components. This approach enables not only accurate detection of specific elements—such as the gripper fingers—but also supports the execution of more complex behaviors guided by linguistic descriptions. Prompts may refer not only to simple visual attributes, such as \textit{``green gripper fingers''} or \textit{``red cross''}, but also to more complex properties, including spatial relationships like \textit{``the blue cube farther from the yellow cube''}. In this section, we evaluate ARRO on tasks that require spatial reasoning in cluttered environments with distractor objects.

\begin{figure}
\centering
\includegraphics[clip,trim=0cm 11cm 0cm 0cm,width=1\linewidth]{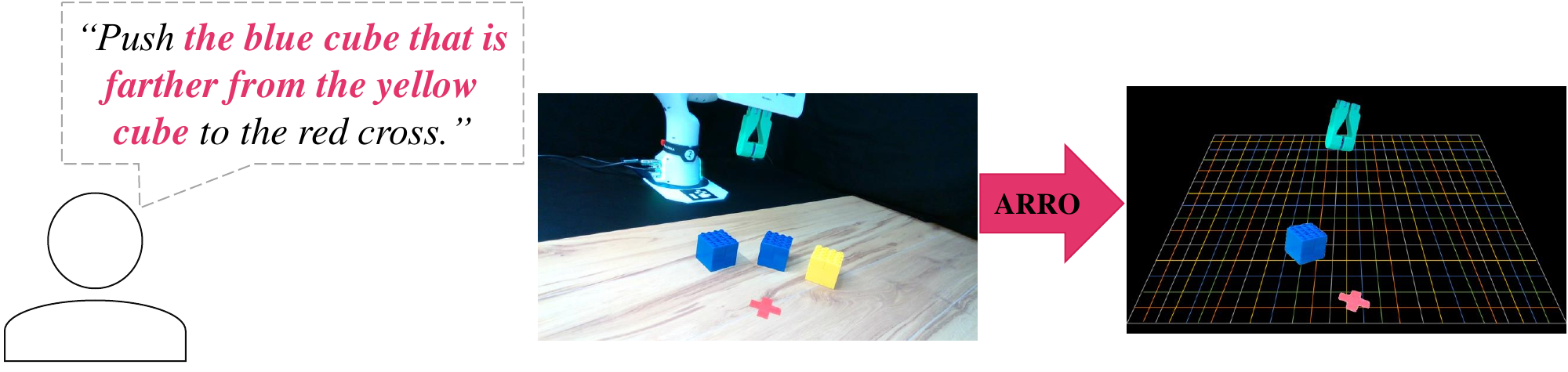}  
\caption[Input scenes before and after ARRO processing for a spatial reasoning task.]{Visual examples for spatial reasoning with distractors. The task is to ``push the blue cube that is farther from the yellow cube to the red cross''. The images show the unmodified input scene and the corresponding ARRO version, where only task-relevant elements are retained and overlaid on a structured virtual grid.}
\label{fig:distractor}
\end{figure}

For this purpose, we extend the push task by introducing distractor objects into the environment. The task configuration is illustrated in Figure~\ref{fig:distractor}. This setup tests whether the system can move beyond simple object detection (e.g., identifying ``\textit{the blue cube}'') and instead interpret spatial cues embedded in natural language, such as ``\textit{the blue cube that is farther from the yellow cube}.'' We design two task variants for this purpose. In the first, the robot receives commands like ``push the \textit{blue cube on the left/right} to the red cross,'' while the second variant increases complexity by referencing relative positions, e.g., ``push the \textit{blue cube that is closer to/farther from the yellow cube} to the red cross.'' As shown in Figure~\ref{fig:annotated} and Figure~\ref{fig:distractor}, the Vision-Language Model accurately parses these linguistically complex prompts and grounds the correct object, even under visual clutter. Once the target is identified, ARRO filters the scene by masking out distractors, enabling the visuomotor policy to act solely on task-relevant observations. This approach allows the system to resolve ambiguity and execute actions based on nuanced spatial relationships.

\begin{figure}
\centering
\includegraphics[clip,trim=0cm 0cm 0cm 0cm,width=0.8\linewidth]{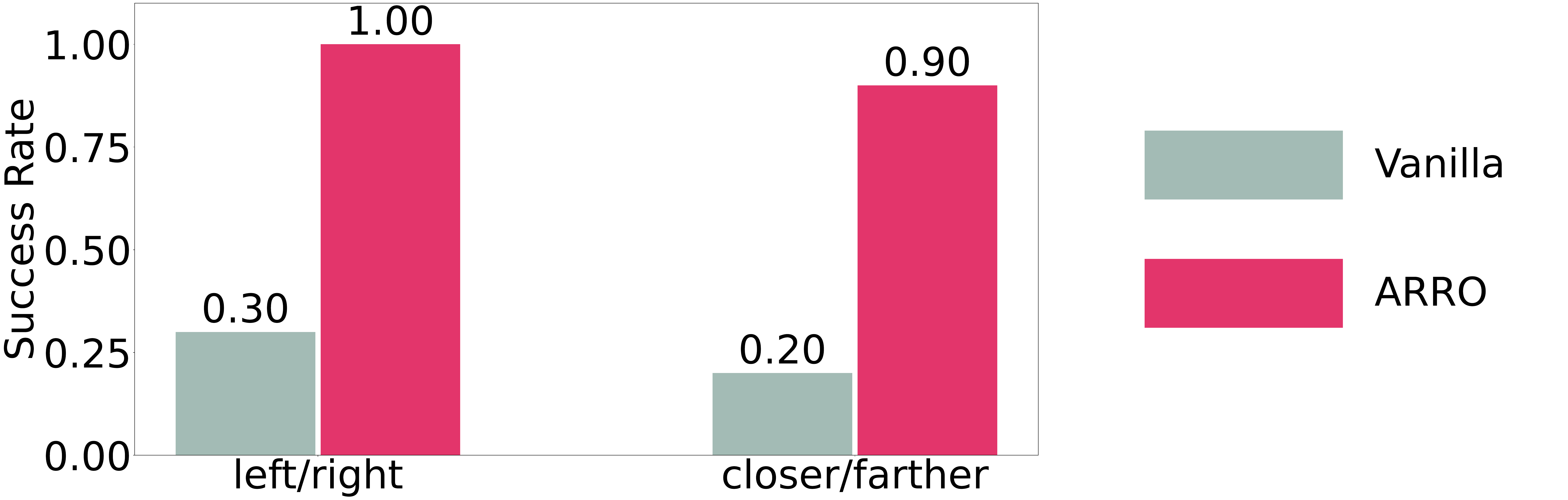}  
\caption[Success rates for spatial reasoning instructions using Vanilla and ARRO approaches.]{Success rates for spatial reasoning tasks under visual domain shifts. We evaluate two instruction types: \textit{left/right} and \textit{closer/farther}. ARRO significantly outperforms the vanilla baseline, highlighting its ability to support tasks requiring spatial reasoning.}
\label{fig:distractor_barplot}
\end{figure}

As depicted in Figure~\ref{fig:distractor_barplot}, ARRO demonstrates consistently high accuracy in identifying and manipulating the correct object in both task variants, reflecting a strong grasp of spatial relationships conveyed through language. 
By comparison, the Vanilla Diffusion Policy performs considerably worse, often failing to select the appropriate target—sometimes interacting with distractor objects or displaying inconsistent behavior, such as shifting between multiple candidates or executing actions with unclear intent. Even in successful attempts, the resulting motion trajectories are typically less precise.

This ability to resolve spatial ambiguity plays a critical role in ARRO’s overall performance. Leveraging language-informed reasoning, the VLM not only identifies relevant objects but also incorporates spatial understanding to guide the masking process, ensuring that only task-relevant regions are preserved. As a result, ARRO remains effective even in complex, cluttered scenes where accurate semantic grounding is essential for reliable task execution. The results further highlight ARRO’s strength in handling both visual clutter and spatially grounded instructions. By selectively retaining the most relevant visual components, ARRO enables the policy to stay focused on the intended task—a capability that is especially important for real-world deployment, where spatial commands and distracting elements often challenge robotic perception and control.

\subsubsection{Evaluation on Generalist Policies*}
\footnotetext{* This subsection presents results contributed by my coauthor Tobias Jülg, as part of our joint work~\citep{mirjalili2025augmented}.}

In addition to evaluating task-specific diffusion policies, we explore how ARRO can enhance the robustness of generalist policies, such as Octo~\cite{octo} and OpenVLA~\cite{openvla}. To this end, we examine these models on two representative manipulation tasks: push-v1 and pick-v2. Both models are fine-tuned using the same task prompts, to ensure consistency across experiments.

For Octo, we fine-tuned the octo-base-1.5 checkpoint for 20,000 steps on both datasets. We used a batch size of 128 and a window size of 1, applying full fine-tuning across all model weights. The pick-v2 dataset consists of 463 real-world demonstrations of a Franka Research 3 (FR3) robot performing picking up a red cube. All input images were resized to $256 \times 256$ using a Lanczos filter prior to training.
OpenVLA was also fine-tuned for 20,000 steps on the same tasks. Here, we used a smaller batch size of 6 and a learning rate of $2 \times 10^{-5}$. Input frames were first resized to $256 \times 256$ using a Lanczos filter and then internally downscaled to $224 \times 224$ using OpenVLA’s built-in resizing.

To assess generalization, we introduced significant visual domain shifts during inference—such as changes in the background and robot appearance—to simulate deployment in unfamiliar settings. Table~\ref{tab:vla} reports performance under three input configurations: (1) \textit{Vanilla}, using the original RGB images; (2) \textit{Masked}, where the segmented foreground is placed on a black background; and (3) \textit{ARRO}, which replaces the background with a structured, repeatable grid to provide consistent visual cues. For reference, we also include the success rates of the Vanilla variant tested under the original visual distribution as a performance upper bound.

The results reveal that both generalist models benefit from preprocessing with Masked and ARRO inputs, achieving higher success rates than with unaltered RGB frames. Among the two, ARRO consistently provides the strongest performance under domain shifts, likely due to its ability to retain essential spatial context while filtering out irrelevant visual clutter. These findings underscore ARRO’s effectiveness in improving the robustness and reliability of generalist visuomotor models across visual variations in the environment.

\begin{table*}
\centering
\footnotesize
\caption[Performance of OpenVLA and Octo under altered visual conditions]{Evaluation of OpenVLA and Octo on the push-v1 and pick-v1 tasks under conditions of visual domain shift. Both models are trained in the original, unaltered environment and subsequently tested on either the same setup or a modified version featuring visual perturbations. This table is dapted from Mirjalili~\etal~\cite{mirjalili2025augmented}.}
\label{tab:vla}
\begin{tabular}{llcccccccc}
\toprule
& & \multicolumn{4}{c}{\textbf{Push}} & \multicolumn{4}{c}{\textbf{Pick}} \\
\cmidrule(lr){3-6} \cmidrule(lr){7-10}
& \textbf{Policy} 
& \textbf{Same Scene} & \multicolumn{2}{c}{\textbf{Altered Scene}} &  
& \textbf{Same Scene} & \multicolumn{3}{c}{\textbf{Altered Scene}} \\
& & \textbf{Vanilla} & \textbf{Vanilla} & \textbf{ARRO} & 
  & \textbf{Vanilla} & \textbf{Vanilla} & \textbf{Masked} & \textbf{ARRO} \\
\midrule
& Octo     & 0.3 & 0   & 0.10 & & 0.5 & 0   & 0.3 & 0.4 \\
& OpenVLA  & 0.1 & 0   & 0.05 & & 0.6 & 0   & 0.4 & 0.4 \\
\bottomrule
\end{tabular}
\end{table*}

\subsection[Simulation-Based Evaluation]{Simulation-Based Evaluation*}
\footnotetext{* This section presents results contributed by my coauthor Tobias Jülg, as part of our joint work~\citep{mirjalili2025augmented}.}

To further assess ARRO's performance, we conducted additional evaluations in simulation, targeting both real-to-sim transfer and cross-embodiment scenarios using MuJoCo. For every condition, models were evaluated over 100 rollout episodes to ensure statistical reliability. In the cross-embodiment experiments, we replaced the Franka Emika Research 3 (FR3) arm with a UR5e manipulator, while preserving the rest of the environment to isolate the impact of embodiment changes. Both robot models were obtained from the MuJoCo Menagerie~\cite{menagerie2022github}. These evaluations provide a rigorous test of policy transferability across embodiments and domain shifts. 

In order to evaluate ARRO’s capabilities in simulation, we use two datasets: pick-v2 (real-world) and sim-pick-v2 (simulated). Both tasks involve picking up a red cube a red cube from the table surface. The sim-pick-v2 dataset replicates the pick-v2 task within a MuJoCo~\cite{mujoco} simulation environment and consists of 900 episodes.

\begin{figure}[t]
  \centering
  \includegraphics[width=0.32\linewidth, trim=0 0 0 0, clip]{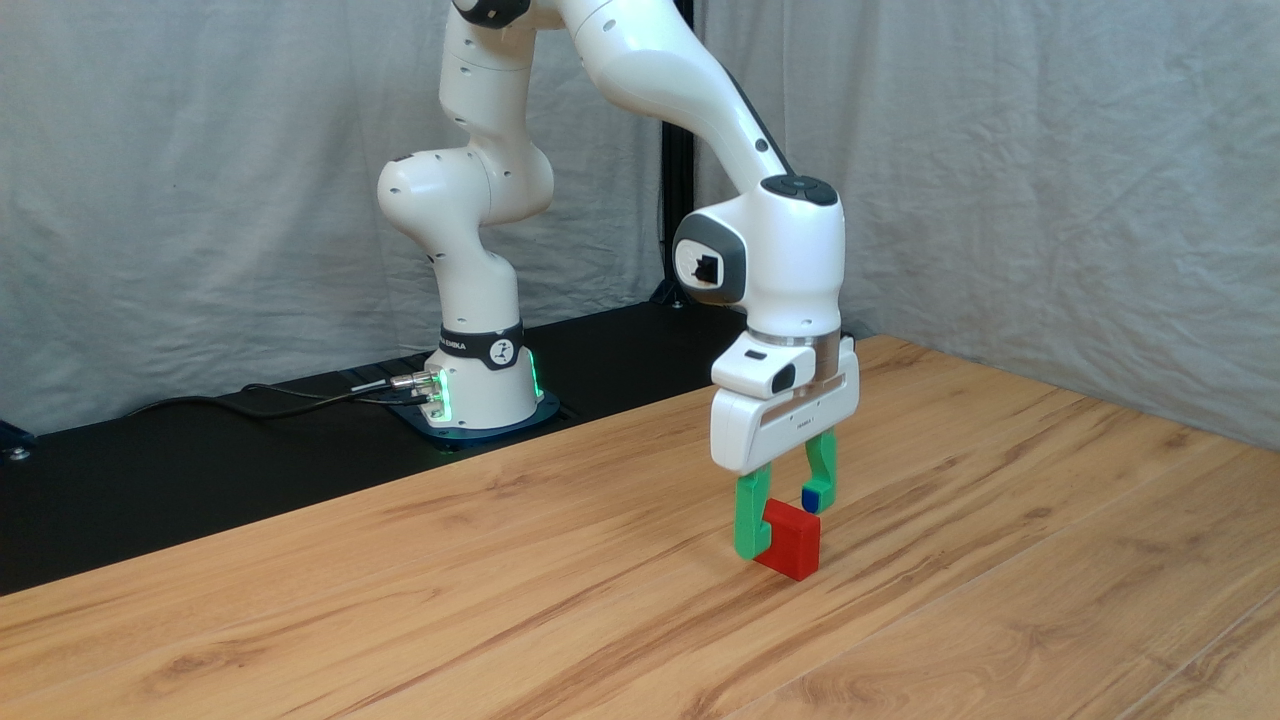}
    \hfill
  \includegraphics[width=0.32\linewidth, trim=0 0 0 0, clip]{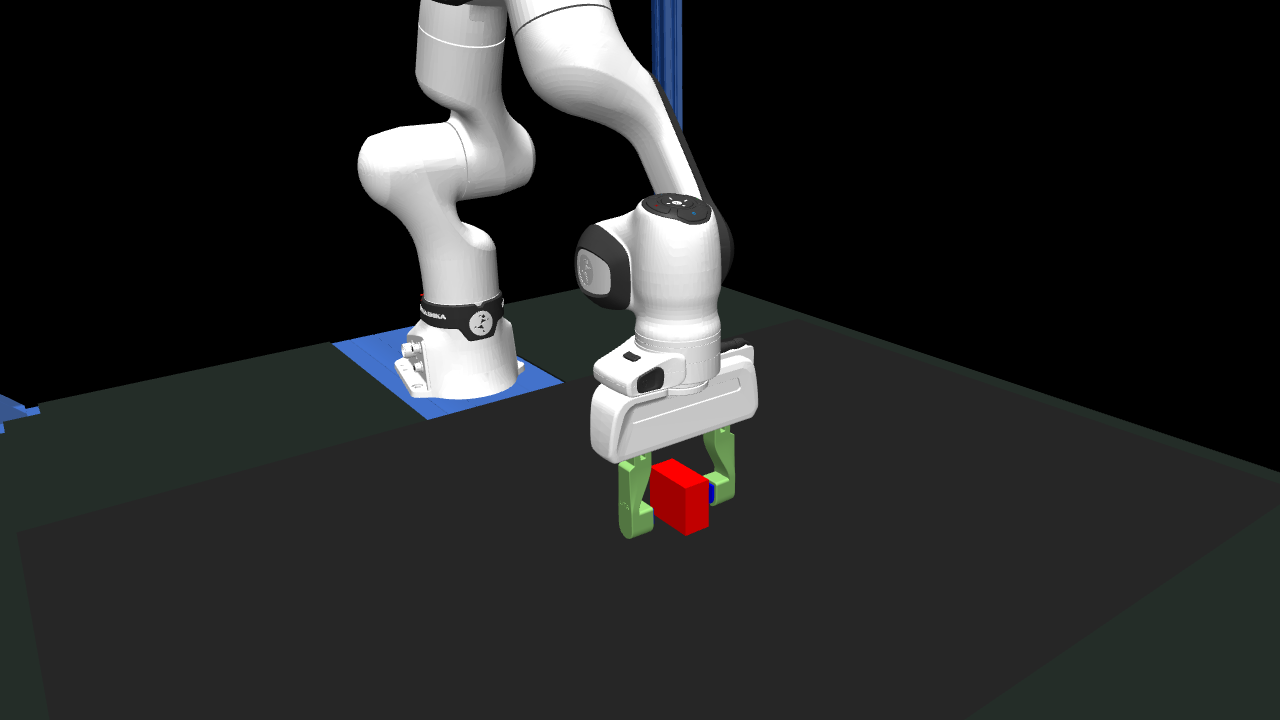}
      \hfill
  \includegraphics[width=0.32\linewidth, trim=0 0 0 0, clip]{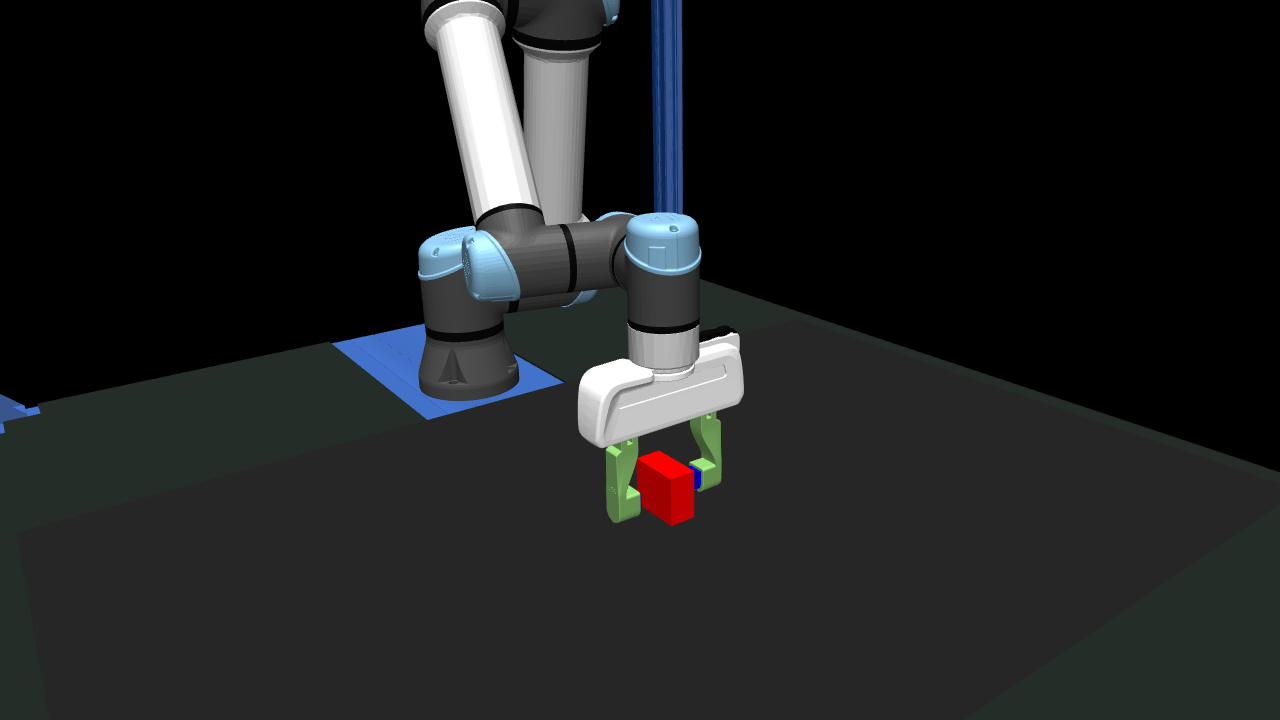}
\caption[Setup for the pick task across real, simulated, and cross-embodiment environments]{
Overview of the experimental configurations for the pick task across real and simulated settings. \textit{From left to right}: the real-world setup using the FR3 arm, its MuJoCo-based simulation for the sim-pick-v2 dataset, and a UR5e-based variation used for cross-embodiment evaluation in simulation. This figure is adapted from Mirjalili~\etal~\cite{mirjalili2025augmented}.}

  \label{fig:sim_exp_setup}
\end{figure}

Figure~\ref{fig:sim_exp_setup} illustrates the experimental configuration, which includes the physical FR3 robot operating in real-world environment, its virtual replica modeled in the \mbox{MuJoCo}~\cite{mujoco} simulator, and a cross-embodiment variant where the FR3 is substituted with a UR5e robotic arm within the same simulated environment. This setup enables a comprehensive evaluation of ARRO’s generalization capacity across two key dimensions of domain shift: discrepancies between simulation and reality, and variations in robot embodiment.

\subsubsection{Cross-Embodiment Evaluation }
\label{sec:x-embodiment}

As mentioned earlier, to assess ARRO’s capacity for cross-embodiment generalization, we design an evaluation setup in which policies trained on one robotic platform are transferred to a different embodiment. Training is performed on the sim-pick-v2 dataset, collected entirely in the MuJoCo simulator using the Franka Emika Research 3 (FR3) manipulator. During inference, these policies are executed on a distinct robotic arm—the UR5e—while keeping the simulated environment unchanged. This setting allows us to isolate the impact of embodiment differences on policy performance.

We present two outcomes using 100 trials: (1) the standard sim-to-sim success rate, reflecting performance on the original training embodiment, and (2) the cross-embodiment success rate, measuring generalization to the UR5e platform. A summary of the results is provided in Table~\ref{tab:sim}.

\begin{table}
\centering
\footnotesize
\caption[Success rates of policy variants under real-to-sim and cross-embodiment transfer]{Comparison of success rates for real-to-simulation and cross-embodiment evaluations under three visual input formats: Vanilla (V), Masked (M), and ARRO. Results span multiple policy architectures, highlighting the impact of each preprocessing method. 
This table is adapted from Mirjalili~\etal~\cite{mirjalili2025augmented}.}
\label{tab:sim}
\resizebox{\textwidth}{!}{%
\begin{tabular}{lcccccc|cccccc}
\toprule
\textbf{Policy} 
& \multicolumn{3}{c}{\textbf{Real-to-Real}} 
& \multicolumn{3}{c}{\textbf{Real-to-Sim}} 
& \multicolumn{3}{c}{\textbf{Sim-to-Sim}} 
& \multicolumn{3}{c}{\textbf{Cross-Embodiment}} \\
& V & M & ARRO & V & M & ARRO & V & M & ARRO & V & M & ARRO \\
\midrule
Diffusion$^*$ & 1.0* & 0.5* & 1.0* & 0 & 0.00 & 0.00 & 0.95 & 0.15 & 1.00 & 0.00 & 0.12 & 0.99 \\
Octo          & 0.5 & 0.3 & 0.4 & 0 & 0.01 & 0.02 & 0.17 & 0.08 & 0.16         & 0.12 & 0.07 & 0.14 \\
OpenVLA       & 0.6 & 0.4 & 0.4 & 0 & 0.16 & 0.22 & 0.14 & 0.50 & 0.51         & 0.04 & 0.36 & 0.41 \\
\bottomrule
\end{tabular}
}
\end{table}

The results show that Diffusion Policy, trained exclusively on the FR3 robot, fails to transfer its capabilities to the UR5e embodiment, achieving a 0.00 success rate in cross-embodiment evaluations. This result is to be expected since the model was not exposed to the new embodiment during training phase and thus cannot handle the visual domain shifts caused by the embodiment change. However, when combined with input preprocessing techniques—either black background masking or ARRO—the policy demonstrates improved performance. These masking strategies appear to alleviate embodiment-specific dependencies by completely removing visual cues associated with the robot body, resulting in observation inputs that are largely invariant to the platform. Additionally, since Diffusion Policy generates actions in absolute Cartesian space, its outputs remain compatible across different robots, regardless of their kinematic differences.

Compared to Diffusion Policy, Vanilla generalist models such as Octo and OpenVLA exhibit more robustness when faced with embodiment shifts. This robustness can be attributed in part to their exposure to more extensive and diverse training datasets, which include various robot types such as UR5 robot. Still, when paired with either ARRO or the black masking approach, they both exhibit stronger performance on the target UR5e platform.

\subsubsection{Real-to-Simulation Transfer}
\label{sec:real-to-sim}

To evaluate ARRO’s impact on generalization across domains, we measure real-to-simulation transfer performance for three policy models: \textit{Diffusion Policy}, \textit{Octo}, and \textit{OpenVLA}. All models were trained on real-world demonstrations from the pick-v2 dataset, with one exception—Diffusion Policy (indicated by an asterisk in Table~\ref{tab:sim}) was trained on pick-v1.

We evaluated each policy in two settings. The first involved measuring in-distribution performance by conducting 10 trials in the original real-world environment used during training, establishing a baseline for real-to-real success. The second setting focused on out-of-distribution generalization, where the same models were tested in a MuJoCo-based simulation designed to closely replicate the real-world setup. To gauge robustness under domain shift, this real-to-sim evaluation was carried out over 100 simulated episodes.

Table~\ref{tab:sim} summarizes the results, highlighting a sharp decline in performance for all models when transitioning from the real-world environment to simulation without any form of visual preprocessing—often resulting in near-zero success rates. Among the evaluated models, OpenVLA demonstrates the strongest performance, retaining a success rate of 0.22 when paired with ARRO. This suggests that input preprocessing plays a key role in enhancing cross-domain generalization. Octo performs slightly better when paired with ARRO, while Diffusion Policy fails to transfer altogether—likely due to its limited training dataset and architectural differences in its visual encoder.

\begin{figure}
  \centering
  \includegraphics[width=\linewidth]{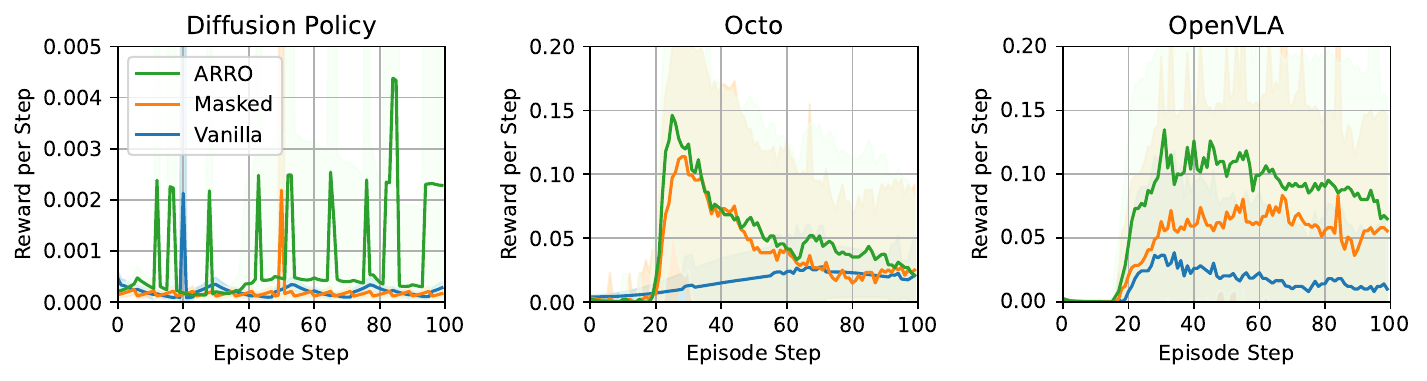}
\caption[Performance comparison on simulated pick-cube task based on evaluation reward]{Mean normalized reward per timestep for the simulated pick-cube task using the FR3 robot. Models were trained on real-world pick-v1 demonstrations and evaluated across 100 simulated episodes. Shaded regions indicate standard deviation. This figure is adapted from Mirjalili~\etal~\cite{mirjalili2025augmented}.}

  \label{fig:real2sim_rewards}
\end{figure}

To complement success rate analysis, we also assess behavioral quality using a reward function based on the distance between the manipulated object and the goal location, following a metric inspired by ManiSkill~\cite{maniskill}. As illustrated in Figure~\ref{fig:real2sim_rewards}, both ARRO and the black masking baseline achieve substantially higher average rewards than the vanilla input, further emphasizing the benefits of visual simplification for real-to-sim transfer performance.

\clearpage
\section{Limitations}
\label{sec:limitations}

While ARRO demonstrates strong performance across various tasks, its effectiveness is influenced by several practical considerations. A key dependency lies in its reliance on recent open-vocabulary segmentation and object detection models, which serve as the basis of its visual preprocessing pipeline. In our implementation, we employ SAM ~\cite{sam} for initial segmentation and Grounding DINO~\cite{groundingdino} for object detection in the first frame. These models support zero-shot inference without task-specific retraining, enabling rapid adaptation to new environments. However, they introduce inference latency that could be a problem for time-critical tasks. Moreover, while minimal prompt tuning sufficed for the domains explored in this work, different operational settings may require more extensive prompt engineering or domain-specific adjustments. Fortunately, ARRO's modular design allows these components to be swapped with faster or more specialized alternatives, offering flexibility in adapting to new task constraints or newer models.

Another practical limitation arises from ARRO’s assumption of a static initial scene. The current pipeline expects all task-relevant objects to be visible in the first frame, which restricts its applicability in scenarios involving dynamic object appearances, movements, or state changes during execution. Addressing this would require dynamic scene updates, potentially achieved by periodically querying the vision-language model at key decision points, such as after object interactions, to capture changes and refresh segmentation prompts. This would enable ARRO to remain responsive and relevant as scenes evolve over time.

Hardware-specific constraints also impact the generality of ARRO. In our setup, the robot's gripper has a distinctive color that facilitates segmentation, aligning with conventions established in prior visuomotor learning works~\cite{umi}. However, such visual modifications may not be practical or desirable across all robotic platforms. For grippers lacking distinctive visual cues, a potential workaround is to initialize each episode from a predefined home position. This consistency allows segmentation to be initialized using a fixed set of keypoints or bounding boxes, enabling reliable initialization across sequences.

Finally, although ARRO effectively filters out background clutter by replacing it with a structured visual canvas, it does not address all sources of visual variation. Lighting conditions and surface reflections, for instance, can alter the appearance of task-relevant objects and affect policy performance. While our experiments showed that moderate lighting changes had minimal impact, more extreme variations could lead to drops in policy performance. One promising direction to mitigate these effects involves applying image inpainting methods~\cite{genaug} to normalize object textures and enhance robustness under visual variations. Together, these considerations outline both the strengths and practical boundaries of ARRO’s current design, while pointing to clear avenues for future improvement.

\section{Conclusion}
\label{sec:conclusion}

In this chapter, we presented ARRO, a visual preprocessing framework aimed at enhancing the robustness of visuomotor policies when faced with visual domain shifts. ARRO improves robustness without the need for calibration or retraining by segmenting out the robot gripper and task-relevant objects and overlaying them onto a structured, virtual background. This compositional strategy effectively removes distracting visual elements while preserving essential task information, resulting in a consistent and generalizable input format.

Extensive evaluations conducted across both physical and simulated environments show that ARRO substantially enhances the robustness of visuomotor control. These gains are evident in both specialized models, such as Diffusion Policy, and broader language-conditioned generalist frameworks like Octo and OpenVLA. The improvements persist across a wide range of challenging conditions—including background variations, changes in robot embodiment and the presence of distractor objects—highlighting ARRO’s versatility and generalization capabilities.

Future work could explore scaling ARRO to large and diverse datasets, such as Open X-Embodiment~\citep{openx}, to further enhance generalization across both tasks and robot embodiments. Another promising direction is to improve ARRO’s robustness to variations in camera placement. In real-world deployments, cameras are often repositioned—intentionally or otherwise—and even slight shifts in camera angle or location can significantly degrade policy performance, particularly for task-specialized models such as diffusion policies. Addressing this challenge may involve incorporating 3D scene reconstruction or augmenting the dataset with real or synthesized views from novel perspectives. Collectively, these extensions would position ARRO as a more general and scalable framework for robust visuomotor learning in complex, real-world robotic environments.

%% file: 3-conclusion.tex
\chapter{Conclusion and Discussion}
\label{chapter:conclusion}

This thesis investigated how foundation models can be leveraged to build robotic systems capable of operating in unstructured, real-world environments. The central aim was to move beyond robotics research confined to controlled laboratory settings and demonstrate robust, semantically informed, and generalizable behavior in complex, everyday scenarios. Through four core contributions, this work showed that integrating large language models (LLMs) and vision-language models (VLMs) into robotic perception and action pipelines significantly enhances autonomy, task comprehension, and adaptability—without the need for retraining or additional data collection.

\sloppy
The first contribution, FM-Loc, introduced a visual place recognition system that constructs semantic descriptors of indoor scenes using foundation models. This method achieved robust localization across wide variations in viewpoint and visual appearance without requiring additional data collection or fine-tuning, making it suitable for real-world deployment.
The second contribution, Lan-grasp, developed a semantically guided object grasping system. By leveraging reasoning from both vision and language models, Lan-grasp generated grasps that were not only functionally appropriate but also aligned with human preferences.
The third contribution, VLM-Vac, explored the integration of a vision-language model into smart robotic vacuum cleaners. This approach distilled knowledge from the VLM into a lightweight model and employed language-guided continual learning, enabling both adaptability and efficiency while reducing dependency on frequent model queries.
The fourth and final contribution, ARRO, introduced a visual augmentation pipeline that filtered out distractors and retained only task-relevant regions. This method substantially improved visuomotor policy robustness and generalization across different embodiments and visual scene modifications.

Together, these contributions demonstrated that foundation models can serve as flexible, zero-shot reasoning modules for embodied agents. A central insight was that real-world robotics cannot rely on continually collecting datasets or retraining models for every new situation. This research reflected a broader shift away from narrowly engineered systems toward robots that demonstrated contextual reasoning and semantic understanding. Across all four projects, such semantic grounding enabled more purposeful and reliable decision making—such as identifying optimal grasp points or distinguishing between objects to clean or avoid in domestic environments. Importantly, foundation models were not limited to perception tasks; they were used as reasoning engines that connected visual understanding to action in meaningful, goal-directed ways.

Despite the progress, several limitations persist. One major constraint is the inherent dependency on foundation models themselves. While these models offer impressive generalization, they are still prone to occasional mistakes. 
Their performance can degrade in ambiguous or out-of-distribution scenarios due to biases in their training data. 
For now, their application is most appropriate in contexts that tolerate occasional errors. However, with the rapid evolution of vision-language models and related technologies, we can expect improvements in both their capability and reliability. This evolving landscape underscores the importance of continually reassessing model choices and refining prompting strategies for optimal performance.

Another limitation lies in the computational demands of foundation models. Their inference typically requires either high-end local hardware or remote API access, which limits practical deployment in edge or resource-constrained settings. To address this, Chapter~\ref{chapter:vacuum} introduced a distillation framework that compressed the knowledge of a VLM into a smaller, more efficient model, complemented by a continual learning mechanism that allows the student model to incrementally learn new knowledge over time. While this approach reduces inference costs and allows for adaptation over time, it remains less semantically expressive than its foundation counterpart. As such, continued progress in developing more efficient foundation models—and in improving hardware acceleration—remains crucial for enabling high-performance, real-time inference directly on robotic platforms without sacrificing generalization or flexibility.

Looking ahead, the landscape of foundation models is rapidly expanding beyond the domains of vision and language—the primary modalities explored in this thesis. Emerging multi-modal architectures that incorporate tactile, audio, and force sensing open new avenues for grounded physical interaction~\cite{yang2023uniaudio, yang2024binding}. These additional sensory channels complement vision and language by enabling robots to perceive the world through richer, more embodied experiences. By fusing diverse modalities, future systems may gain a deeper understanding of object properties, environmental dynamics, and human intent—paving the way for more adaptive, intuitive, and physically capable robotic behavior.
In parallel, recent advances in generative models capable of synthesizing 3D scene representations from images or language prompts~\cite{3d2,3ddiff1} introduce transformative opportunities. Such models could replace traditional 3D reconstruction pipelines, allowing faster and more scalable deployment in complex settings. This is especially impactful for robotic grasping: as discussed in Chapter~\ref{chapter:langrasp}, our prior method relied on a fixed dual RGB-D sensor setup and a turntable to generate 3D mesh models for grasp synthesis. In contrast, 3D foundation models may offer a more flexible and lightweight alternative, producing geometry-rich representations with minimal hardware requirements.
Beyond these immediate use cases, generative 3D models also hold promise for improving visual robustness. Access to 3D representations could help mitigate the effects of camera viewpoint variation—a persistent challenge for visuomotor policies. This may be achieved by augmenting training datasets with synthetic views or reconstructing consistent scene geometry at inference time, aligning perception across training and deployment. As discussed in Chapter~\ref{chapter:arro}, this capability enhances the robustness of visuomotor policies when deployed in unstructured, dynamic real-world settings.
Altogether, the emergence of foundation and generative models across novel modalities unlocks capabilities previously out of reach. As these systems grow in scope and generality, embracing new sensory inputs and model architectures will be essential to advance the frontiers of embodied intelligence and address long-standing challenges in robotics.

Another promising direction lies in integrating foundation models with classical robotic algorithms. For instance, as discussed in Chapter~\ref{chapter:fm-loc}, language-guided semantic perception can be paired with traditional localization techniques to enhance both robustness and contextual awareness. Future research could extend this idea by incorporating more advanced methods—such as particle filters~\cite{particle}—to build hybrid systems that tightly couple the semantic understanding of foundation models  with algorithmic structure of traditional approaches. Such systems have the potential to combine the broad generalization and adaptability of foundation models with the formal guarantees, and interpretability offered by classical robotics. This integration may yield a more effective balance between generalization and precision, enabling robust and reliable performance in diverse and unpredictable environments.

Beyond perception and localization, another critical area for advancing real-world robotics is long-horizon manipulation. Scaling robotic manipulation to complex, multi-step tasks remains a significant challenge. While recent methods such as Lan-grasp, discussed in Chapter~\ref{chapter:langrasp}, illustrate how language can effectively guide object grasping, future extensions could explore more complex manipulations—such as dynamically adapting the grasp for downstream tasks like object handover or tool use. Similarly, the ARRO framework introduced in Chapter~\ref{chapter:arro} could be extended to support multi-stage tasks that require fine-grained visuomotor control, including the operation of household appliances or dexterous manipulation using multi-fingered hands. Evaluating these capabilities in extended-horizon settings would provide deeper insight into the strengths and limitations of current policy learning pipelines and contribute toward the reliable deployment of robots in everyday human environments.

Finally, the long-term deployment of knowledge distillation, as discussed in Chapter~\ref{chapter:vacuum}, presents a promising avenue for future research. These approaches are particularly valuable because they enable the practical application of foundation model knowledge in everyday settings where high-performance hardware may be unavailable. While early results show encouraging efficiency and competence, key questions remain about their capacity to acquire and retain knowledge over extended periods of time. Investigating their behavior across long periods—spanning months or even years of incremental fine-tuning—could yield important insights into their robustness, adaptability, and real-world viability.

In summary, advancing toward truly autonomous, robust, and generalizable robots will require synergistic progress across foundation models, classical algorithms, sensor fusion, and long-term learning strategies. These directions form a rich and exciting research frontier for the years ahead, and the foundation laid in this thesis provides a concrete starting point. Throughout this work, four key contributions were introduced to support this vision. By exploring how foundation models can be leveraged for embodied use, this thesis advanced the development of intelligent, robust, and deployable robotic systems. Each chapter demonstrated how large-scale pretraining can be adapted to real-world robotic contexts—bridging the gap between general-purpose reasoning and the practical demands of physical interaction. Together, these contributions help chart a path toward robotic systems that can perceive, understand, and act intelligently across diverse and unstructured environments.

%% file: 4-remark.tex
\chapter*{Final Thoughts}

\vspace{0.5cm}

Looking back, it has been humbling to realize how quickly the field of AI and robotics
evolves. Some of the problems I tackled early in my PhD—once considered challenging—
could now be approached more elegantly with models and tools that did not exist at the
time. It is a strange and sobering thought that even my most recent work, of which I
am genuinely proud, may soon appear limited in light of what comes next. But rather
than discouraging, this impermanence is what makes the scientific process so meaning-
ful. Progress depends on our willingness to question our assumptions, to embrace new
paradigms, and to keep wondering what might be possible.

\vspace{0.3cm}
As Carl Sagan once said: 

\begin{quote}
\textit{“Somewhere, something incredible is waiting to be known.”} \\

\end{quote}